\documentclass[letterpaper, 10 pt, journal,twoside]{IEEEtran} % Comment this line out if you need a4paper

\IEEEoverridecommandlockouts   % Comment this line out if you need a4paper

\usepackage[colorlinks,bookmarksopen,bookmarksnumbered,citecolor=red,urlcolor=blue]{hyperref}

\usepackage{bm}
\usepackage{amsmath} % assumes amsmath package installed
\usepackage{amssymb}  % assumes amsmath package installed
\usepackage{dsfont}
\usepackage{graphicx}

\usepackage{psfrag,graphicx,epsfig}
\usepackage{epstopdf}
\usepackage{xspace}
\usepackage{subcaption}
\usepackage{float}
\usepackage{placeins}
\usepackage{multirow}
\usepackage{pgf,tikz}
\usepackage{nowidow}
\usepackage{lineno}
\usepackage{xcolor}
\usepackage{color}

\DeclareMathOperator*{\argmax}{arg\,max}
\usepackage{siunitx}
\usepackage{color}
\usepackage{lineno}
\usepackage{algorithmic}% Needed to meet printer requirements.
\usepackage{gensymb}

\allowdisplaybreaks

\allowdisplaybreaks

\newcommand{\pyrobocop}{\textsc{PyRoboCOP}}

\newcommand{\fig}[1]{Fig.~\ref{#1}}
\newcommand{\tab}[1]{Table~\ref{#1}}
\newcommand{\eq}[1]{(\ref{#1})}

\newcommand{\revise}[1]{\textcolor{black}{#1}} % Change 'red' to any color you prefer

\title{
Robust Pivoting Manipulation using Contact Implicit Bilevel Optimization
}

\author{Yuki Shirai$^{\ddagger}$, Devesh K. Jha$^{\ddagger\dagger}$, and Arvind U. Raghunathan$^{\ddagger}$ %   
\thanks{$^{\ddagger}$Yuki Shirai, Devesh K. Jha and Arvind U. Raghunathan are with Mitsubishi Electric Research Laboratories (MERL), Cambridge, MA, USA 02139 {\tt\small \{shirai,jha,raghunathan\}@merl.com}}
\thanks{$^{\dagger}$ Corresponding author}}%}%

\begin{document}

\maketitle

\begin{abstract}
Generalizable manipulation requires that robots be able to interact with novel objects and environment. This requirement makes manipulation extremely challenging as a robot has to reason about complex frictional interactions with uncertainty in physical properties of the object and the environment. In this paper, we study robust optimization for planning of pivoting manipulation in the presence of uncertainties. We present insights about how friction can be exploited to compensate for inaccuracies in the estimates of the physical properties during manipulation. Under certain assumptions, we derive analytical expressions for stability margin provided by friction during pivoting manipulation. This margin is then used in a Contact Implicit Bilevel Optimization (CIBO) framework to optimize a trajectory that maximizes this stability margin to provide robustness against uncertainty in several physical parameters of the object. We present analysis of the stability margin with respect to several parameters involved in the underlying bilevel optimization problem.  We demonstrate our proposed method using a 6 DoF manipulator for manipulating several different objects. \revise{We also design and validate an MPC controller using the proposed algorithm which can track and regulate the position of the object during manipulation.}
\end{abstract}

\maketitle
% \thispagestyle{plain}
% \pagestyle{plain}

%%%%%%%%%%%%%%%%%%%%%%%%%%%%%%%%%%%%%%%%%%%%%%%%%%%%%%%%%%%%%%%%%%%%%%%%%%%%%%%%

%bilevel trajectory optimization algorithm to design a controller that maximizes this stability margin to provide robustness against uncertainty in physical properties of the object.

\section{Introduction}\label{sec:introduction}
\begin{figure*}
    \centering
    \includegraphics[width=0.95\textwidth]{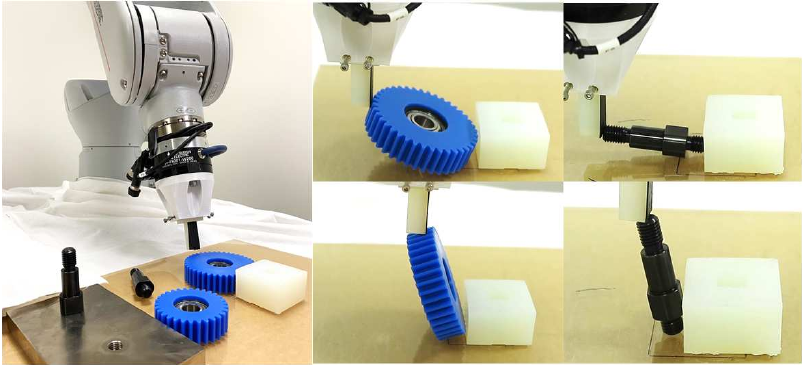} % 
    \caption{We consider the problem of reorienting parts for assembly using pivoting manipulation primitive. Such reorientation could possibly be required when the parts being assembled are too big to grasp in the initial pose (such as the gears) or the parts to be inserted during assembly are not in the desired pose (such as the pegs). The figure shows some instances during the implementation of our  controller to reorient a gear and a peg.}
    \label{fig:pivoting_abstractfig}
\end{figure*}

\IEEEPARstart{C}{ontacts} are central to most manipulation tasks as they can provide additional dexterity to robots to interact with their environment~\cite{mason2018toward}. 
% \revise{For example,  
% in \fig{fig:pivoting_abstractfig}, the objective of the task is assembly. \fig{fig:pivoting_abstractfig} shows that the robot is unable to accomplish grasping since the size of the gear is too large and the peg is an undesired pose. To accomplish the task, we propose to use the pivoting manipulation with extrinsic contacts and thus the robot is able to reorient the parts. Therefore, the robot can successfully conduct the grasping, resulting in the successful assembly.} 
It is desirable that a robot should be able to interact with unknown objects in unknown environments during operation and thus achieve generalizable manipulation. 
\revise{However, designing systems which can achieve such behavior is difficult. Such behavior requires that a robot should be able to reason about and generate plans that are robust to uncertainties arising from a variety of different reasons.} Robust planning for frictional interaction with objects with uncertain physical properties is challenging as the mechanical stability of the object depends on these physical properties. Inspired by this problem, we consider the task of robust pivoting manipulation in this paper. The pivoting task considered in this paper requires that the slipping contact be maintained at the two external contact points which presents unique challenges for robust planning. 
% In particular, we consider the problem of re-orienting objects with uncertain mass and Center of Mass (CoM) location using pivoting.
We are interested in ensuring mechanical stability via friction to compensate for uncertainty in the physical properties (e.g., physical parameters, coefficient of friction, \revise{contact location}.) of the objects during manipulation. We present a novel formulation and an optimization technique that can solve robust manipulation trajectories for manipulation problems.% for the proposed pivoting manipulation. 

%We present a bilevel trajectory optimization method that can maximize this frictional stability under some simplifying assumptions. We present analysis showing that friction can provide stability to uncertainty in mass and CoM location of the object. 
Robust planning (and control) for frictional interaction is challenging due to the  hybrid nature of underlying frictional dynamics. Consequently, a lot of classical robust planning and control techniques are not applicable to these systems in the presence of  uncertainties~\cite{drnach2021robust,9812069,yuki2021chance}. While concepts of stability margin or Lyapunov stability have been well studied in the context of nonlinear dynamical system controller design~\cite{vidyasagar2002nonlinear}, such notions have not been explored in contact-rich manipulation problems. This can be mostly attributed to the fact that a controller has to reason about the mechanical stability constraints of the frictional interaction to ensure stability. Mechanical stability closely depends on the contact configuration during manipulation, and thus a planner (or controller) has to ensure that the desired contact configuration is either maintained during the task or it can maintain stability even if the contact sequence is perturbed. Analysis of such systems is difficult in the presence of friction as it leads to differential inclusion system (see~\cite{raghunathan2020stability}) . One of the key insights we present in this paper is that friction provides mechanical stability margin during a contact-rich task. We call the mechanical stability provided by friction as \textit{Frictional Stability}. This \textit{frictional stability} can be exploited during optimization to allow stability of manipulation in the presence of uncertainty. We show the effect of several different parameters on the stability of the manipulation using the proposed approach. In particular, we consider the effect of contact modes and point of contact between the robot \& object on the stability of the manipulation. %We believe that our proposed ideas could also be used for designing feedback controllers to correct contact trajectories based on estimates of contact states.

We study pivoting manipulation where the object being manipulated has to maintain slipping contact with two external surfaces (see \fig{fig:mechanics_pivoting_eq}). A robot can use this manipulation to reorient parts on a planar surface to allow grasping or assist in assembly by manipulating objects to a desired pose (see \fig{fig:pivoting_abstractfig}). Note that this manipulation is challenging as it requires controlled slipping (as opposed to sticking contact~\cite{hou2018fast, hogan2020tactile, shirai2023tactile}). Ensuring robustness for slipping contact is challenging due to the equality constraints for the contact forces compared to inequality constraints for sticking contact. To ensure mechanical stability of the two-point pivoting in the presence of uncertainty, we derive a sufficient condition for stability which allows us to compute a margin of stability. This margin is then used in a bilevel optimization technique, CIBO (Contact Implicit Bilevel Optimization).
\revise{Our proposed CIBO designs an optimal control trajectory while maximizing the worst-case margin along the entire trajectory for manipulation.}
% , to design an optimal control trajectory while maximizing this margin. 
Through numerical simulations as well as physical experiments, we verify that CIBO is able to achieve more robustness compared to the basic trajectory optimization.

% What Yuki wants to describe in this section: 
% \begin{enumerate}
% \item contacts are important
% \item ideally we would like to do manipulation such as pivoting for unknown objects
% \item it is challenging because it is not clear how to formulate the planning/control problem under uncertainty. 
% \item also, slipping control is challenging since it can switch to different mode.
% \item Thus, in this paper, recognizing that friction forces can account for uncertainty, we can generate open-loop trajectories
% \end{enumerate}

% We use vision-based tactile sensors co-located at the fingers of a parallel jaw gripper to detect stability of an object in an unknown grasp. In Figure~\ref{fig:mechanics_placement}, we show a sticking contact configuration between an object and its external environment during a placement attempt. The Figure~\ref{fig:mechanics_placement} shows the force and moments experienced by the object in the shown contact configuration. It can be seen that the object will undergo rotation about the axis of grasp, when the object is pushed against the external surface. 

\textbf{Contributions.} This paper has the following contributions.
\begin{enumerate}
    \item We present analysis of mechanical stability of pivoting manipulation with uncertainty in mass, CoM location, \revise{contact location,} and coefficient of friction.
    \item We present a robust contact-implicit bilevel optimization (CIBO) technique which can be used to optimize the mechanical stability margin to compute robust trajectories for pivoting manipulation. For objects with non-convex shapes, we present a formulation with mode-based optimization. 
     %during packing and assembly-like scenarios.
\end{enumerate}
The proposed method is demonstrated for reorienting parts using a 6 DoF manipulator (see \fig{fig:pivoting_abstractfig}. Please see a video demonstrating hardware experiments at this link\footnote{\url{https://www.youtube.com/watch?v=ojlZDaGytSY}}).
A preliminary version of this work was initially presented at a conference~\cite{9811812}. However, compared to the previous work, this paper has the following major differences:
\begin{enumerate}
    % \item We present a generic formulation for stability during the pivoting manipulation task.
    \item We present analysis of the proposed manipulation considering patch contact, \revise{uncertain mass on a slope, robot finger contact location}, and stochastic friction coefficients at the different points of contact.
    \item \revise{We present a mode-based optimization formulation which can be used for computing robust trajectories for objects with non-convex geometry.}
    \item \revise{We also implement a closed-loop controller with vision feedback which operates in an MPC fashion where we use CIBO for re-computation of controller up on state feedback. We show that we are able to achieve additional robustness for the closed-loop controller.}
    % \item We present formulation and verification for recovery from failure during the proposed pivoting manipulation.
\end{enumerate}

In Section~\ref{sec:related_work}, we present work which is relevant to our proposed work. In Section~\ref{sec:mechanics}, we present the mechanics of pivoting manipulation. Section~\ref{sec:sec_formulation} presents an analysis of frictional stability margin considering different sources of uncertainty. In Section~\ref{sec:robust_to}, we present the proposed contact-implicit bilevel optimization (CIBO) for robust pivoting manipulation. Section~\ref{sec:results} presents numerical results of trajectory optimization as well as experimental evaluation using a manipulator arm and several different objects. Finally, the paper is concluded in Section~\ref{sec:discussion} with some topics for future research.
% We believe that this is the first attempt in literature to analyze and propose a solution to the problem of detecting stability of an object depending on the contact formation between the object and environment. We believe that the proposed work would be useful for several manipulation problems like stable feedback pivoting, part re-orientation during assembly, etc.
\section{Related Work}\label{sec:related_work}

%\textcolor{red}{shouldn't we add more papers?}

% In this section, we present some literature which is closely related to the work we present in this paper. 
Contact modeling has been extensively studied in mechanics as well as robotics literature~\cite{todorov2010implicit, drumwright2011evaluation, 9366782, shirai2022iros, 9561521, 9739950}, \revise{\cite{Yoshida_Regrasp2009, Dafle_gripper2015}}. One of the most common contact models is based on the linear complementarity problem (LCP). LCP-based contact models have been extensively used for performing trajectory optimization in manipulation~\cite{9812069,jin2021trajectory} as well as locomotion~\cite{posa2014direct, 8648229}. More recently, there has also been some work for designing robust manipulation techniques for contact-rich systems using stochastic optimization~\cite{yuki2021chance,drnach2021robust, 9812069, shirai2023covariance}. These problems consider stochastic complementarity systems and consider robust optimization for the underlying stochastic system. However, these problems consider a dynamical model and do not explicitly consider the mechanical stability during planning. Our work is motivated by the concepts of stability under multiple contacts in legged locomotion. \revise{Quasi-static} stability with multiple contacts has been widely studied in legged locomotion~\cite{4598894, 8383993, 8416785, 8358969, 9113247}. These works consider the problem of mechanical stability of the legged robot under multiple contacts by considering the stability polygon defined by the frictional contacts. 
The planning framework for optimizing contact wrench cone margin during locomotion is able to achieve robust locomotion results \cite{dai2016planning, 8358969, 8289420}.
Similar to the concept of these works, we present the idea of frictional stability which defines the extent to which multiple points of contact can compensate for unknown forces and moments in the presence of uncertainty in the mass, CoM location, \revise{contact location,} and frictional parameters. This idea exploits contact forces to ensure stability of the object during the two-point pivoting.  % of the object being manipulated. 
Our work is also related to manipulation by shared grasping~\cite{hou2020manipulation} which discusses mechanics of shared grasping and shows impressive demonstrations. In contrast to the work presented in~\cite{hou2020manipulation}, we present a robust contact-implicit bilevel optimization (CIBO) framework that can be used to find feasible solutions in the presence of uncertainty during the pivoting manipulation and avoids consideration of different modes during planning.

In~\cite{hogan2020tactile}, authors consider stabilization of a table-top manipulation task during online control. They consider a decomposition of the control task in object state control and contact state control. The contact state was detected using vision-based tactile sensors~\cite{donlon2018gelslim,li2020f}. As the task mostly required sticking contact for stability, the tactile feedback was designed to make corrections to push the system away from the boundary of friction cone at the different contact locations. However, the authors did not consider the problem of designing trajectories which can provide robustness to uncertainty. Furthermore, the authors only considered controlled sticking in~\cite{hogan2020tactile} which is, in general, easier than controlled slipping.  
\revise{
Similarly, in \cite{DanicaIn-hand2015}, authors design and validate their sliding controller for in-hand tool pivoting. In \cite{DanicaIn-hand2016}, the authors extend their sliding controller in \cite{DanicaIn-hand2015} such that the sliding controller is able to achieve adaptive control for friction coefficients using visual and force measurements, showing impressive demonstrations. 
Also, authors in \cite{Cruciani_inhand2017} consider pivoting manipulation with a parallel gripper without relying on fast and precise robotic systems.
In contrast to their work in \cite{DanicaIn-hand2015, DanicaIn-hand2016, Cruciani_inhand2017}, we present the pivoting manipulation with extrinsic contacts, which introduces additional complexity of the manipulation, and other uncertain parameters such as mass, CoM location, and robot contact location. 
The work in \cite{Dafle_Extrinsic2014} discusses dexterous in-hand manipulation including extrinsic contact.
% Also, authors in \cite{Cruciani_inhand2017} 
However, the work in \cite{Dafle_Extrinsic2014} does not consider uncertainty in physical parameters.
}
Other previous works that study stable pivoting also consider sticking contact during pivoting using multiple points of contact~\cite{hou2018fast}. The problem in~\cite{hou2018fast} is inherently stable as the object is always in stable grasp. Furthermore, the authors do not consider any uncertainty during planning. Similarly, authors in~\cite{aceituno2020global} present a mixed integer programming formulation to generate contact trajectory given a desired reference trajectory for the object for several manipulation primitives. 
In contrast, this work proposes a bilevel optimization technique which maximizes the minimum margin from instability that the object experiences during an entire trajectory.
Another related work is presented in~\cite{han2020local} where the authors study the feedback control during manipulation of a half-cylinder. The idea there is to design a reference trajectory and then use a local controller by building a funnel around the reference trajectory by linearizing the dynamics. The online control is computed by solving linear programs to locally track the reference trajectory.  

From the above discussion, we can arrive at the following conclusion. In contrast to most of the related work, this proposed work presents a novel formulation for two-point pivoting which requires slipping contact formation between the object and the environment. Furthermore, in comparison to most of the work on contact implicit trajectory optimization, we present a contact implicit bilevel optimization (CIBO) for robust trajectory optimization for manipulation. Even though this method is illustrated on a particular pivoting manipulation problem in this paper, the proposed optimization algorithm could be used for other robust manipulation problems based on the mechanics of the manipulation task.

% \revise{A two-phase gripper to reorient and
% grasp. Propose the design of the gripper. Maybe just cite. }

% \revise{Extrinsic Dexterity: In-Hand Manipulation with External Forces. Using extrinsic dexterity, the robot can still do very cool motion. No discussion in robustness. Extrinsic contact }

% \revise{In-hand manipulation using
% three-stages open loop pivoting. Use q-learning. But in open loop. no discussion on robustness, extrinsic contact. }

% Vision-based tactile sensors have received a lot of attention in the recent years~\cite{donlon2018gelslim,li2020f}. These vision-based tactile sensors can observe very high resolution contact-patch formed at these sensors when an object is grasped at the fingers by a gripper. These high resolution images can then be used for several different tasks. These sensors have been used to perform various estimation tasks using these visuo-tactile sensors. For example, incipient slip detection for grasped objects is a basic requirement for 

\section{Mechanics of Pivoting}\label{sec:mechanics}
% What We need to say:
% 1. Stability margin Concept. Given fx, fy (u) and length, parameters, we can have safety margin so that we can discuss stability
% exact formulation of safety margin in formulation
% during manipulation, you may not be able to get the precise parameters. Then your manipulation task can fail.
% however, in practice, we don't need to have so much accurate parameters. of course, we can do it better with better estimated parameters but we have uncertainty during the manipulation since inherently friction already compensate for uncertainty as long as it satisfies static equilibrium of force moment and friction cones. For example accordingly friction magnitude can change to some extent. This would not happen to other example. This is due to the property of friction forces. our prime goal in this paper is to somehow utilize this frictional feature so that we can consider the most robust nominal trajectory

\begin{figure}
    \centering    \includegraphics[width=0.4\textwidth]{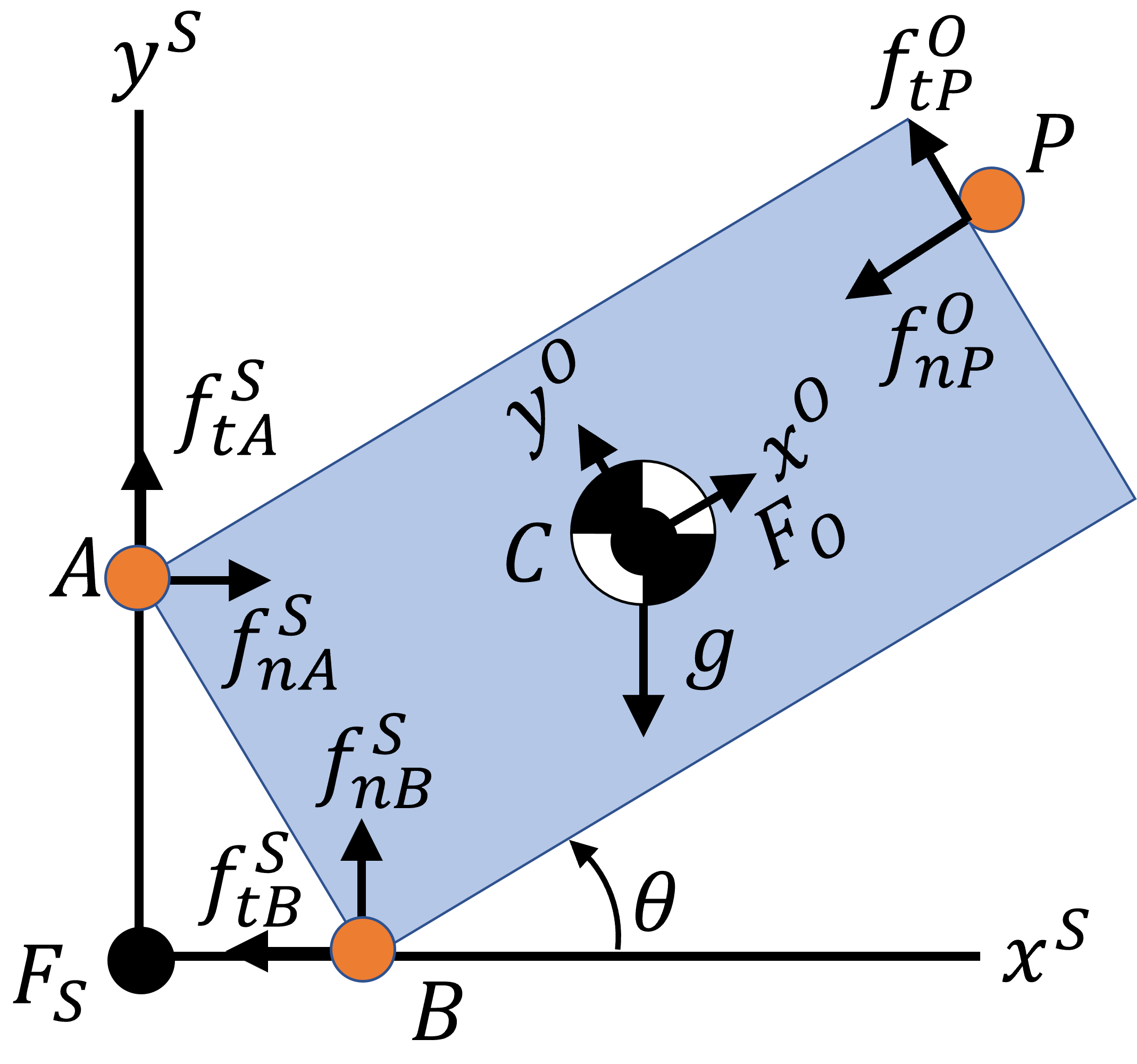} 
    \caption{A schematic showing the free-body diagram of a rigid body during pivoting manipulation \revise{when the relative angle between $F_W$ and $F_S$ is zero.} Point $P$ is the contact point with a manipulator.
    \revise{ The black circle represents the origin of each frame. 
    The object experiences four forces corresponding to two friction forces  from external contact points $A$ and $B$, one control input $f_P$ from the manipulator at point $P$, and gravity at point $C$.
    }
    }
    \label{fig:mechanics_pivoting_eq}
\end{figure}

\begin{figure}
    \centering    \includegraphics[width=0.4\textwidth]{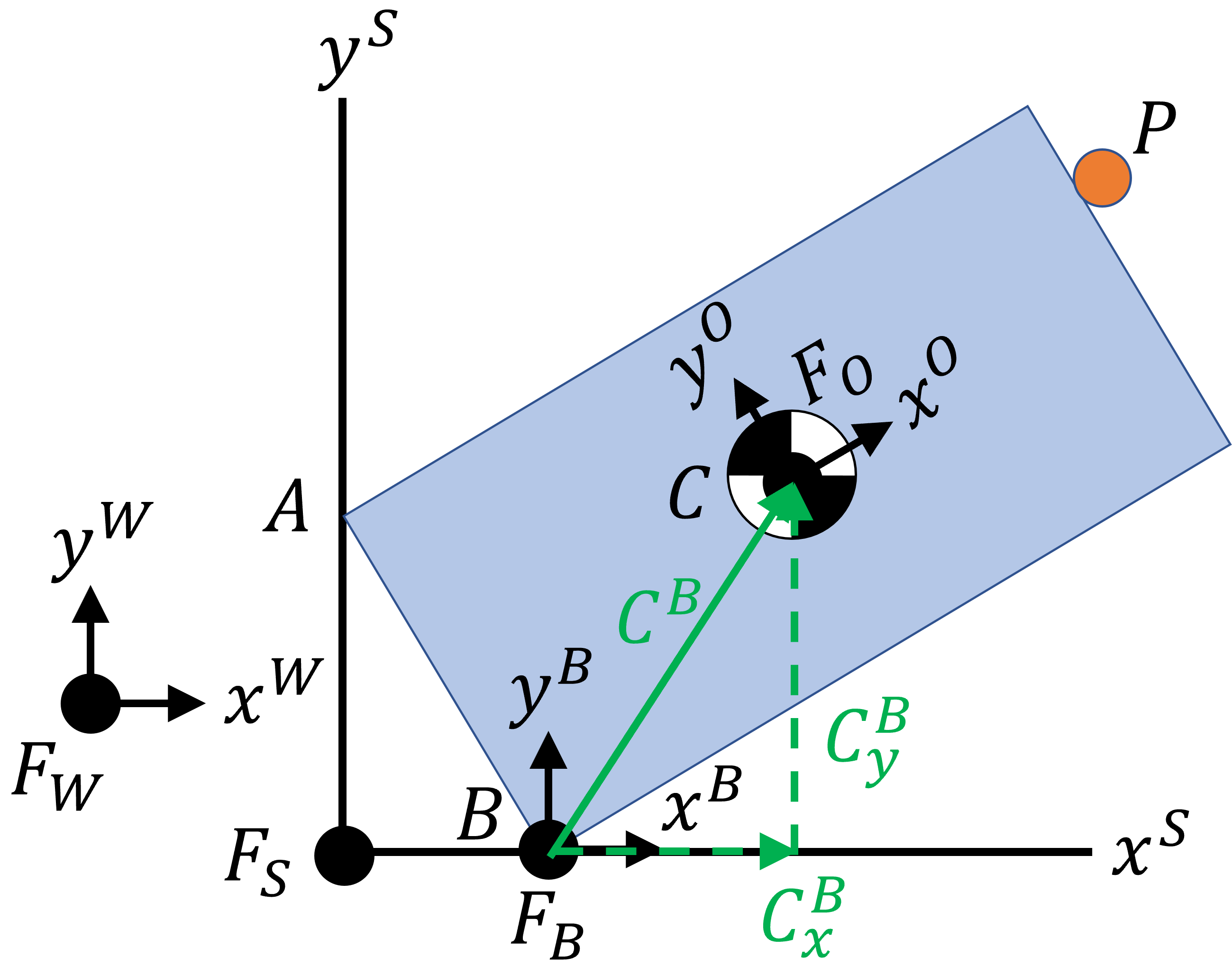} 
    \caption{\revise{A schematic showing the frame definition of a rigid body during pivoting manipulation. $F_W$, $F_S$, $F_O$, and $F_B$ are the world frame, slope frame, object frame, and frame at contact location $B$, respectively. 
    Gravity is defined in $F_W$ where the gravity is parallel to $y$-axis of $F_W$. 
    Pivoting manipulation happens with extrinsic contact $A$ and $B$ defined in $F_S$. $F_O$ is fixed with CoM of an object. $F_B$ is in parallel to $F_S$ with offset $B^S_x$ along $x$-axis of $F_S$.    We also show an example of $i_x^\Sigma$ and $i_x^\Sigma$ in \tab{tab:notion}. In this example,  $C_x^B$ and $C_y^B$ are illustrated.}
    }
    \label{fig:frame_def}
\end{figure}

% \begin{figure}
%     \centering    \includegraphics[width=0.45\textwidth]{Figures/slope_box.png} 
%     \caption{\textcolor{blue}{
%     A schematic showing the free-body diagram of a rigid body on an inclined surface during pivoting manipulation. }}
%     \label{fig:mechanics_pivoting_eq_slope}
% \end{figure}

In this section, we explain quasi-static stability of two-point pivoting in a plane. %We use this to present our proposed concept of \textit{frictional stability} which explains how friction can compensate for the inaccuracy of physical parameters during pivoting.
Before explaining the details, we present our assumptions in this work. The following assumptions are used in the model for the pivoting manipulation task presented in this paper:
\begin{enumerate}
\item The object is rigid.
\item We consider \revise{quasi-static} equilibrium of the object.
\item The external contact surfaces are perfectly flat. 
\item The dimensions and pose of the object is perfectly known.
% \item The frictional parameters for the contact between the object and manipulator are perfectly known.
\item The object makes point contacts.
\end{enumerate}
% The above assumptions are common in manipulation problems. 
% Regarding the fifth assumption, we should mention that other parameters such as coefficient of friction can be also uncertain. However, uncertainty in coefficient of friction would lead to stochastic complementarity system, which is left as a future work~\cite{yuki2021chance}. 

\subsection{Mechanics of Pivoting with External Contacts}
\begin{table}[]
    \centering
        \caption{\revise{Notation of variables for analysis of frictional stability margin. In $\Sigma$ column, we indicate the frame of variables. We use the following indices for defining variables in this table: $j \in \{A, B, C, P\} $ for representing the location of frames, $i \in \{A, B, P\} $ for representing contact location, and $\Sigma \in \{W, S, O, B\} $ for representing a frame. }
        }

\begin{tabular}{|c|c|c|c|}
\hline Name & Description & Size & $\Sigma$ \\
\hline
% $j$ & indices for location. $j \in \{A, B, C, P\} $ &   &  \\
 % $i$ & indices for contact location.  $i \in \{A, B, P\} $ &   &  \\
 % % $q$ &  $q \in \{A, B\} $ &   &  \\
 % $\Sigma$ & indices for frame.  $\Sigma \in \{W, S, O, B\} $ &   &  \\
 $F_\Sigma$ & $\Sigma$ frame.   &   &  \\
  $f_{nj}^\Sigma$ & normal force at  $j$ in frame $F_\Sigma$& $\mathbb{R}^1$  & $\Sigma$ \\
    $f_{tj}^\Sigma$ & friction force at  $j$ in frame $F_\Sigma$& $\mathbb{R}^1$  & $\Sigma$ \\
 % $f_{nq}$ & normal force at contact $q$ & $\mathbb{R}^1$  & $S$ \\
 % $f_{tq}$ & friction force at contact $q$ &  
 % $\mathbb{R}^1$  & $S$ \\
 %  $f_{np}$ & normal force at contact $P$ & $\mathbb{R}^1$  & $O$ \\
 % $f_{tp}$ & friction force at contact $P$ & $\mathbb{R}^1$  & $O$ \\
   $f_{xj}^\Sigma$ & force at $j$ along $x$-axis in frame $F_\Sigma$  & $\mathbb{R}^1$  & $\Sigma$ \\
   $f_{yj}^\Sigma$ & force  at $j$ along $y$-axis in frame $F_\Sigma$  & $\mathbb{R}^1$  & $\Sigma$ \\
 $m$ & mass & $\mathbb{R}^1$  &  \\
 $g$ & gravity acceleration & $\mathbb{R}^1$  &  $W$ \\
  $l$ & length of an object & $\mathbb{R}^1$  &   \\
    $w$ & width of an object & $\mathbb{R}^1$  &   \\
  $\mu_i$ & coefficient of friction at $i$ & $\mathbb{R}^1$  & \\
  $i_x^\Sigma$ & contact location at $i$ along $x$-axis in frame $F_\Sigma$ & $\mathbb{R}^1$  & $\Sigma$ \\
    $i_y^\Sigma$ & contact location at $i$ along $y$-axis in frame $F_\Sigma$ & $\mathbb{R}^1$  & $\Sigma$ \\
      $\dot{i}_x^\Sigma$ & slipping velocity at $i$ along $x$-axis in frame $F_\Sigma$ & $\mathbb{R}^1$  & $\Sigma$ \\
      $\dot{i}_y^\Sigma$ & slipping velocity at $i$ along $y$-axis in frame $F_\Sigma$ & $\mathbb{R}^1$  & $\Sigma$ \\
     $\theta$ & angle of an object & $\mathbb{R}^1$  & $S$ \\
          $\phi$ & relative angle of frame from $\{F_W\}$ to $\{F_S\}$ & $\mathbb{R}^1$  & $W$ \\
          % $p_y^O$ & finger location in frame $O$ & $\mathbb{R}^1$  & $O$ \\
          %           $\dot{p}_y^O$ & finger slipping velocity in frame $O$ & $\mathbb{R}^1$  & $O$ \\
\hline
\end{tabular}
    \label{tab:notion}
\end{table}

%Before describing frictional stability, we describe our problem setting. 
We consider pivoting where the object maintains slipping contact with two external surfaces (see \fig{fig:mechanics_pivoting_eq}). A free body diagram showing the \revise{quasi-static} equilibrium of the object is shown in \fig{fig:mechanics_pivoting_eq}. 
\revise{The definitions of frames and  variables are summarized in 
\fig{fig:frame_def} and \tab{tab:notion}, respectively.}
% 
% The object experiences four forces corresponding to two friction forces $f_A, f_B$ from external contact points $A$ and $B$, one control input $f_P$ from manipulator at point $P$, and gravity, $mg$ at point $C$ where $m$ is mass of a body.
% We denote $f_{ni}, f_{ti}$ as a normal force and friction force at point $\forall i, i=\{A, B\}$, respectively, defined in ${\{F_W\}}$. $f_{nP}, f_{tP}$ are normal and friction force at point $P$ defined in ${\{F_B\}}$. Note that we define the $[f_x, f_y]^\top = \mathbf{R} [f_{nP}, f_{tP}]^\top$ where $\mathbf{R}$ is a rotation matrix from ${\{F_B\}}$ to ${\{F_W\}}$. We denote $x, y$ position at point in ${\{F_W\}}$ $\forall i, i=\{A, B, C, P\}$ as $i_x, i_y$, respectively. We denote $y$ position of point $P$ in ${\{F_B\}}$ as $p_y \in [-\frac{w}{2}, \frac{w}{2}]$.
% % 
% We define the angle of body with respect to $x$-axis as $\theta$. The coefficient of friction at point $\forall i, i=\{A, B, P\}$ are $\mu_A, \mu_B, \mu_P$, respectively. 
In the later sections, we present trajectory optimization formulation where we consider
\revise{
decision variables at time step $k$ (e.g., $f_{k, ni}$). In this section, we remove $k$ to represent variables for simplicity.}
% In this section, we remove $k$ to represent variables for simplicity. }
% $f_{ni}, f_{ti}$, location variables $i_{x}, i_{y}$ $\forall i, i=\{A, B, C, P\}$, $\theta$, and $p_y$ at each time-step $k$ denoted as $f_{k, ni}, f_{k, ti}, i_{k, x}, i_{k, y}, \theta_k, p_{y, k}$. 

% By setting $B_x = B_y = 0$, the static equilibrium of force in $x$ and $y$ directions the static equilibrium of and moment along point $B$ can be given by:
The \revise{quasi-static} equilibrium conditions for the object \revise{in $F_B$ when the relative angle between  $F_W$ and $F_S$ is zero (see \fig{fig:mechanics_pivoting_eq})} can be represented by the following equations.
% Note that we consider the moment at point $B$ by setting $B_x = B_y = 0$:
% (note we consider the moment at point $B$ by setting $B_x = B_y = 0$):
\begin{subequations}
\begin{flalign}
 f_{nA}^{\revise{B}} + f_{tB}^{\revise{B}} + f_{xP}^{\revise{B}}   =0,\label{forceeq1}\\
f_{tA}^{\revise{B}} + f_{nB}^{\revise{B}} + mg + f_{yP}^{\revise{B}}   = 0,  \label{forceeq2}\\
A_x^{\revise{B}} f_{tA}^{\revise{B}} - A_y^{\revise{B}}f_{nA}^{\revise{B}} + C_x^{\revise{B}}mg + P_x^{\revise{B}}f_{yP}^{\revise{B}} - P_y^{\revise{B}}f_{xP}^{\revise{B}} = 0 \label{moment_eq1}
% -\frac{l_\text{com}}{2}c_{\theta - \gamma}f_{tA} + \frac{l_\text{com}}{2}s_{\theta - \gamma}f_{nA} -\frac{l_\text{com}}{2}c_{\theta + \gamma}f_{nb} +\frac{l_\text{com}}{2}s_{\theta + \gamma}f_{tb}
% (A_x-B_x)f_{tA} - (A_y-B_y)f_{nA} + (C_x-B_x)mg + (P_x-B_x)f_{y} - (P_y - B_y) f_x = 0
\end{flalign}
\label{force_eq}
\end{subequations}
\revise{Note that because we define $F_B$ as parallel to $F_S$, all force variables in $F_B$ and $F_S$ are the same. }
We consider Coulomb friction law which results in friction cone constraints as follows:
\begin{equation}
 |f_{tA}^{\revise{B}}|  \leq \mu_A f_{nA}^{\revise{B}}, |f_{tB}^{\revise{B}}|  \leq \mu_B f_{nB}^{\revise{B}}, \quad f_{nA}^{\revise{B}}, f_{nB}^{\revise{B}} \geq 0,
% f_{tA} + f_{nB} + mg + f_{yP}   = 0,  \label{forceeq2}\\
% A_xf_{tA} - A_yf_{nA} + C_xmg + P_xf_{y} - P_y f_x = 0
% -\frac{l_\text{com}}{2}c_{\theta - \gamma}f_{tA} + \frac{l_\text{com}}{2}s_{\theta - \gamma}f_{nA} -\frac{l_\text{com}}{2}c_{\theta + \gamma}f_{nb} +\frac{l_\text{com}}{2}s_{\theta + \gamma}f_{tb}
% (A_x-B_x)f_{tA} - (A_y-B_y)f_{nA} + (C_x-B_x)mg + (P_x-B_x)f_{y} - (P_y - B_y) f_x = 0
\label{general_FC}
\end{equation}
To describe sticking-slipping complementarity constraints, we have the following complementarity constraints at point $A, B$:
\begin{subequations}
\begin{flalign}
 0 \leq  \revise{\dot{A}_{y+}^B} \perp \mu_A f_{nA}^{\revise{B}}-f_{tA}^{\revise{B}} \geq 0,  \\
 0 \leq   \revise{\dot{A}_{y-}^B} \perp \mu_A  f_{nA}^{\revise{B}}+f_{tA}^{\revise{B}} \geq 0, \\
 0 \leq  \revise{\dot{B}_{x+}^B} \perp \mu_B f_{nB}^{\revise{B}}-f_{tB}^{\revise{B}} \geq 0,  \\
 0 \leq   \revise{\dot{B}_{x-}^B} \perp \mu_B  f_{nB}^{\revise{B}}+f_{tB}^{\revise{B}} \geq 0
 \end{flalign}
 \label{slippingAB}
\end{subequations}
where the slipping velocities  follows \revise{ $\dot{A}_y^B=\dot{A}_{y+}^B-\dot{A}_{y-}^B, \dot{B}_x^B=\dot{B}_{x+}^B-\dot{B}_{x-}^B$}.
$\dot{A}_{y+}^B, \dot{A}_{y-}^B$ represent the slipping velocity \revise{at $A$} along positive and negative directions for \revise{$y$-axis in $F_B$}, respectively.
\revise{$\dot{B}_{x+}^B, \dot{B}_{x-}^B$ represent the slipping velocity at $B$ along positive and negative directions for $x$-axis in $F_B$}, respectively.
The notation $0 \leq a \perp b \geq 0$ means the complementarity constraints $a \geq 0, b \geq 0, a b=0$.
Since we consider slipping contact during pivoting, we have "equality" constraints in friction cone constraints at points $A, B$:
\begin{equation}
 f_{tA}^{\revise{B}}  =\mu_A f_{nA}^{\revise{B}}, f_{tB}^{\revise{B}}  =-\mu_B f_{nB}^{\revise{B}}
% f_{tA} + f_{nB} + mg + f_{yP}   = 0,  \label{forceeq2}\\
% A_xf_{tA} - A_yf_{nA} + C_xmg + P_xf_{y} - P_y f_x = 0
% -\frac{l_\text{com}}{2}c_{\theta - \gamma}f_{tA} + \frac{l_\text{com}}{2}s_{\theta - \gamma}f_{nA} -\frac{l_\text{com}}{2}c_{\theta + \gamma}f_{nb} +\frac{l_\text{com}}{2}s_{\theta + \gamma}f_{tb}
% (A_x-B_x)f_{tA} - (A_y-B_y)f_{nA} + (C_x-B_x)mg + (P_x-B_x)f_{y} - (P_y - B_y) f_x = 0
\label{slipping_friction_cone}
\end{equation}
To realize stable pivoting, actively controlling position of point $P$ is important. Thus, we consider the following complementarity constraints that represent the relation between the slipping velocity \revise{$\dot{P}_y$}  at point $P$ in \revise{$F_O$} and friction cone constraint at point $P$:
\begin{subequations}
\begin{flalign}
 0 \leq   \dot{P}_{y+}^{\revise{O}} \perp \mu_p f_{nP}^{\revise{O}}-f_{tP}^{\revise{O}} \geq 0  \\
 0 \leq   \dot{P}_{y-}^{\revise{O}} \perp \mu_p  f_{nP}^{\revise{O}}+f_{tP}^{\revise{O}} \geq 0 
 \end{flalign}
 \label{slippingP}
\end{subequations}
where \revise{$\dot{P}_y^O=\dot{P}_{y+}^O-\dot{P}_{y-}^O$}.

\section{Robust Pivoting Formulation}\label{sec:sec_formulation}
% \subsection{Robust Pivoting Formulation}\label{subsec:robust_pivoting} 
In this section, we present a generic formulation for robust pivoting manipulation. In particular, we use the \revise{quasi-static} equilibrium conditions~\eqref{force_eq} in the presence of disturbances to formulate the robust planning problem. In particular, using sufficiency for stability of the object during manipulation we can estimate the bound of disturbance that can be tolerated during manipulation. Since this bound would depend on the pose of the object, we reason about the margin throughout the manipulation trajectory during the optimization problem formulation. We present the general idea in the following paragraph.

In the most general case, we assume that there is an external force $F_{ext}^{\revise{B}}$ and moment $M_{ext}^{\revise{B}}$ acting on the object during manipulation. Let us assume that the $x$ and $y$ component of the external force \revise{in $F_B$} are represented as $F_{ext,x}^{\revise{B}}$ and $F_{ext,y}^{\revise{B}}$ respectively. Then the \revise{quasi-static} equilibrium conditions~\eqref{force_eq} can be rewritten as follows:
\begin{subequations}
\begin{flalign}
 f_{nA}^{\revise{B}} + f_{tB}^{\revise{B}} + f_{xP}^{\revise{B}}+F_{ext,x}^{\revise{B}}  =0,\label{robust_forceeq1}\\
f_{tA}^{\revise{B}} + f_{nB}^{\revise{B}} + mg + f_{yP}^{\revise{B}}+F_{ext,y}^{\revise{B}}   = 0,  \label{robust_forceeq2}\\
A_x^{\revise{B}}f_{tA}^{\revise{B}} - A_y^{\revise{B}}f_{nA}^{\revise{B}} + C_x^{\revise{B}}mg + P_x^{\revise{B}}f_{yP}^{\revise{B}} 
- P_y^{\revise{B}} f_{xP}^{\revise{B}}  \nonumber \\+M_{ext}^{\revise{B}} = 0 
\label{robust_moment_eq1}
% -\frac{l_\text{com}}{2}c_{\theta - \gamma}f_{tA} + \frac{l_\text{com}}{2}s_{\theta - \gamma}f_{nA} -\frac{l_\text{com}}{2}c_{\theta + \gamma}f_{nb} +\frac{l_\text{com}}{2}s_{\theta + \gamma}f_{tb}
% (A_x-B_x)f_{tA} - (A_y-B_y)f_{nA} + (C_x-B_x)mg + (P_x-B_x)f_{y} - (P_y - B_y) f_x = 0
\end{flalign}
\label{robust_force_eq}
\end{subequations}
Note that $F_{ext}^{\revise{B}}$ and $M_{ext}^{\revise{B}}$ may not be independent of each other. They are related via the the point of application of force $F_{ext}^{\revise{B}}$ in the \revise{quasi-static} equilibrium conditions~\eqref{robust_force_eq}. These equations may not be satisfied for all possible values of $F_{ext}^{\revise{B}}$ and $M_{ext}^{\revise{B}}$. Since the contact forces can be readjusted in~\eqref{robust_force_eq}, the \revise{quasi-static} equilibrium can be satisfied for a certain range of $F_{ext}^{\revise{B}}$ and $M_{ext}^{\revise{B}}$. A generic analysis for estimating this margin or bound for which these disturbances can be compensated by contact forces is a bit involved as such a bound is dependent on the point and angle of application of the external force $F_{ext}^{\revise{B}}$. In the following sections, we present some specific cases which can be analyzed by making some simplifying assumptions on these disturbances. 
\revise{For brevity, we omit superscript $B$ of variables in the following sections because we consider \revise{quasi-static} equilibrium in $F_B$ unless we consider \revise{quasi-static} equilibrium in a different frame (see Sec~\ref{sec:stability_margin_finger_contact_location}). }
% \eq{slippingP} means that $\dot{p}_y$ is non-zero only at the boundary of friction cone.

% Additionally, one can have the following complementarity constraints to discuss the case where the object can lose contact at point B and can incur rotation about point A: 
% \begin{equation}
%  0 \leq   B_y \perp   f_{nB} \geq 0 
%  \label{complementarity_in_air}
% \end{equation}
% f_{tA} + f_{nB} + mg + f_{yP}   = 0,  \label{forceeq2}\\
% A_xf_{tA} - A_yf_{nA} + C_xmg + P_xf_{y} - P_y f_x = 0
% -\frac{l_\text{com}}{2}c_{\theta - \gamma}f_{tA} + \frac{l_\text{com}}{2}s_{\theta - \gamma}f_{nA} -\frac{l_\text{com}}{2}c_{\theta + \gamma}f_{nb} +\frac{l_\text{com}}{2}s_{\theta + \gamma}f_{tb}
% (A_x-B_x)f_{tA} - (A_y-B_y)f_{nA} + (C_x-B_x)mg + (P_x-B_x)f_{y} - (P_y - B_y) f_x = 0

% In this work, we do not consider \eq{complementarity_in_air} during optimization since \eq{complementarity_in_air} leads to stochastic complementarity system.

\subsection{Frictional Stability Margin}

%Otherwise, the manipulation may not be able to be completed due to uncertain parameters. However, in reality, uncertainty always exists and the manipulation can be completed under uncertainty to some extent. This is because friction forces inherently compensate for uncertainty as long as they satisfy static equilibrium of force and moments and friction cone constraints. the body can lift up by sticking at point $A$.  Hence, in this case, we can imagine that 

% In model-based manipulation, it is important to have precise estimate of physical parameters. However, it is desirable that a robot can compensate for uncertainty during manipulation of novel objects. 
% In this paper, we provide some insights about how a robot can use the stability margin in static equilibrium during manipulation to compensate for uncertainty in gravitational forces and moments. \fig{fig:concept} shows our proposed concept of frictional stability and bilevel robust trajectory optimization. The latter is discussed in Sec~\ref{sec:robust_to}. 
The robust quasi-static equilibrium conditions shown in~\eqref{robust_force_eq} can be used to explain the concept of stability margin. The stability margin is given by the magnitude of the external force $F_{ext}^{\revise{B}}$ and moment $M_{ext}^{\revise{B}}$ which can be satisfied in~\eqref{robust_force_eq} in any stable configuration of the object. This margin would depend on the contact force between the object and the environment as well as the control force used by the manipulator during the task. This provides the intuition that one can design a control trajectory such that the stability margin can be maximized.

We briefly provide some physical intuition about frictional stability for a few specific cases. First suppose that uncertainty exists in mass of a body. In the case when the actual mass is lower than  estimated, the friction force at point $A$ would increase while the friction force at point $B$ would decrease, compared to the nominal case. In contrast, suppose if the actual mass of the body is heavier than that of what we estimate, then the body can tumble along point $B$ in the clockwise direction. In this case, we can imagine that the friction force at point $A$ would decrease while the friction force at point $B$ would increase. However, as long as the friction forces are non-zero, the object can stay in contact with the external environment.
Similar arguments could be made for uncertainty in CoM location. The key point to note that the friction forces can re-distribute at the two contact locations and thus provide a margin of stability to compensate for uncertain gravitational forces and moments. We call this margin as \textit{frictional stability}.

In the following sections, we present the mathematical formulation of \textit{frictional stability} for cases when the mass, CoM location, friction coefficients, or \revise{finger contact location} are not known perfectly.
% In \fig{fig:concept},  if the actual CoM location is more left (represented as light-blue) than that of what we estimate (represented as black), then the body can lift up by sticking at point $A$.  Hence, in this case, we can imagine that the friction force at point $A$ would increase while the friction force at point $B$ would decrease. In contrast, if the actual CoM location is more right, then the body can tumble along point $B$ in the clockwise direction. In this case, we can imagine that the friction force at point $A$ would decrease while the friction force at point $B$ would increase. We could provide a similar insight when the body has uncertain CoM location. 

% In the next few subsections, we formalize the intuition for frictional stability margin proposed above and derive sufficient condition for stability of the object during pivoting.

%ly formulate frictional stability and confirm that what we describe above is actually confirmed mathematically. Here we consider the body with uncertain mass.

\subsection{Stability Margin for Uncertain Mass}\label{sec:sec_uncertain_mass}
 For simplicity, we denote $\epsilon$ as uncertain weight with respect to the estimated weight. Also, to emphasize that we consider the system under uncertainty, we put superscript $\epsilon$ for each friction force variable. Thus, the \revise{quasi-static} equilibrium conditions in \eq{force_eq} can be rewritten as:
\begin{subequations}
\begin{flalign}
f_{nA}^\epsilon + f_{tB}^\epsilon + f_{xP}  =0\label{forceeq11},\\
f_{tA}^\epsilon + f_{nB}^\epsilon + (mg + \epsilon) + f_{yP}   = 0,  \label{forceeq21}\\
A_xf_{tA}^\epsilon - A_yf_{nA}^\epsilon + C_x (mg+\epsilon) + P_xf_{yP}  = P_y f_{xP} \label{moment_eq11}
\end{flalign}
\label{force_eq_mass}
\end{subequations}
% equation with super script B
% \begin{subequations}
% \begin{flalign}
% f_{nA}^\epsilon + f_{tB}^\epsilon + f_{xP}^{\revise{B}}  =0\label{forceeq11},\\
% f_{tA}^\epsilon + f_{nB}^\epsilon + (mg + \epsilon) + f_{yP}^{\revise{B}}   = 0,  \label{forceeq21}\\
% A_x^{\revise{B}}f_{tA}^\epsilon - A_y^{\revise{B}}f_{nA}^\epsilon + C_x^{\revise{B}} (mg+\epsilon) + P_x^{\revise{B}}f_{yP}^{\revise{B}}  = P_y^{\revise{B}} f_{xP} \label{moment_eq11}
% % -\frac{l_\text{com}}{2}c_{\theta - \gamma}f_{tA} + \frac{l_\text{com}}{2}s_{\theta - \gamma}f_{nA} -\frac{l_\text{com}}{2}c_{\theta + \gamma}f_{nb} +\frac{l_\text{com}}{2}s_{\theta + \gamma}f_{tb}
% % (A_x-B_x)f_{tA} - (A_y-B_y)f_{nA} + (C_x-B_x)mg + (P_x-B_x)f_{y} - (P_y - B_y) f_x = 0
% %In order to realize the robustness, the more contact points are preferred since the system obtains more frictional stability. Therefore, we aim at ensuring contacts during pivoting (i.e., slipping). 
% \end{flalign}
% \label{force_eq_mass}
% \end{subequations}
Then, using \eq{slipping_friction_cone} and \eq{moment_eq11}, we obtain:
\begin{equation}
f_{nA}^\epsilon = \frac{-{C_x}\left(mg + \epsilon\right) -{P_x}f_{yP} + {P_y}f_{xP}}{\mu_A {A_x} - {A_y}}
\label{fna_111}
\end{equation}
To ensure that the body maintains contact with the external surfaces, we would like to enforce that the body experience non-zero normal forces at the both contacts.
To realize this, we have $f_{nA}^\epsilon \geq 0, f_{nB}^\epsilon \geq 0$ as conditions that the system needs to satisfy. Consequently, by simplifying \eq{fna_111}, we get the following:
\begin{subequations}
\begin{flalign}
\epsilon \geq \frac{P_yf_{xP} - P_xf_{yP} - C_xmg}{C_x}, \text{ if } C_x>0,  \label{fnacond1}\\
\epsilon \leq \frac{P_yf_{xP} - P_xf_{yP} - C_xmg}{C_x}, \text{ if } C_x<0  \label{fnacond2}
% -\frac{l_\text{com}}{2}c_{\theta - \gamma}f_{tA} + \frac{l_\text{com}}{2}s_{\theta - \gamma}f_{nA} -\frac{l_\text{com}}{2}c_{\theta + \gamma}f_{nb} +\frac{l_\text{com}}{2}s_{\theta + \gamma}f_{tb}
% (A_x-B_x)f_{tA} - (A_y-B_y)f_{nA} + (C_x-B_x)mg + (P_x-B_x)f_{y} - (P_y - B_y) f_x = 0
\end{flalign}
\label{fna_cond_mass}
\end{subequations}
Note that the upper-bound of $\epsilon$ means that the friction forces can exist even when we make the mass of the body lighter up to $\frac{\epsilon}{g}$. The lower-bound of $\epsilon$ means that the friction forces can exist even when we make the mass of the body heavier up to $\frac{\epsilon}{g}$. 
\eq{fna_cond_mass} provides some useful insights. \eq{fna_cond_mass} gives either upper- or lower-bound of $\epsilon$ for $f_{nA}^\epsilon$ according to the sign of $C_x$ (the moment arm of gravity). This is because the uncertain mass would generate an additional moment along with point $B$ in the clock-wise direction if $C_x >0$ and in the counter clock-wise direction if $C_x <0$. 
% In other words, the heavier body can make itself tumble and lose contact at point $A$ so it is not preferred if $C_x>0$, and lighter body can make itself again tumble and lose contact at point $A$ so it is not preferred if $C_x<0$  based on $f_{nA}^\epsilon \geq 0$. 
If $C_x = 0$, we have an unbounded range for $\epsilon$, meaning that the body would not lose contact at point $A$ no matter how much uncertainty exists in the mass. 

\eq{fna_cond_mass} can be reformulated as an inequality constraint: 
\begin{equation}
 C_x(\epsilon - \epsilon_A) \geq 0
% f_{tA} + f_{nB} + mg + f_{yP}   = 0,  \label{forceeq2}\\
% A_xf_{tA} - A_yf_{nA} + C_xmg + P_xf_{y} - P_y f_x = 0
% -\frac{l_\text{com}}{2}c_{\theta - \gamma}f_{tA} + \frac{l_\text{com}}{2}s_{\theta - \gamma}f_{nA} -\frac{l_\text{com}}{2}c_{\theta + \gamma}f_{nb} +\frac{l_\text{com}}{2}s_{\theta + \gamma}f_{tb}
% (A_x-B_x)f_{tA} - (A_y-B_y)f_{nA} + (C_x-B_x)mg + (P_x-B_x)f_{y} - (P_y - B_y) f_x = 0
\label{fna_cond_mass_one}
\end{equation}
where $\epsilon_A = \frac{P_yf_{xP} - P_xf_{yP} - C_xmg}{C_x}$.

We can derive condition for $\epsilon$ based on $f_{nB}^\epsilon \geq 0$ from \eq{slipping_friction_cone}, \eq{forceeq11}, and \eq{forceeq21}:
\begin{equation}
 \epsilon \leq \mu_A f_{xP} -f_{yP} -mg
% f_{tA} + f_{nB} + mg + f_{yP}   = 0,  \label{forceeq2}\\
% A_xf_{tA} - A_yf_{nA} + C_xmg + P_xf_{y} - P_y f_x = 0
% -\frac{l_\text{com}}{2}c_{\theta - \gamma}f_{tA} + \frac{l_\text{com}}{2}s_{\theta - \gamma}f_{nA} -\frac{l_\text{com}}{2}c_{\theta + \gamma}f_{nb} +\frac{l_\text{com}}{2}s_{\theta + \gamma}f_{tb}
% (A_x-B_x)f_{tA} - (A_y-B_y)f_{nA} + (C_x-B_x)mg + (P_x-B_x)f_{y} - (P_y - B_y) f_x = 0
\label{fnb_cond_mass}
\end{equation}
We only have upper-bound on $\epsilon$ based on $f_{nB}^\epsilon \geq 0$, meaning that the contact at point $B$ cannot be guaranteed if the actual mass is lighter than $\mu_A f_{xP} -f_{yP} -mg$. 
%Note that the conditions  we derive in this paper are sufficient but not necessary conditions.

\subsection{Stability Margin for Uncertain CoM Location}\label{sec:stability_margin_com_location}
% We consider the case with uncertain CoM location. We have a similar discussion we have in Sec~\ref{sec_uncertain_mass}. 
We denote $\revise{d_x^{{O}}, d_y^{{O}}}$ as residual CoM locations with respect to the estimated CoM location in $F_{\revise{O}}$ coordinate, respectively. Thus, the residual CoM location in $\revise{F_W}$, $\revise{d_x^{{W}}, d_y^{{W}}}$, are represented by $\revise{d_x^{{W}}} = d \cos({\theta + \theta_d}), \revise{d_y^{{W}}} = d \sin({\theta + \theta_d})$, where $d = \sqrt{\revise{\left({d_x^{{O}}}\right)^2 + \left({d_y^{{O}}}\right)^2}}$, $\theta_d = \arctan{\frac{\revise{{d_y^{{O}}}}}{\revise{{d_x^{{O}}}}}}$.  For notation simplicity, we use $r$ to represent $\revise{d_x^{{W}}}$.  In this paper, we put superscript $r$ for each friction force variable. The \revise{quasi-static} equilibrium conditions in \eq{force_eq} can be rewritten as follows:
\begin{subequations}
\begin{flalign}
f_{nA}^r + f_{tB}^r + f_{xP}  =0\label{forceeq12},\\
f_{tA}^r + f_{nB}^r + mg + f_{yP}   = 0,  \label{forceeq22}\\
A_xf_{tA}^r - A_yf_{nA}^r + (C_x + r) mg + P_xf_{yP}  = P_y f_{xP}  \label{moment_eq12}
\end{flalign}
\label{force_eq_location}
\end{subequations}
Then, using \eq{slipping_friction_cone} in \eq{force_eq_location}, we obtain:
\begin{subequations}
\begin{flalign}
r \leq \frac{P_yf_{xP} -P_x f_{yP}}{mg} - C_x \label{fna_fnb_r1},\\
r \geq  - \frac{\frac{\mu_A A_x - A_y}{1 + \mu_A}(-f_{xP}-f_{yP}-mg) -P_yf_{xP} + P_xf_{yP}}{mg}  -C_x  \label{fna_fnb_r2}
% -\frac{l_\text{com}}{2}c_{\theta - \gamma}f_{tA} + \frac{l_\text{com}}{2}s_{\theta - \gamma}f_{nA} -\frac{l_\text{com}}{2}c_{\theta + \gamma}f_{nb} +\frac{l_\text{com}}{2}s_{\theta + \gamma}f_{tb}
% (A_x-B_x)f_{tA} - (A_y-B_y)f_{nA} + (C_x-B_x)mg + (P_x-B_x)f_{y} - (P_y - B_y) f_x = 0
\end{flalign}
\label{fna_fnb_r}
\end{subequations}
where \eq{fna_fnb_r1}, \eq{fna_fnb_r2} are obtained based on $f_{nA}^r \geq 0, f_{nB}^r \geq 0$, respectively. \eq{fna_fnb_r} means that the object would lose contact at $A$ if the actual CoM location is more to the right than our expected CoM location while the object would lose the contact at $B$ if the actual CoM location is more to the left.

\subsection{Stability Margin for Stochastic Friction}\label{subsec:stochasticfriction_planning}
In this section, we present modeling and analysis of pivoting manipulation in the presence of stochastic friction coefficients. In particular, we consider stochastic friction at the two different contact points $A$ and $B$. We do not consider stochastic friction at the contact point between the robot and the manipulator since that leads to stochastic complementarity constraints (please see~\cite{shirai2023covariance, yuki2021chance} for detailed analysis on stochastic complementarity constraints). 
We make the assumption that the friction coefficients at $A$ and $B$ are partially known. In particular, we assume that the friction coefficients for contact at $A$ could be represented as $\mu_A=\hat{\mu}_A+\revise{\tilde{\mu}_A}$
where $\revise{\tilde{\mu}_A}$ is the uncertain stochastic variable.
% $\revise{\tilde{\mu}_A}\sim\mathcal N(0,\sigma_{\mu_A}^2)$. 
Similarly, the friction coefficient at $B$ could be represented as  $\mu_B=\hat{\mu}_B+\revise{\tilde{\mu}_B}$ where
  $\revise{\tilde{\mu}_B}$ is the uncertain stochastic variable. Note that we do not need to need to know any information regarding the \revise{probabilistic} distribution \revise{(e.g., probability density function of Gaussian distribution, beta distribution.)} of the unknown part. 
% $\revise{\tilde{\mu}_B}\sim \mathcal N(0,\sigma_{\mu_B}^2)$. 
We can rewrite~\eqref{robust_force_eq} for this case as follows. We put superscript $\mu$ for each friction variable:
\begin{subequations}
\begin{flalign}
 f_{nA}^\mu + \hat{f}_{tB}^\mu + f_{xP}+\epsilon_B  =0,\label{robust_forceeq_friction_friction}\\
\hat{f}_{tA}^\mu + f_{nB}^\mu + mg + f_{yP}+\epsilon_A   = 0,  \label{robust_forceeq_friction_friction}\\
A_x\hat{f}_{tA}^\mu+A_x\epsilon_A - A_yf_{nA}^\mu + C_xmg  \nonumber \\+ P_xf_{yP} - P_y f_{xP}=  0\label{robust_moment_eq_friction_friction}
\end{flalign}
\label{robust_force_eq_friction_friction}
\end{subequations}
where, $f_{tA}^\mu=\hat{f}_{tA}^\mu+f_{nA}^\mu\revise{\tilde{\mu}_A}$ and $f_{tB}^\mu=\hat{f}_{tB}^\mu+f_{nB}^\mu\revise{\tilde{\mu}_B}$. The above equations are obtained by representing $f_{nA}\revise{\tilde{\mu}_A}$ as $\epsilon_A$ for contact at $A$ and similarly, $\epsilon_B$ for the contact at $B$. Thus, $\epsilon_A$ and $\epsilon_B$ are the uncertain contact forces for the contacts at $A$ and $B$. The robust formulation that we consider in this paper considers the worst-case effect of these uncertainties on the stability of the object during manipulation. Thus, we try to maximize the bound of these variables $\epsilon_A$ and $\epsilon_B$ using our proposed bilevel optimization. It is noted that $\epsilon_A$ and $\epsilon_B$ are the stability margin for this particular case of stochastic friction.

To ensure that the body maintains contact, we impose $f_{nA}^\mu \geq 0, f_{nB}^\mu \geq 0$, so that we get the following inequalities for $\epsilon_A, \epsilon_B$:
\begin{subequations}
\begin{flalign}
-\mu_Af_{xP}  + \epsilon_A + mg + f_{yP} \leq \mu_A\epsilon_B
\\
\epsilon_B \leq -\mu_B (\epsilon_A + mg + f_{yP})- f_{xP}
 \label{moment_eq12_condition}
\end{flalign}
\label{force_eq_location_friction_condition}
\end{subequations}
To ensure slipping contact even in the presence of uncertainties, we need to satisfy friction cone constraints specified earlier in~\eqref{general_FC}, \eq{slipping_friction_cone}. Using these constraints, we can find the upper and lower bound for the variables $\epsilon_A$ and $\epsilon_B$: 
\begin{subequations}
\begin{flalign}
(\hat{\mu}_A+\revise{\tilde{\mu}_A}) f_{nA}^\mu = \hat{f}_{tA}^\mu+\revise{\tilde{\mu}_A}f_{nA}^\mu \\
(\hat{\mu}_B+\revise{\tilde{\mu}_B}) f_{nB}^\mu = -\hat{f}_{tB}^\mu-\revise{\tilde{\mu}_B}f_{nB}^\mu
\end{flalign}
\label{eq:sliping_eq_friction_uncertainty}
\end{subequations}
To get a lower bound for the variables $\epsilon_A$ and $\epsilon_B$, we make a assumption regarding the uncertainty for the friction coefficients at $A$ and $B$. We assume that the unknown part is bounded above by the known part, i.e., $\revise{\tilde{\mu}_i}\leq \hat{\mu}_i$, $\forall i=A,B$. Note that this is not a restrictive assumption. What this implies is that the above parameter has bounded uncertainty. For simplicity, we assume that uncertainty is bounded by the known part of the parameter. For example, if the friction coefficient is modeled as a stochastic random variable, then we assume that we know the mean of the friction parameter and the standard deviation is bounded by some multiple of mean (note that this bound is just for simplification and one can assume any practical bound for uncertainty).
% In practice, the variances of $\revise{\tilde{\mu}_A}$ and $\revise{\tilde{\mu}_B}$ are relatively small. Thus, we can argue that each realization of $\revise{\tilde{\mu}_A}$ and $\revise{\tilde{\mu}_B}$ are much smaller than $\nu_{\mu_A}$ and $\nu_{\mu_B}$, respectively.  
Consequently, we can derive the following relations:
\begin{subequations}
\begin{flalign}
-\hat{\mu}_A f_{nA}^\mu\leq \epsilon_A \leq \hat{\mu}_A f_{nA}^\mu
\\
-\hat{\mu}_B f_{nB}^\mu\leq \epsilon_B \leq \hat{\mu}_B f_{nB}^\mu
\end{flalign}
\label{eq:sliping_eq_friction_uncertainty_simple}
\end{subequations}
Thus, we get constraints~\eqref{force_eq_location_friction_condition} and~\eqref{eq:sliping_eq_friction_uncertainty_simple} for the stability margin by considering the stability and the friction cone constraints in the presence of uncertain friction coefficients. These constraints are used to estimate the stability margin during the proposed bilevel optimization.

\subsection{\textcolor{blue}{}
Stability Margin for Finger Contact Location}\label{sec:stability_margin_finger_contact_location}
\begin{figure}
    \centering    \includegraphics[width=0.4\textwidth]{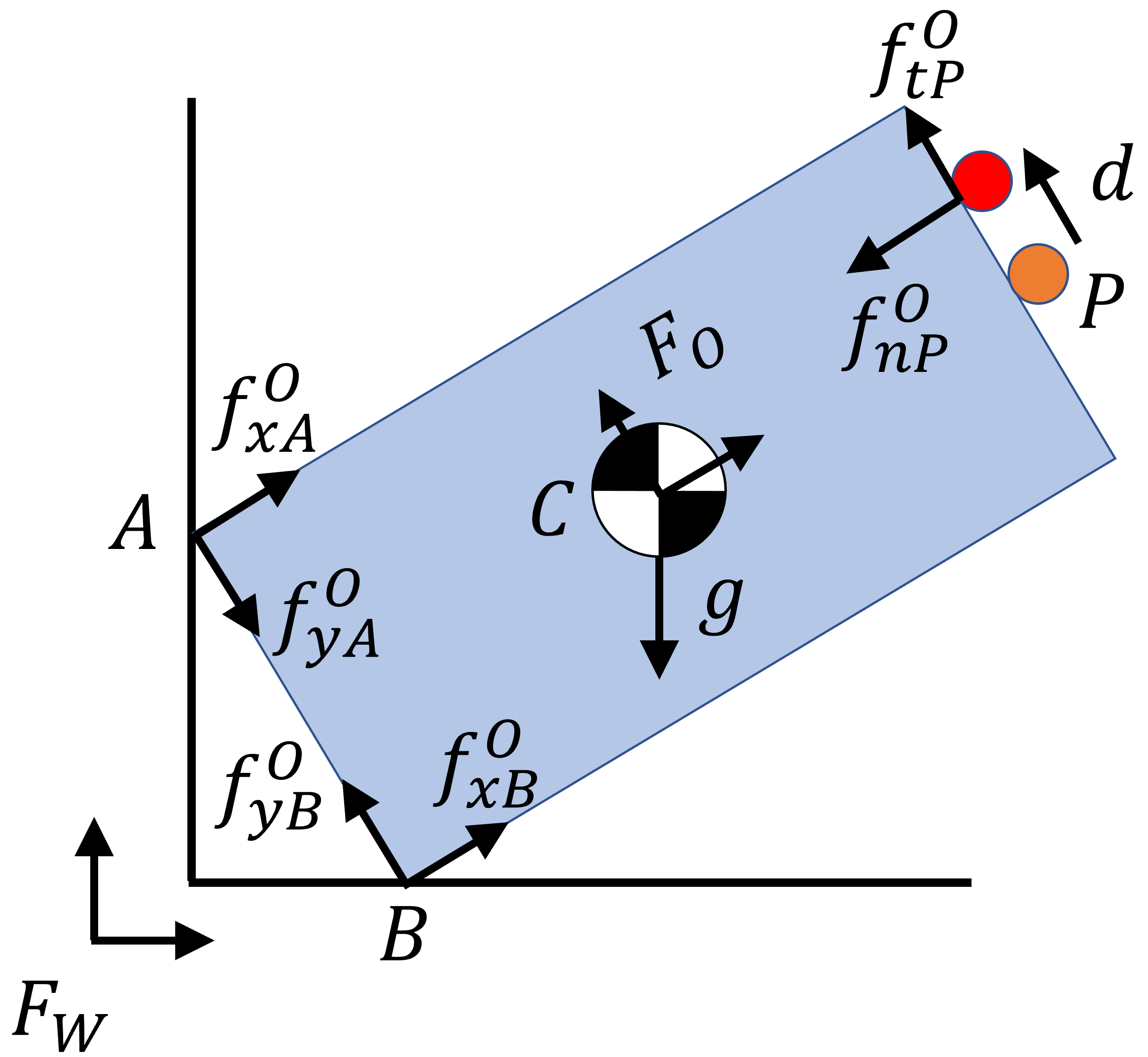} 
    \caption{\revise{
    A schematic showing the free-body diagram of a rigid body during pivoting manipulation. We consider the stability margin of finger location due to imperfect control of stiffness controller in a robotic manipulator.}}
    \label{fig:mechanics_pivoting_finger_margin}
\end{figure}

\revise{
We consider another case of uncertainty which might arise due to an imperfect robot controller or due to imperfect pose information of the object. For this case, we consider the stability margin $d$ of finger contact location on an object, as illustrated in \fig{fig:mechanics_pivoting_finger_margin}. There could be multiple reasons for this uncertainty. One possible reason could be due to imperfect state information for the object being manipulated which can lead to imprecise information about the finger contact location. Another reason could be imprecise stiffness controller of the robot. It is noted that we use a stiffness controller for a position controlled robot to  implement the computed force trajectory. Due to compliance of the object and the robot, the actual robot trajectory is different from the planned and thus, this could lead to this uncertainty. 
% We consider the stability margin  due to imperfect stiffness controller from robotic manipulators. 
We can formulate the following \revise{quasi-static} equilibrium in $F_O$.  We put superscript $d$ for each extrinsic friction variable:
\begin{subequations}
\begin{flalign}
f_{xA}^{O, d} + f_{xB}^{O, d} + mg\sin{\theta} + f_{nP}^O  =0\label{forceeq12_finger},\\
f_{yA}^{O, d} + f_{yB}^{O, d} + mg\cos{\theta} + f_{tP}^O  =0\label{forceeq22_finger}\\
\sum_{i\in\{A, B\}}\left(
i_x^O f_{yi}^{O, d} - i_y^O f_{xi}^{O, d}
\right)
% A_x^O f_{yA}^{O, d} - A_y^O f_{xA}^{O, d} + B_x^O f_{yB}^{O, d} - B_y^O f_{xB}^{O, d} 
 \nonumber \\+P_x^O f_{tP}^O - (P_y^O + d) f_{nP}^O = 0  \label{moment_eq12_finger}
\end{flalign}
\label{force_eq_location_finger}
\end{subequations}
Note that $-A_x^O = -B_x^O = P_x^O =  \frac{l}{2}, A_y^O = -B_y^O = \frac{w}{2}$. Using this relation, we can simplify \eq{force_eq_location_finger}. In particular, we use $f_{xA}^{O, d} \geq 0,  f_{xB}^{O, d} \geq 0,  f_{nP}^{O} \geq 0$ and thus we can get the following bound for $d$:
\begin{subequations}
\begin{flalign}
 \underline{d} \leq d \leq \bar{d} 
\label{forceeq12_finger_margin},\\
\underline{d} = -A_x\frac{mg\cos{\theta} + 2f_{tP}}{f_{nP}}   - A_y \frac{mg\sin{\theta} + f_{nP}}{f_{nP}}  -P_y^O,\\
\bar{d} = -A_x\frac{mg\cos{\theta} + 2f_{tP}}{f_{nP}}   + A_y \frac{mg\sin{\theta} + f_{nP}}{f_{nP}}  -P_y^O
\label{forceeq22_finger_margin}
\end{flalign}
\label{force_eq_location_finger_margin}
\end{subequations}
}
\revise{
When $f_{nP}^O \rightarrow 0$, the equation suggests that $\bar{d}$ tends to infinity and $\underline{d}$ tends to negative infinity. As $f_{nP}^O = 0$ implies no force at point $P$, the finger's placement becomes inconsequential as it does not affect the \revise{quasi-static} equilibrium of the object.}

\revise{
We can consider that uncertainty in finger contact location and uncertainty in the geometry of an object have a similar influence on the manipulation. This is because the relative pose of the object with respect to the robot changes for both cases, resulting in the potential contact mode changes.  
}
% As 
% Consider the case where $f_{np} \rightarrow
%  0$. In this case, \eq{force_eq_location_finger_margin} indicate that $\bar{d}\rightarrow
% \infty$ and $\underline{d}\rightarrow
% -\infty$. Because $f_{np} = 0$ means there is no force at point $P$, the finger can be placed anywhere since finger position does not matter to the static equilibrium of the object. 

\revise{
\subsection{Stability Margin for Uncertain Mass on a Slope}\label{sec:sec_uncertain_mass_slope}
We consider the case where we tilt the two external walls by the angle of $\phi$. 
% as illustrated in \fig{fig:mechanics_pivoting_eq_slope}. 
Our discussion in Sec.~\ref{sec:sec_uncertain_mass} still holds. The only difference arises from gravity terms. Hence, the \revise{quasi-static} equilibrium conditions in $F_B$ can be rewritten as:
\begin{subequations}
\begin{flalign}
f_{nA}^\epsilon + f_{tB}^\epsilon + f_{xP} + (mg + \epsilon) \sin{\phi}  =0\label{forceeq11_slope},\\
f_{tA}^\epsilon + f_{nB}^\epsilon + f_{yP} + (mg + \epsilon) \cos{\phi}    = 0,  \label{forceeq21_slope}\\
A_xf_{tA}^\epsilon - A_yf_{nA}^\epsilon + \left(C_x \cos{\phi} -  C_y \sin{\phi}\right) (mg+\epsilon) \nonumber \\ + P_xf_{yP} - P_y f_{xP} = 0 \label{moment_eq11_slope}
\end{flalign}
\label{force_eq_mass_slope}
\end{subequations}
Following the same logic in Sec.~\ref{sec:sec_uncertain_mass}, we can get the following bound for the stability margin $\epsilon$ under uncertain mass when the object is on a slope: 
\begin{subequations}
\begin{flalign}
\epsilon \geq \frac{P_yf_{xP} - P_xf_{yP} - (C_x\cos{\phi}-C_y \sin{\phi}) mg}{C_x\cos{\phi}-C_y\sin{\phi}}, \nonumber \\ \text{ if } C_x\cos{\phi}>C_y\sin{\phi} \label{fnacond1_slope}\\
\epsilon \leq \frac{P_yf_{xP} - P_xf_{yP} - (C_x\cos{\phi}-C_y \sin{\phi}) mg}{C_x\cos{\phi}-C_y\sin{\phi}}, 
\nonumber \\ \text{ if } C_x\cos{\phi}<C_y\sin{\phi}  \label{fnacond2_slope}
\end{flalign}
\label{fna_cond_mass_slope}
\end{subequations}
As a result, \eq{fnacond1_slope} and \eq{fnacond2_slope} result in the following inequality constraint:
\begin{equation}
 \left(C_x\cos{\phi}-C_y\sin{\phi}  \right)(\epsilon - \epsilon_A) \geq 0
\label{eq:uncertain_mass_slope}
\end{equation}
where $\epsilon_A = \frac{P_yf_{xP} - P_xf_{yP} - (C_x\cos{\phi}-C_y \sin{\phi}) mg}{C_x\cos{\phi}-C_y\sin{\phi}}$. We also derive the bound on $\epsilon$ using $f_{nB}^\epsilon\geq 0$, \eq{fnacond1_slope}, and \eq{fnacond2_slope}:
\begin{equation}
 \left(\mu_A \sin{\phi} - \cos{\phi}  \right)\epsilon \geq f_{yP} -\mu_A f_{xP} 
\label{eq:uncertain_mass_slope_upper_bound}
\end{equation}
Note that the sign of $\mu_A \sin{\phi} - \cos{\phi}$ can change depending on the angle of slope. In this paper, we choose $\phi$ such that the sign of $\mu_A \sin{\phi} - \cos{\phi}$ does not change during manipulation. 
}

\revise{The discussion in this section for manipulation under uncertain mass on a slope can be easily extended with other uncertain parameters such as CoM location, friction, and finger contact location. }

\subsection{Pivoting with Patch Contact between the object and the manipulator}\label{subsec:Pivoting_manipulation}

\begin{figure}[t]
    \centering
    \includegraphics[width=0.35\textwidth]{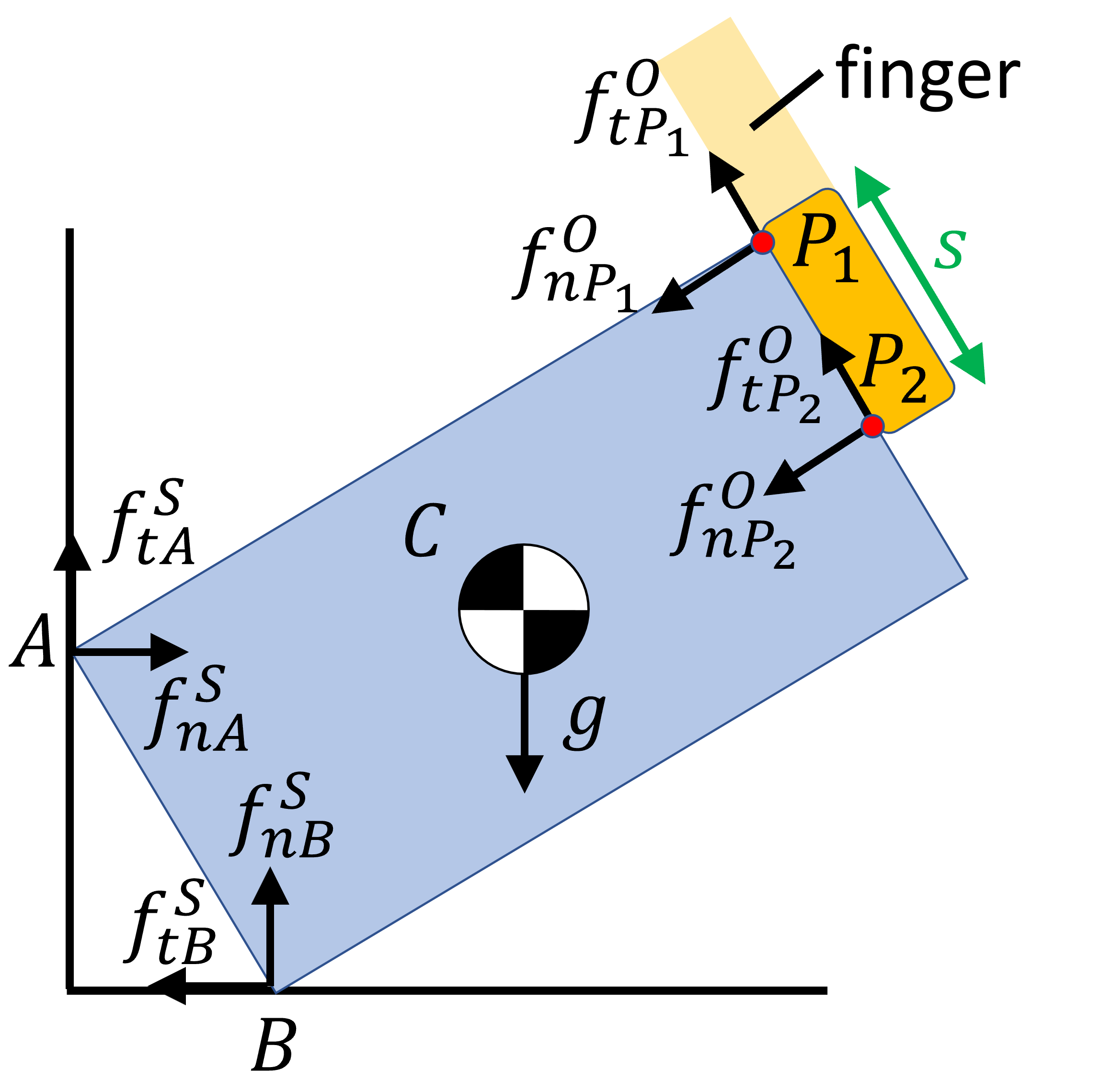} % 
    \caption{A schematic showing the free-body diagram of a rigid body during pivoting manipulation with patch contact. We approximate patch contact as two point contacts $P_1$ and $P_2$ with the same force distribution. We assume that $P_1$ always lies on the vertex of the object for this simplistic patch contact model. \revise{$s$ is the distance between point contact $P_1$ and $P_2$ along $y$-axis of $F_O$.}}
    \label{fig:mechanics_pivoting_patch_eq}
\end{figure}

In the previous sections, we considered point contact between the manipulator and the object. This could be potentially restrictive. Moreover, this may not be a realistic assumption when a robot is interacting with objects. 
In this section, we present a slightly modified formulation by considering patch contact between the object and the manipulator. We would like to analyze and understand how patch contact would compare against a point contact model for stability during pivoting manipulation. \fig{fig:mechanics_pivoting_patch_eq} shows the simplest patch contact model during the pivoting task we consider in this paper. Using this model, we can write the following quasi-static equilibrium:
\begin{subequations}
\begin{flalign}
 f_{nA} + f_{tB} + \revise{f_{xP_{1}}+ f_{xP_{2}}}   =0,\label{forceeq1}\\
 f_{tA} + f_{nB} + mg + \revise{f_{yP_{1}}+ f_{yP_{2}}}= 0,\label{forceeq2}\\
 A_xf_{tA}- A_yf_{nA} + C_xmg  \nonumber \\ + \sum_{i=1}^2 \left(P_{{i}_x} f_{yP_{i}} - P_{{i}_y} f_{xP_{i}}\right) = 0
% f_{tA} + f_{nB} + mg + 2f_{yP}= 0,  \label{forceeq2}\\
% A_xf_{tA} - A_yf_{nA} + C_xmg + \sum_{i=1}^2 \left(P_{ix} f_{y} - P_{iy} f_x\right) = 0 \label{moment_eq1}
\end{flalign}
\label{force_balance_patch_contact}
\end{subequations}
where $P_{{i}_x}, P_{{i}_y}$ represent $x$ and $y$ coordinate of $P_1$ and $P_2$ \revise{in $F_O$}, respectively. 
\revise{In this work, we assume that patch contact as two point contacts $P_1$ and $P_2$ as the same force distribution, which indicates that $f_{xP_{1}} = f_{xP_{2}}, f_{yP_{1}} = f_{yP_{2}}$.}  
\revise{$s$} is the distance between point contact $P_1$ and $P_2$ and \revise{$s$} is a decision variable, meaning that location of $P_2$ is a decision variable and can change over time. In this work, we assume that $P_1$ does not move over time, which simplifies the model of patch contact. 

Using the above \revise{quasi-static} equilibrium conditions with $f_{nA}\geq 0, f_{nB}\geq 0$, we can solve and find the upper and the lower bound of stability margin under the various uncertainties described earlier in the previous subsections. We will present some results in the later section using this formulation and compare them against the point contact formulation.

\revise{\textit{Remark 1}: The patch contact discussion in this section can be extended into the patch contact at extrinsic contact with sliding contacts. We can approximate the extrinsic patch contact as two-point contacts with the same force distribution. Then, we can formulate the quasi-static equilibrium and derive the bound of the stability margin.}

\section{Robust Trajectory Optimization}\label{sec:robust_to}
% also want to say that we denote friction stability as pure function of \epsilon, making inner loop of bilevel optimization linear
%\color[red]{This section can be expanded and re-written to explain everything properly.}\\
% \devesh{This section can be expanded and explained in detail. May be consider and explain the case of patch contact as well as uncertainty due to friction. Also include the formulation for mode-based optimization. The uncertainty due to friction might be a bit different treatment than earlier. The other option to consider is to include recovery from failure using control.}\\
\begin{figure}[t]
    \centering
\includegraphics[width=0.49\textwidth]{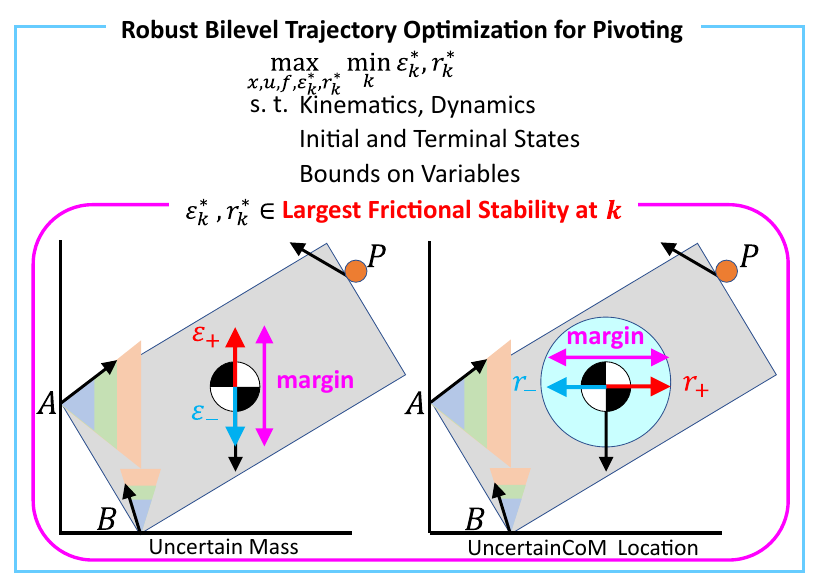} % 
    \caption{Conceptual schematic of our proposed frictional stability and robust trajectory optimization for pivoting. Due to slipping contact, friction forces at points $A, B$ lie on the edge of friction cone. Given the nominal trajectory of state and control inputs, friction forces can account for uncertain physical parameters to satisfy \revise{quasi-static} equilibrium. We define the range of disturbances that can be compensated by contacts as frictional stability. The above figure shows the case of uncertain mass and CoM location.}
    \label{fig:concept}
\end{figure}
% concept.eps
% In our proposed optimization formulation, we maximize the worst frictional stability margin over the entire trajectory where we obtain the maximum frictional stability at each time step given $x, u$, leading to a bilevel optimization problem.
Using the notion of \textit{frictional stability} introduced in the previous section, we describe our proposed contact implicit bilevel optimization (CIBO) method for robust optimization of manipulation trajectories. The proposed method explicitly considers frictional stability under uncertain physical parameters. It is noted that the proposed method considers robustness under slipping contact which results in equality for friction cone constraints (see \fig{fig:concept}).
After describing the formulation for convex objects, we also describe how to extend the proposed CIBO to consider objects with non-convex geometry. Our proposed method is also presented as a schematic in \fig{fig:concept}. As shown in \fig{fig:concept}, the proposed CIBO considers frictional stability margin along the entire trajectory for manipulation and then maximizes the minimum margin in the proposed framework. This is also explained in \fig{fig:cibo}, where we show that we estimate the bound of stability margin in the lower level optimization and maximize the minimum margin in the upper level optimization. Before introducing our proposed bilevel optimization, we present a baseline contact-implicit TO which can be formulated as an MPCC. 

%\textcolor{red}{equality constraints}

%non-convex shape of objects and patch contact in this section.

\subsection{Contact-Implicit Trajectory Optimization for Pivoting}

% Here we first show our general formulation of optimization with uncertainty and show its robust counterpart. we first describe the optimization problem for trajectory generation during pivoting.
The purpose of our optimal control is to 
\revise{find optimal control input sequences under constraints for pivoting manipulation. In particular, we consider the objective function for achieving the minimum motion of objects under kinematics constrains, \revise{quasi-static} equilibrium, friction cone constraints, and sticking-slipping complementarity constraints as follows:}
% The purpose of our optimal control is to regulate the contact state and object state simultaneously given by:
% The purpose of our optimal control is to regulate the contact state and object state simultaneously. Our optimal control problem is:
% 
\begin{subequations}
\begin{flalign}
% \min _{x, u, \lambda}\sum_{k=0}^{N-1} \phi(x_k, u_k, \lambda_k)\\
\min _{x, u, f}  \sum_{k=1}^{N} ({x}_{k} - x_g)^{\top} Q ({x}_{k} - x_g)+\sum_{k=0}^{N-1}u_{k}^{\top} R u_{k} \\
\text{s. t. } i_{k, x}, i_{k, y} \in FK(\theta_k, \revise{P}_{k, y}^{\revise{O}}), \eq{force_eq}, \eq{slipping_friction_cone}, \eq{slippingP},  \label{const2}\\
% \eq{force_eq}, \eq{slipping_friction_cone}, \eq{slippingP}, \label{const3}\\
% \sum_{c=1}^{C} f_{k, c} + mg  = 0\\
% \sum_{c=1}^{C} \left(p_{k, c} - q_{k}\right) \times \lambda_{k, c}  = 0\\
x_{0} = x_s, x_{N} = x_g,
x_{k} \in \mathcal{X}, u_{k} \in \mathcal{U}, 0\leq f_{k, ni} \leq f_{u} \label{bounds_variables}
% f_{k, c, n} \geq 0 ,
%  0 \leq   \dot{p}_{k, c, j+} \perp \mu_c \lambda_{{k,c, n}}-\lambda_{k, c, t} \geq 0  \\
%  0 \leq   \dot{p}_{k, c, j-} \perp \mu_c \lambda_{{k,c, n}}+\lambda_{k, c, t} \geq 0 
\end{flalign}
\label{equation_control}
\end{subequations}
where $x_k = [\theta_k, \revise{P}_{k, y}^{\revise{O}}, \dot{\theta}_k, \revise{\dot{P}}_{k, y}^{\revise{O}}]^\top$, $u_k=[f_{k, nP}, f_{k, tP}]^\top$, $f_k = [f_{k, nA},f_{k, nB}]^\top$, $Q=Q^{\top} \geq 0,R=R^{\top} > 0$.
\revise{The input of \eq{equation_control} consists of physical parameters such as mass, length, and width of the object and the optimization parameters such as $Q$ and $R$. The output of  \eq{equation_control} consists of trajectories of $x_k, u_k, f_k, \forall k \in \{0, 1, \ldots, N\}$. }
We use explicit Euler to discretize the dynamics with sample time $\Delta$. The function $FK$ represents forward kinematics to specify each contact point $i$ and CoM location. $\mathcal{X}$ and $\mathcal{U}$ are convex polytopes, consisting of a finite number of linear inequality constraints.  $f_u$ is an upper-bound of normal force at each contact point. Note that we impose \eq{force_eq}, \eq{slipping_friction_cone} at each time step $k$. $x_s, x_g$ are the states at $k=0,k=N$, respectively.
%  Index $k, c, j$ represent time-step, contact point, appropriate slipping direction, respectively. $x_k$ is the decision variable of states including $q_k$, which is the $x, y, \theta$. $u_k = [\lambda_{k,p,n}, \lambda_{k,p,t}]^\top$ where $\lambda_{k,p,n}, \lambda_{k,p,t}$ are the normal and shear forces at the point P (the point where a robot touches). $\lambda_k$ has the rest of other friction forces at other contact points. 

% In this work, we work on the problem where uncertainty exists in $m, p_{k, c}, \mu_c $. The robust version of \eq{equation_control} can be formulated easily except for equality constraints. 

% \begin{enumerate}
% \item how to realize the control with state and contact force simultaneously? use additional cost?
% \item robust equality constraints
% \end{enumerate}
% The question is eventually what we want to do.

%of Pivoting Considering Frictional Stability
\subsection{Robust CIBO}\label{bilevel_sec}

\begin{figure}
    \centering    \includegraphics[width=0.49\textwidth]{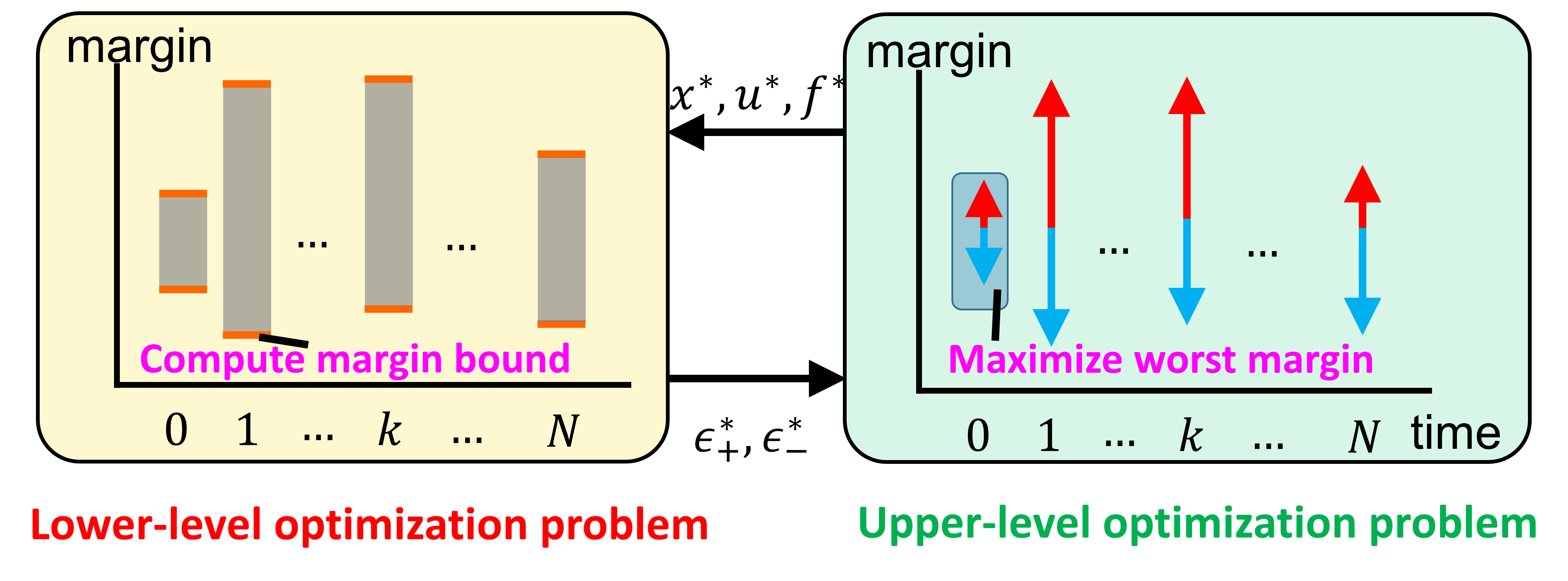} % 
    \caption{This figure illustrates the idea of the proposed contact implicit bilevel optimization, CIBO. Given the trajectory of $x, u, f$, the stability margin over the trajectory can be computed as shown in lower-level optimization problem. Then, given the computed stability margin over the trajectory $\epsilon$, the upper-level optimization problem maximizes the worst-case stability margin over the trajectory by optimizing the trajectory of $x, u, f$. Our CIBO simultaneously optimizes the lower-level optimization problem and the upper-level optimization problem.
    In the right plot, red and blue arrows represent the stability margin along positive and negative directions, respectively. Our CIBO optimizes the stability margin for each direction. 
    }
    \label{fig:cibo}
\end{figure}

% As described in Sec~\ref{sec:mechanics}, considering frictional stability is critical for robust manipulation. % To robustify the trajectory optimization in \eq{equation_control}, we need to incorporate frictional stability in \eq{equation_control}. 
In this section, we present our formulation where we incorporate frictional stability in trajectory optimization to obtain robustness.
In particular, we first focus on discussing the optimization problem with uncertain mass, CoM location, and \revise{finger contact location}. We later discuss the optimization problem of uncertain coefficient of friction in Sec~\ref{subsec:opt_friction}.

An important point to note is that the optimization problem would be ill-posed if we naively add \eq{force_eq_mass}, \eq{force_eq_location}, and/or \revise{\eq{force_eq_location_finger_margin}} to \eq{equation_control} since there is no $u$ to satisfy all uncertainty realization in equality constraints \cite{yalmip_2018}.  
Therefore, our strategy is that we plan to find an optimal nominal trajectory that can ensure external contacts under uncertain physical parameters. 
% In particular, we here focus on discussing uncertain mass or CoM location uncertainty and we later discuss the optimization problem of uncertain coefficient of friction in Sec~\ref{subsec:opt_friction}.
In other words, we aim at maximizing the worst-case stability margin over the trajectory given the maximal frictional stability at each time-step $k$ (also shown in \fig{fig:concept}). Thus,  we maximize the following objective function:
\begin{equation}
\min _{k} \epsilon_{k, +}^* - \max _{k} -\epsilon_{k, -}^*
\label{bilevel_obj}
\end{equation}
where $\epsilon_{k, +}^*, \epsilon_{k, -}^*$ are non-negative variables. Note that $\epsilon_{k, +}^*, \epsilon_{k, -}^*$ are the largest uncertainty in the positive and negative direction, respectively, at instant $k$ given $x, u, f$, which results in non-zero contact forces (i.e., stability margin, see also \fig{fig:concept}).
% 
% that represents the magnitude of friction force (i.e., frictional stability margin) under $\epsilon_{k, +}^*, \epsilon_{k, -}^*$ along positive and negative directions of uncertainty at instant $k$ given $x, u, f$, respectively (see also Figure~\ref{fig:concept}).
% 
% the maximum stability margin along positive and negative direction of uncertainty at $k$ given $x, u, f$, respectively. 
\eq{bilevel_obj} calculates the smallest stability margin over time-horizons by subtracting the stability margin along the positive direction from that along the negative direction. 
Hence, we formulate a bilevel optimization problem which consists of two lower-level optimization problems as follows (see also \fig{fig:cibo}):
\begin{subequations}
\begin{flalign}
\max_{x, u, f, \epsilon_+^*, \epsilon_-^*} (\min _{k} \epsilon_{k, +}^* - \max _{k} -\epsilon_{k, -}^*)  \ \\
% \max_{x, u, f, \epsilon_+^*, \epsilon_-^*} (\min _{k} \epsilon_{k, +}^* + q\min _{k} \epsilon_{k, -}^* )  \ \\
\text{s. t. } \quad \text{\eq{const2}, \eq{bounds_variables}}, \\
\epsilon_{k, +}^* \in \argmax_{\epsilon_{k, +}} \{\epsilon_{k, +}: A_k\epsilon_{k, +} \leq b_k , \epsilon_{k, +} \geq 0 \}, \label{bi-const1} \\
\epsilon_{k, -}^* \in \argmax_{\epsilon_{k, -}} \{\epsilon_{k, -}: -A_k\epsilon_{k, -} \leq b_k , \epsilon_{k, -} \geq 0 \}
%  \text{s. t. } \eq{fna_cond_mass_one}, \eq{fnb_cond_mass}\label{force_eq_inner}
%  \sum_{c=1}^{C} p_{k, c} \times \lambda_{k, c} + p_{k, u} \times R\left(x_{k, o}\right) u_k + r_{k, g} \times (\bar{m}g + \epsilon_k)  = 0\\
% \lambda_{k, c, n} \geq 0 ,\\
% \mu_c \lambda_{{k,c, n}}-\lambda_{k, c, t} = 0\label{friction_eq_inner}
% \mu_c \lambda_{{k,c, n}}+\lambda_{k, c, t} = 0 
%   0 \leq   \dot{p}_{k, c, j+} \perp \mu_c \lambda_{{k,c, n}}-\lambda_{k, c, t} \geq 0  \\
%  0 \leq   \dot{p}_{k, c, j-} \perp \mu_c \lambda_{{k,c, n}}+\lambda_{k, c, t} \geq 0 
\end{flalign}
\label{equation_sm_1}
\end{subequations}
where $A_k \in\mathbb{R}^{2 \times 1}, b_k \in\mathbb{R}^{2 \times1}$ represent inequality constraints in \eq{fna_cond_mass_one} and \eq{fnb_cond_mass} \revise{or \eq{eq:uncertain_mass_slope} and  \eq{eq:uncertain_mass_slope_upper_bound} if the object is on a slope.} 
 $A_k\epsilon_{k, +} \leq b_k , \epsilon_{k, +} \geq 0,$ and $-A_k\epsilon_{k, -} \leq b_k , \epsilon_{k, -} \geq 0$ represent the lower-level constraints for each lower-level optimization problem while \eq{const2}, \eq{bounds_variables} represent the upper-level constraints. $\epsilon_+, \epsilon_-$ are the lower-level objective functions while $\min _{k} \epsilon_{k, +}^* - \max _{k} -\epsilon_{k, -}^* $ is the upper-level objective function. $\epsilon_{k, +}, \epsilon_{k, -}$ are the lower-level decision variables of each lower-level optimization problem while $x, u, f, \epsilon_+^*, \epsilon_-^*$ are the upper-level decision variables. 

% Note that $A_k, b_k$ are nonlinear function with respect to $x, u, f$.  

\eq{equation_sm_1} considers the largest one-side frictional stability margin along positive and negative direction at $k$. Therefore, by solving these two lower-level optimization problems, we are able to obtain the maximum frictional stability margin along positive and negative direction. 
% We initially plan to solve the lower-level optimization problem which objective function is $\epsilon_k^2$ to obtain the largest magnitude of $\epsilon_k$ but this objective function makes the lower-level optimization non-convex, which is difficult to solve.
% In contrast, 
The advantage of \eq{equation_sm_1} is that since the lower-level optimization problem are formulated as two linear programming problems, we can efficiently solve the entire bilevel optimization problem using the Karush-Kuhn-Tucker (KKT) condition as follows:
\begin{subequations}
\begin{flalign}
 w_{k, +, j}, w_{k, -, j} \geq 0, C_k\epsilon_{k, +} \leq d_k , E_k\epsilon_{k, -} \leq d_k,\\
w_{k, +, j}(C_k\epsilon_{k, +} - d_k)_j = 0, \\
w_{k, -, j}(E_k\epsilon_{k, -} - d_k)_j = 0, \\
% \epsilon_{k, +}^* \in \argmax_{\epsilon_+} \{\epsilon_{k, +}: A_k\epsilon_{k, +} \leq b_k , \epsilon_{k, +} \geq 0 \}, \\
% \epsilon_{k, -}^* \in \argmax_{\epsilon_-} \{\epsilon_{k, -}: -A_k\epsilon_{k, -} \leq b_k , \epsilon_{k, -} \geq 0 \}, \\
\nabla (-\epsilon_{k, +}) + \sum_{j=1}^{3}w_{k, +, j} \nabla (C_k\epsilon_{k, +} - d_k)_j= 0,\\
\nabla (-\epsilon_{k, -}) + \sum_{j=1}^{3}w_{k, -, j} \nabla (E_k\epsilon_{k, -} - d_k)_j= 0
% t_+ \leq \epsilon_{k, +}, t_- \leq \epsilon_{k, -}, \forall k
%  \text{s. t. } \eq{fna_cond_mass_one}, \eq{fnb_cond_mass}\label{force_eq_inner}
%  \sum_{c=1}^{C} p_{k, c} \times \lambda_{k, c} + p_{k, u} \times R\left(x_{k, o}\right) u_k + r_{k, g} \times (\bar{m}g + \epsilon_k)  = 0\\
% \lambda_{k, c, n} \geq 0 ,\\
% \mu_c \lambda_{{k,c, n}}-\lambda_{k, c, t} = 0\label{friction_eq_inner}
% \mu_c \lambda_{{k,c, n}}+\lambda_{k, c, t} = 0 
%   0 \leq   \dot{p}_{k, c, j+} \perp \mu_c \lambda_{{k,c, n}}-\lambda_{k, c, t} \geq 0  \\
%  0 \leq   \dot{p}_{k, c, j-} \perp \mu_c \lambda_{{k,c, n}}+\lambda_{k, c, t} \geq 0 
\end{flalign}
\label{kkt_equations}
\end{subequations}
where $C_k = [A_k^\top, -1]^\top \in\mathbb{R}^{3 \times 1}, d_k = [b_k^\top, 0]^\top \in\mathbb{R}^{3\times 1}, E_k = [-A_k^\top, -1]^\top \in\mathbb{R}^{3 \times 1}$.
$w_{k, +, j}$ is Lagrange multiplier associated with  $(C_k\epsilon_{k, +} \leq d_k)_j$, where $(C_k\epsilon_{k, +} \leq d_k)_j$ represents the $j$-th inequality constraints in $C_k\epsilon_{k, +} \leq d_k$. $w_{k, -, j}$ is Lagrange multiplier associated with  $(E_k\epsilon_{k, -} \leq d_k)_j$. 
Using the KKT condition and epigraph trick, we eventually obtain a single-level large-scale nonlinear programming problem with complementarity constraints: 
% where $\epsilon_+^*, \epsilon_-^*$ are introduced instead of $\epsilon^*$ so \eq{equation_sm_1} is able to solve lins
\begin{subequations}
\begin{flalign}
\max_{x, u, f, \epsilon_+^*, \epsilon_-^*} (t_+ + \alpha t_- ) \label{cost_bilevel}  \ \\
\text{s. t. } \quad \text{\eq{const2}, \eq{bounds_variables}, \eq{kkt_equations}},\\
% w_{k, +, j}, w_{k, -, j} \geq 0\\
% C_k\epsilon_{k, +} \leq d_k , E_k\epsilon_{k, -} \leq d_k,\\
% w_{k, +, j}(C_k\epsilon_{k, +} - d_k)_j = 0, \\
% w_{k, -, j}(E_k\epsilon_{k, -} - d_k)_j = 0, \\
% % \epsilon_{k, +}^* \in \argmax_{\epsilon_+} \{\epsilon_{k, +}: A_k\epsilon_{k, +} \leq b_k , \epsilon_{k, +} \geq 0 \}, \\
% % \epsilon_{k, -}^* \in \argmax_{\epsilon_-} \{\epsilon_{k, -}: -A_k\epsilon_{k, -} \leq b_k , \epsilon_{k, -} \geq 0 \}, \\
% \nabla (-\epsilon_{k, +}) + \sum_{j=1}^{3}w_{k, +, j} \nabla (C_k\epsilon_{k, +} - d_k)_j= 0,\\
% \nabla (-\epsilon_{k, -}) + \sum_{j=1}^{3}w_{k, -, j} \nabla (E_k\epsilon_{k, +} - d_k)_j= 0, \\
t_+ \leq \epsilon_{k, +}, t_- \leq \epsilon_{k, -}, \forall k
%  \text{s. t. } \eq{fna_cond_mass_one}, \eq{fnb_cond_mass}\label{force_eq_inner}
%  \sum_{c=1}^{C} p_{k, c} \times \lambda_{k, c} + p_{k, u} \times R\left(x_{k, o}\right) u_k + r_{k, g} \times (\bar{m}g + \epsilon_k)  = 0\\
% \lambda_{k, c, n} \geq 0 ,\\
% \mu_c \lambda_{{k,c, n}}-\lambda_{k, c, t} = 0\label{friction_eq_inner}
% \mu_c \lambda_{{k,c, n}}+\lambda_{k, c, t} = 0 
%   0 \leq   \dot{p}_{k, c, j+} \perp \mu_c \lambda_{{k,c, n}}-\lambda_{k, c, t} \geq 0  \\
%  0 \leq   \dot{p}_{k, c, j-} \perp \mu_c \lambda_{{k,c, n}}+\lambda_{k, c, t} \geq 0 
\end{flalign}
\label{kkt_convertion}
\end{subequations}
% where $C_k \in\mathbb{R}^{3 \times 1}, d_k \in\mathbb{R}^{3}$ represent inequality constraints associated with $\epsilon_{k, +}$.  $E_k \in\mathbb{R}^{3 \times 1}, d_k \in\mathbb{R}^{3}$ represent inequality constraints associated with $\epsilon_{k, -}$.
where $\alpha$ is a weighting scalar. 
Note that we derive \eq{kkt_convertion} for the case with an uncertain mass parameter but this formulation can be easily converted to the case where uncertainty exists in CoM location by replacing $A_k, b_k$ in \eq{equation_sm_1} with \eq{fna_fnb_r}.
\revise{Similarly, we can consider uncertainty in finger contact location by replacing $A_k, b_k$ in \eq{equation_sm_1} with \eq{force_eq_location_finger_margin}.}
% with $\epsilon_k^* \in \argmax_{\epsilon} \{\epsilon_k^2:\text{\eq{fna_fnb_r}} \} $ in \eq{equation_sm}. 
Therefore, by solving tractable \eq{kkt_convertion}, we can efficiently generate robust trajectories that are robust against uncertain mass, CoM location, \revise{and contact location} parameters. 

% \textit{Remark 1}:
% In practice, we can add $\min _{x, u, f} \sum_{k=0}^{N-1} ({x}_{k} - x_g)^{\top} Q ({x}_{k} - x_g)+u_{k}^{\top} R u_{k}$ to \eq{cost_bilevel} to realize a smooth trajectory and avoid jerky control inputs. 

\textit{Remark \revise{2}}:
If we consider the case where uncertainty exists in both mass and CoM location simultaneously, we would have a nonlinear coupling term $(C_x+r)(mg + \epsilon)$ in \revise{quasi-static} equilibrium of moment. This makes the lower-level optimization non-convex optimization, making it extremely challenging to solve during bilevel optimization.
Once the lower-level optimization becomes a non-convex optimization problem, there is no guarantee that the lower-level optimization finds globally optimal solutions, resulting in finding a very sub-optimal controller.
\revise{Similarly, all of the constraints (e.g., considering sticking-slipping contact at point contact $A$ and $B$ requires complementarity constraints) which results in non-convex constraints cannot be handled in our CIBO.}

\subsection{Robust CIBO for Frictional Uncertainty}
\label{subsec:opt_friction}
We consider the case where the system has uncertainty in the friction coefficients at $A$ and $B$ as discussed in Sec~\ref{subsec:stochasticfriction_planning}. In order to design a robust open-loop controller for the system,  we can use the similar formulation presented in Sec~\ref{bilevel_sec}.
The proposed formulation aims at maximizing the stability margin from stochastic friction. In particular, to avoid non-convex optimization as the lower-level optimization problem, we consider the stability margin along positive and negative direction for both $\epsilon_A$ and $\epsilon_B$, as we discuss in Sec~\ref{bilevel_sec}. By borrowing the optimization problem \eq{equation_sm_1},  the proposed formulation can be seen as follows.
For simplicity, we abbreviate subscript $k$. 
% \begin{subequations}
% \begin{flalign}
% \max_{\Gamma} (\min _{k} \epsilon_{A, +}^* - \max _{k} -\epsilon_{A, -}^* + \min _{k} \epsilon_{B, +}^* - \max _{k} -\epsilon_{B, -}^*)  \ \\
% \text{s. t. } \quad \text{\eq{const2}, \eq{bounds_variables}}, \\
% \epsilon_{A, +}^* \in \argmax_{\epsilon_{A, +}} \{\epsilon_{A, +}:  g(x, u, f, \epsilon_{A, +}, \epsilon_{B}^*)\leq 0 , \nonumber \\ \epsilon_{A, +} \geq 0, \epsilon_{B}^* \in [-\epsilon_{B, -}^*, \epsilon_{B, +}^*]  \}, \label{bi-const1-friction1} \\
% \epsilon_{A, -}^* \in \argmax_{\epsilon_{A, -}} \{\epsilon_{A, -}:  g(x, u, f, -\epsilon_{A, -}, \epsilon_{B}^*)\leq 0 , \nonumber \\ \epsilon_{A, -} \geq 0, \epsilon_{B}^* \in [-\epsilon_{B, -}^*, \epsilon_{B, +}^*]  \}, \label{bi-const1-friction2} \\
% \epsilon_{B, +}^* \in \argmax_{\epsilon_{B, +}} \{\epsilon_{B, +}:  g(x, u, f, \epsilon_{B, +}, \epsilon_{A}^*)\leq 0 , \nonumber \\ \epsilon_{B, +} \geq 0, \epsilon_{A}^* \in [-\epsilon_{A, -}^*, \epsilon_{A, +}^*]  \}, \label{bi-const1-friction3} \\
% \epsilon_{B, -}^* \in \argmax_{\epsilon_{B, -}} \{\epsilon_{B, -}:  g(x, u, f, -\epsilon_{B, -}, \epsilon_{A}^*)\leq 0 , \nonumber \\ \epsilon_{B, -} \geq 0, \epsilon_{A}^* \in [-\epsilon_{A, -}^*, \epsilon_{A, +}^*]  \}, \label{bi-const1-friction4} 
% \end{flalign}
% \label{eq:bilvel_friction}
% \end{subequations}
% Actually, we can instead have the following optimization problem by moving the constraints:  
\begin{subequations}
\begin{flalign}
% \max_{\Gamma} (\min _{k} \epsilon_{A, +}^* - \max _{k} -\epsilon_{A, -}^* + \min _{k} \epsilon_{B, +}^* - \max _{k} -\epsilon_{B, -}^*)  \ \\
\max_{x, u, f, \epsilon_{A,+}^*, \epsilon_{A, -}^*, \epsilon_{B,+}^*, \epsilon_{B, -}^*} \sum_{c\in \mathcal{C}} (\min _{k} \epsilon_{c, +}^* - \max _{k} -\epsilon_{c, -}^*)\\
\text{s. t. } \quad \text{\eq{const2}, \eq{bounds_variables}}, \\
\epsilon_{A}^* \in [-\epsilon_{A, -}^*, \epsilon_{A, +}^*], \epsilon_{B}^* \in [-\epsilon_{B, -}^*, \epsilon_{B, +}^*], \label{eq:stochastic_friction_bilevel_const0} \\
\epsilon_{A, +}^* \in \argmax_{\epsilon_{A, +}} \{\epsilon_{A, +}:  g(x, u, f, \epsilon_{A, +},  \epsilon_{B}^*)\leq 0,  \nonumber \\
\epsilon_{A, +} \geq 0 \}, \label{bi-const1-friction12} \\
\epsilon_{A, -}^* \in \argmax_{\epsilon_{A, -}} \{\epsilon_{A, -}:  g(x, u, f, -\epsilon_{A, -}, \epsilon_{B}^*)\leq 0, \nonumber \\
\epsilon_{A, -} \geq 0,   \}, \label{bi-const1-friction22} \\
\epsilon_{B, +}^* \in \argmax_{\epsilon_{B, +}} \{\epsilon_{B, +}:  g(x, u, f, \epsilon_{B, +}, \epsilon_{A}^*)\leq 0, \nonumber \\ \epsilon_{B, +} \geq 0, \}, \label{bi-const1-friction32} \\
\epsilon_{B, -}^* \in \argmax_{\epsilon_{B, -}} \{\epsilon_{B, -}:  g(x, u, f, -\epsilon_{B, -}, \epsilon_{A}^*)\leq 0, \nonumber \\
\epsilon_{B, -} \geq 0  \}, \label{bi-const1-friction42} 
\end{flalign}
\label{eq:bilvel_friction2}
\end{subequations}
where $g$ summarizes the constraints for each lower-level optimization problem and $\mathcal{C} = \{A, B\}$. 
For each lower-level optimization problem, we consider that another uncertain friction is in the range of optimal stability margin.
For instance, \eq{bi-const1-friction12} is one of the four lower-level optimization problems which aims at maximizing the stability margin under stochastic friction forces at $A$, given stochastic friction force at $B$, $\epsilon_{B}^*$. \eq{eq:stochastic_friction_bilevel_const0}  ensures that $\epsilon_{B}^*$ needs to be within the range of stability margin computed from other two lower-level optimization problems \eq{bi-const1-friction32} and \eq{bi-const1-friction42}.

The resulting optimization introduces many complementarity constraints through the KKT condition because of four lower-level optimization problems, but the resulting computation is still tractable. We discuss computational results in Sec~\ref{subsec::computation}.

\revise{\textit{Remark 3}: In practice, the choice of the particular parameter for the which one should use CIBO to obtain robust trajectories depends on the amount of uncertainty in different parameters associated with the manipulation task. For instance, if we have access to the CAD model of the objects, we can have a good guess of mass and CoM location of the object and thus the major source of uncertainty can be from other parameters such as coefficients of friction.}

\subsection{Robust CIBO for Non-Convex Objects}\label{subsec:mode_based_optimization}
\begin{figure*}[t]
    \centering
    \includegraphics[width=0.8\textwidth]{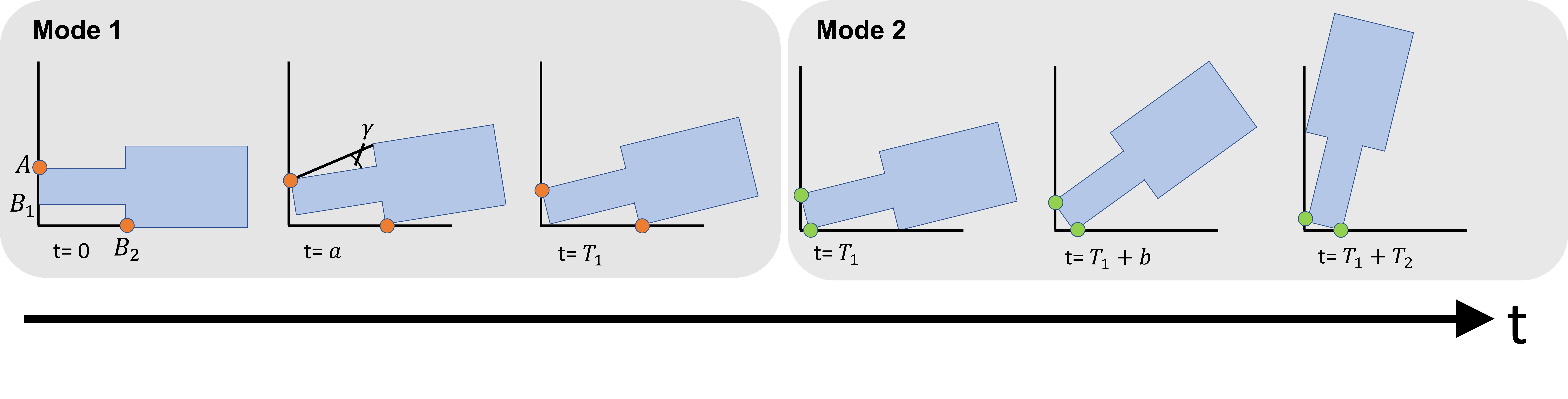}  
    \caption{A schematic of pivoting for a non-convex shape object where contact set changes over time. During mode 1, the peg rotates with contact at $A$ and $B_2$. During mode 2, the peg rotates with contact at $A$ and $B_1$. 
    $\gamma$ represents one of the kinematic features of peg, which is used to discuss the result in Sec~\ref{fig:openloop_result}. 
    }
    \label{fig:mode_concept}
\end{figure*}

% \devesh{Rephrase to use the formulation in previous subsections}\\
% While our proposed framework in \cite{9811812} is able to design the robust open-loop controller under uncertain physical parameters, it is unable to design the controller for non-convex shape objects such as pegs as shown in \fig{fig:pivoting_abstractfig}. 
The method introduced in the previous subsections assumes convex geometry of the object being manipulated and can not be applied to objects with non-convex geometry (such as pegs as shown in \fig{fig:pivoting_abstractfig}). This is because non-convex objects could result in different contact formations between the object and the environment
 and it is not trivial to identify a feasible contact sequence. In \cite{9811812}, the proposed optimization \eq{kkt_convertion} was solved sequentially for pegs with non-convex geometry. As illustrated in \fig{fig:mode_concept},  we first solve the optimization for a particular contact set (i.e., mode 1 in \fig{fig:mode_concept}) and then solve the optimization for another contact set (i.e., mode 2 in \fig{fig:mode_concept}) given the solution obtained from the first optimization. 
While this method works, it requires extensive domain knowledge. We observed that the second stage optimization can result in infeasible solutions given the solution from the first stage optimization. Thus, we had to carefully specify the parameters of optimization and, in particular, the initial state and terminal state constraints. Such a hierarchical approach has difficulty in finding a feasible solution once the object becomes more complicated. 

To overcome these issues, in general, complementarity constraints can be used to model the change of contact. However, introducing complementarity constraints inside the lower-level optimization makes the lower-level optimization non-convex optimization. Hence, the KKT condition is not a necessary and sufficient condition for optimality but rather a necessary condition. Thus, it is not guaranteed to find globally optimal safety margins over the trajectory. 

In this work, we propose another approach to deal with the non-convexity of the object. Inspired by \cite{9812069}, we formulate the optimization that optimizes the trajectory given mode sequences instead of optimizing mode sequences. It is worth noting that our framework still optimizes the trajectory over the time duration of each mode given the sequence of the mode. Our goal is that the optimization has a larger feasible space so that less domain knowledge is required.%, resulting in less parameter tuning. 

Using the formulation presented in~\cite{9812069}, we present a mode-based formulation for non-convex shaped objects. 
See \cite{9812069} for more details regarding mode-based optimization. For simplicity of exposition, we only present the formulation considering two modes. But one can easily extend this to problems with multiple modes.
% We consider the case where the system has two contact sets and we define the mode as $m = \{\left(l, b\left(l\right)| l\in \mathcal{L}, b\left(l\right) \in \{1, 2\}\right)\}$ where $\mathcal{L}$ is a set of pairs of indicies that defines the complementarity relationship. 
% As shown in FIGURE, we use $m_1$ and $m_2$ to represent mode 1 (i.e., contact on $A$ and $B_1$) and mode 2 (i.e., contact on $A$ and $B_2$), respectively. 
% Thus, we can define each mode as 
For each contact mode, the system has the different constraints. For brevity, we abbreviate the subscript $k$:
\begin{subequations}
\begin{flalign}
% \min _{\tilde{x}, u, f} \sum_{k=0}^{N-1} (\tilde{x}_{k} - x_g)^{\top} Q (\tilde{x}_{k} - x_g)+u_{k}^{\top} R u_{k} + \sum_{l=1}^2 T_l \\
% \text{s. t. } 
i_{x}, i_{y} \in FK_{m}(\theta_k, \revise{P}_{k, y}^{\revise{O}}),  \forall i  \in \{A, B_m\}\\
g_m(f_{nA}, f_{tA}, f_{nB_{1}}, f_{tB_{1}}, f_{nP}, f_{tP}, \revise{P}_{y}^{\revise{O}}) \ \text{if} \ m = 1
\\
g_m(f_{nA}, f_{tA}, f_{nB_{2}}, f_{tB_{2}}, f_{nP}, f_{tP}, \revise{P}_{y}^{\revise{O}}) \ \text{if} \ m = 2 \label{eq:mode2_dynamics_const}
\\
 f_{tA}  =\mu_A f_{nA}, f_{tB_{1}}  =-\mu_{B_1} f_{nB_{1}}, f_{tB_{2}}  =-\mu_{B_2} f_{nB_{2}}
 \\
 \eq{slippingP}, x_{k} \in \mathcal{X}, u_{k} \in \mathcal{U}, 0\leq f_{k, ni} \leq f_{u}
\end{flalign}
\label{eq:mode_const}
\end{subequations}
where $m \in \{1, 2\}$ to represent each contact mode. 
$g_m$ represents the quasi-static model of pivoting manipulation for mode $m$.
It is worth noting that since we decompose the optimization problem into the two mode optimization problem, complementarity constraints are encoded for each mode. %\devesh{Yuki : the above equations are not clear. What is g? have we introduced it earlier?}

 What the optimization problem needs to perform is that for each mode, it only considers the associated constraints and does not consider constraints associated with different contact mode. For example, during mode 1, the optimization should consider only constraints associated with mode 1 and should not consider constraints such as \eq{eq:mode2_dynamics_const}. 
 Another thing the optimization needs to consider is that it needs to scale $\dot{\theta}, \revise{\dot{P}}_{y}^{\revise{O}}$ since we would like to optimize over the time duration. To achieve that, we employ the scaled time variables as discussed in \cite{9812069}. As a result, 
 we recast the quasi-static model by introducing a new state variable with a scaled time, $\tilde{x}_k = \left[\theta_k, \revise{P}_{k, y}^{\revise{O}}, \frac{\dot{\theta}_k}{T}, \frac{\revise{\dot{P}}_{k, y}^{\revise{O}}}{T}\right]^\top$ where $T = T_1$ during mode 1 and $T = T_2$ during mode 2.

 For two contact modes, we can remodel our optimization \eq{equation_control} as follows:
\begin{subequations}
\begin{flalign}
\min _{\tilde{x}, u, f} \sum_{k=0}^{N-1} (\tilde{x}_{k} - x_g)^{\top} Q (\tilde{x}_{k} - x_g)+u_{k}^{\top} R u_{k} + \sum_{l=1}^2 T_l \\
\text{s. t. }  \revise{h_1}(\tilde{x}_k, u_k, f_k) \leq 0, \text{for} \ k\Delta \leq 1 \\
\revise{h_2}(\tilde{x}_k, u_k, f_k) \leq 0, \text{for} \ k\Delta > 1
\label{bounds_variables_mode}
\end{flalign}
\label{eq:mode_change}
\end{subequations}
where $\tilde{x}_k = \left[\theta_k, \revise{P}_{k, y}^{\revise{O}}, \frac{\dot{\theta}_k}{T_1}, \frac{\revise{\dot{P}}_{k, y}^{\revise{O}}}{T_1}\right]^\top$ for $k\Delta \leq 1$ and $\tilde{x}_k = \left[\theta_k, \revise{P}_{k, y}^{\revise{O}}, \frac{\dot{\theta}_k}{T_2}, \frac{\revise{\dot{P}}_{k, y}^{\revise{O}}}{T_2}\right]^\top$ for $k\Delta > 1$.
We use $\revise{h_1}$ and $\revise{h_2}$ to represent all constraints for each mode. 
Given \eq{eq:mode_change}, we can obtain bilevel optimization formulation for non-convex shape objects by following the logic in Sec~\ref{bilevel_sec}.

\subsection{Robust CIBO with Patch Contact}
The formulation for robust CIBO is similar to the point contact case except that the underlying equilibrium conditions are different. The \revise{quasi-static} equilibrium conditions for the patch contact case were earlier presented in~\eqref{force_balance_patch_contact}. Using these equations and the analysis presented in Sections~\ref{sec:sec_uncertain_mass} through~\ref{subsec:stochasticfriction_planning}, it is straightforward to compute the constraints for the corresponding robust CIBO similar to~\eqref{equation_sm_1}. More explicitly, this can be achieved by computing the appropriate constraints of the type $A_k\epsilon_{k, +} \leq b_k$ and $-A_k\epsilon_{k, -} \leq b_k$ using~\eqref{force_balance_patch_contact} and the frictional stability margin discussion in Sec~\ref{subsec:Pivoting_manipulation}.

% Given patch contact model, we can formulate optimization problem in the same fashion using \eq{equation_sm_1}. We simply need to replace  with the constraints discussed in Sec~\ref{subsec:Pivoting_manipulation}.
% \textcolor{red}{where do we say considering complementarity constraints would make optimization intractable? here or in Sec~\ref{subsec:Pivoting_manipulation}?}

% \input{feedbackmanipulation/instability_detection}
\section{Experimental Results}\label{sec:results}
%We verify our proposed optimal control presented in Sec~\ref{sec:robust_to} for pivoting. 
In this section, we verify the performance of our proposed approach for pivoting. Through the experiments we present in this section, we evaluate the efficacy of the proposed planner in several different settings and the computational requirement of the method. We also present results of implementation of the proposed planner on a robotic system using a 6 DoF manipulator arm and compare it against a baseline trajectory optimization method.

% We present experiments to answer the following questions:
% \begin{enumerate}
%     \item How much robustness can the proposed bilevel optimization method provide over the baseline method?
%     \item Can we demonstrate robustness of the proposed optimization during control of pivoting manipulation?
% \end{enumerate}

%We first show the result of our proposed controller \eq{kkt_convertion} compared to a baseline trajectory generated by solving \eq{equation_control}. %We also implement our controller using a physical manipulator arm and evaluate the performance.

\begin{table}[t]
    \caption{{Parameters of objects. $m, l, w$ represent the mass, length, and the width of the object, respectively. For pegs, the first element in $l, w$ are $l_1, w_1$ and the second element in $l, w$ are $l_2, w_2$, respectively, shown in \fig{fig:hardwareresults}.} For pegs, since they are made of the same material and they make contact on the same environment, we can assume $\mu_B = \mu_{B_1} = \mu_{B_2}$. }
    \centering
    \begin{tabular}{c|c|c|c|c}
     & $m$ [g] & $l$ [mm] & $w$ [mm] & $\mu_A, \mu_B, \mu_P$\\
         \hline\hline gear 1 & 140 & 84 & 20 & 0.3, 0.3, 0.8\\
         \hline gear 2 & 100 & 121 & 9.5 & 0.3, 0.3, 0.8\\
         \hline gear 3 & 280 & 84 & 20 & 0.3, 0.3, 0.8\\
         \hline peg 1 & 45 & 36, 40 & 20, 28 & 0.3, 0.3, 0.8\\
         \hline peg 2 & 85 & 28, 40 & 10, 11 & 0.3, 0.3, 0.8
         \\
         \hline peg 3 & 85 & 28, 40 & 10, 27.5 & 0.3, 0.3, 0.8
    \end{tabular}
    \label{parameter_table}
\end{table}

\begin{figure*}
    \centering
    \includegraphics[width=0.9\textwidth]{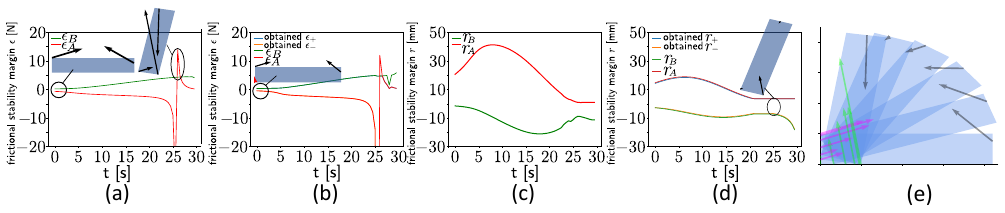} % 
    \caption{Trajectory of frictional stability margin. $\epsilon_A, \epsilon_B$ are bounds of $\epsilon$ from \eq{fna_cond_mass_one}, \eq{fnb_cond_mass}. $r_A, r_B$ are bounds of $r$ from \eq{fna_fnb_r}.  $\epsilon_+, \epsilon_-, r_+, r_i$ are solutions obtained from CIBO. (a), (b): Trajectory of frictional stability of gear 1 based on uncertain mass obtained from baseline optimization, our CIBO, respectively. (c), (d): Trajectory of frictional stability of gear 1 based on uncertain CoM location obtained from baseline optimization, CIBO, respectively. (e): Snapshots of pivoting motion for gear 1 obtained from CIBO considering uncertain CoM location.}
    \label{fig:openloop_result}
\end{figure*}

\begin{figure}
    \centering
    \includegraphics[width=0.35\textwidth]{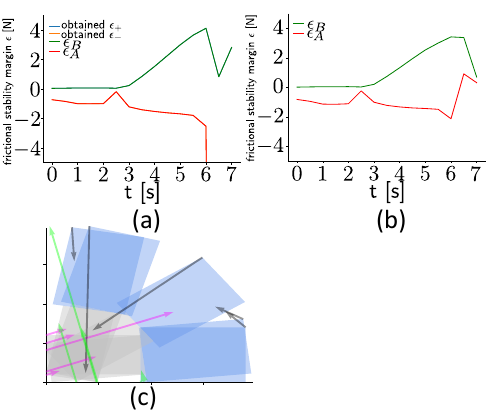} % 
    \caption{(a), (b): Trajectory of frictional stability margin of peg 1 based on uncertain mass obtained from CIBO, baseline optimization, respectively. Note that here we solve CIBO sequentially for each mode (i.e., hierarchical planning), instead of using the proposed mode-sequence-based optimization.  (c): Snapshots of pivoting motion for peg 1, obtained from CIBO considering uncertain mass.}
    \label{fig:openloop_result_peg1}
\end{figure}

\begin{figure*}
    \centering
    \includegraphics[width=0.98\textwidth]{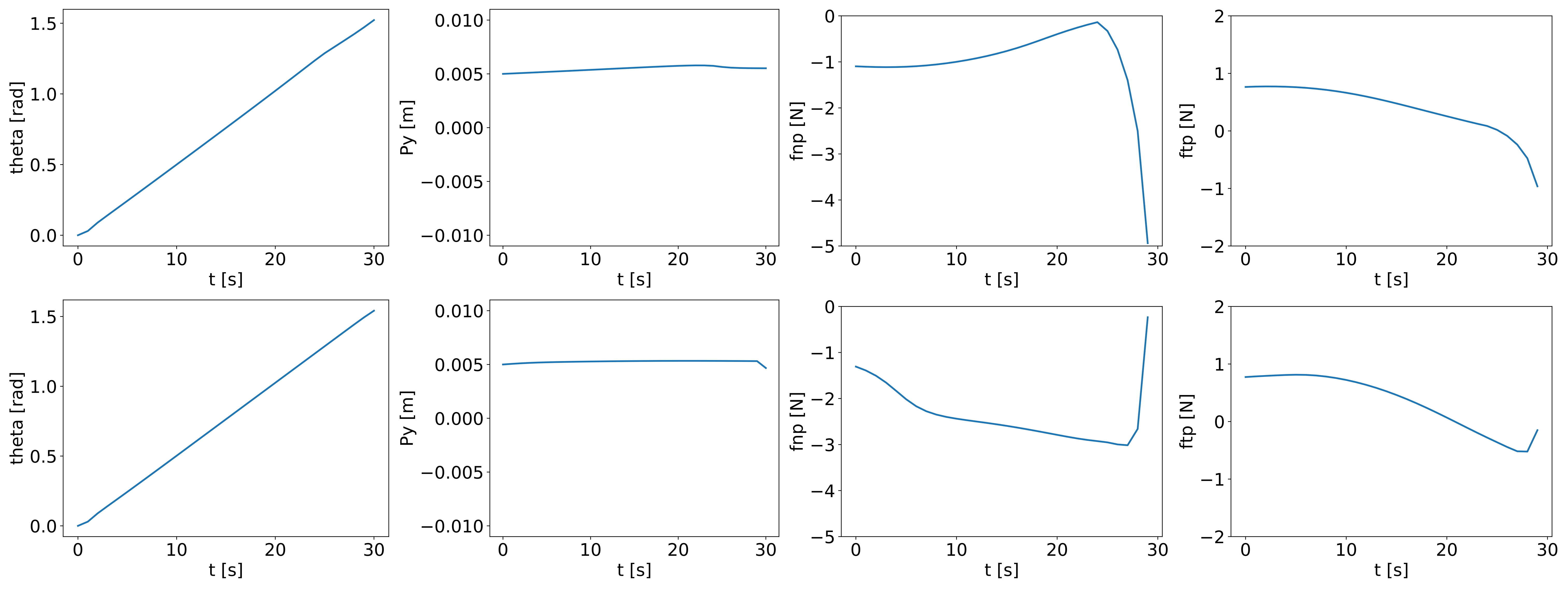} % 
    \caption{We show the time history of object angle, finger position, and contact forces from a manipulator during pivoting of gear 1. The top row shows the result using CIBO \eq{kkt_convertion} considering CoM uncertainty and the bottom one shows the result using \eq{equation_control} (i.e., it does not consider robustness criteria in the formulation explicitly.). The top row results and the bottom row results are used in visualizing the stability margin in \fig{fig:openloop_result} (d), (c), respectively.}
    \label{fig:benchmark_vs_robust_com}
\end{figure*}

% \begin{figure*}
%     \centering
%     \includegraphics[width=1\textwidth]{Figures/fc2_cropped.pdf} % 
%     \caption{Time history of frictional stability margin with time history of control inputs $u$. (a): The trajectory obtained from the baseline optimization is shown with stability margin considering uncertain mass. (b): The trajectory obtained from our proposed optimization is shown with stability margin considering uncertain mass. (c): The trajectory obtained from the baseline optimization is shown with stability margin considering uncertain CoM location. (d): The trajectory obtained from our proposed optimization is shown with stability margin considering uncertain CoM location. (e): Snapshots of motion of pivoting for gear 1 obtained from our proposed optimization considering uncertain CoM location. (e): Snapshots of motion of pivoting for peg 2 obtained from our proposed optimization considering uncertain mass. (g), (h): Time history of stability margin of (f) obtained from our proposed optimization, the benchmark optimization, respectively. Our proposed could generate more robust trajectories than baseline optimization. In (a), at $t=0$ s, $f_{nB}$ is almost zero so that the stability margin obtained from \eq{fnb_cond_mass} would be zero. In contrast, in (b), our proposed optimization could realize non-zero $f_{nB}$ as shown as a red arrow in (b). In (c), at $t=25$ s,  the trajectory has smallest stability margin. In (d), our bilevel optimization is able to increase the stability margin at $t=25$ s. }
%     \label{fig:openloop_result}
% \end{figure*}

\begin{table}[t]
    \caption{{Worst-case stability margin over the control horizon obtained from optimization for gear 1. Note that the stability margin for the solution of the benchmark optimization is analytically calculated.}}
    \centering
    \begin{tabular}{c|c|c}
     & $\epsilon_{+}^*$, $\epsilon_{-}^* $  [N]& $r_{+}^*$, $r_{-}^*$ [mm]\\
         \hline\hline Benchmark optimization \eq{equation_control} & 0.10, 0.66 & 1.5, 0.85\\
         \hline Ours  \eq{kkt_convertion} with mass uncertainty & 0.34, 0.50 & N/A \\
         \hline Ours  \eq{kkt_convertion} with CoM uncertainty & N/A & 3.43, 2.70
    \end{tabular}
    \label{epsilon}
\end{table}

\begin{table}[t]
    \caption{{Obtained worst stability margins over the time horizons from optimization for peg 1. Note that the stability margin for the solution of the benchmark optimization is analytically calculated.}}
    \centering
    \begin{tabular}{c|c|c}
     & $\epsilon_{+}^*$, $\epsilon_{-}^* $  [N]& $r_{+}^*$, $r_{-}^*$ [mm]\\
         \hline\hline Benchmark optimization \eq{equation_control} & 0.035, 0.018 & 31, 0\\
         \hline Ours  \eq{kkt_convertion} with mass uncertainty & 0.050, 0.021 & N/A \\
         \hline Ours  \eq{kkt_convertion} with CoM uncertainty & N/A & 38, 0
    \end{tabular}
    \label{epsilon_peg}
\end{table}

\subsection{Experiment Setup}
We implement our method in Python using IPOPT solver \cite{80fe29bf9dc245ffa5c8bd7b3eee2902} with \pyrobocop\ wrapper \cite{9812069}. 
We use HSL MA86 \cite{hsl2007collection} as a linear solver for IPOPT. 
The optimization problem is implemented on a computer with Intel i7-12800K.

We demonstrate our algorithm on several different objects, as detailed in \tab{parameter_table}. During optimization, we set 
$Q=\text{diag}(0.1, 0), R=\text{diag}(0.01, 0.01)$. We use $\alpha= 1$ when we run \eq{kkt_convertion}. We set $x_s = [0, \frac{w}{4}]^\top, \theta_g = \frac{\pi}{2}$. Note that we only enforce terminal constraints for convex shape objects. For non-convex shape objects, we do not enforce terminal constraints since the peg cannot achieve $\theta_N = \frac{\pi}{2}$ unless we consider another contact mode (see \fig{fig:mode_concept}).
In \pyrobocop\ wrapper, we did warm-start for the state at $k=0, N$ by setting initial and terminal states as initial guesses. We did not explicitly conduct a warm-start for other decision variables and we set them to 0. 
%Regarding pegs, because the contact points change during pivoting, we decouple one trajectory optimization problem with complementarity constrains expressing the contact-noncontact configuration and solve it as a multi-stage optimization problem where each stage problem solves the case with a certain contact configuration.

%As shown in 

We use a Mitsubishi Electric Factory Automation (MELFA) RV-5AS-D Assista 6 DoF arm (see \fig{fig:pivoting_abstractfig}) for the experiments. The robot has a pose repeatability of $\pm 0.03$mm. The robot is equipped with Mitsubishi Electric F/T sensor 1F-FS001-W200 (see \fig{fig:pivoting_abstractfig}). To implement the computed force trajectory during manipulation, we use the default stiffness controller for the robot. By selecting an appropriate stiffness matrix \cite{stiffness_control}, we design a reference trajectory that would result in the desired interaction force required for manipulation~\cite{9838102, https://doi.org/10.48550/arxiv.2204.10447}. 
\revise{More specifically, we use the following relationship, $x^r_k=x_k+u_k K_s^{-1}$, where, $x_{k}$ and $f_k$ are the configuration trajectory and the force trajectory at time step $k$, respectively, obtained by CIBO as output. $K_s$ is the stiffness matrix for the robot which is appropriately chosen. At each time step $k$, we command $x^r_k$ as a reference trajectory of the robot's internal position controller.}
Note that this trajectory is implemented in open-loop and we do not design a controller to ensure that the computed force trajectory is precisely tracked during execution.

% \textcolor{red}{Explain MPC setting: Devesh could you work on this?}
\revise{For the MPC experiments discussed in Section~\ref{sec:error_recovery}, the object states are tracked using AprilTag \cite{apriltag_2011icra}. The robot states are tracked using the robot's joint encoders. The contact states at contact $A, B, P$ in \fig{fig:mechanics_pivoting_eq} are estimated using the object state, the robot state, and the known geometry of the object. }
% To track the reference force profile in open-loop fashion, we use a stiffness controller which is designed using the default stiffness controller \cite{stiffness_control} of the robot, as illustrated in \fig{fig:control_block}. \devesh{Yuki : please be careful in the choice of words. We are not tracking the force profile. We are simply designing a reference position trajectory so that we can perform indirect force control. Note that the there is no feedback loop tracking the force trajectory.}
% We test our method on several different objects listed in \tab{parameter_table}.
% talk stiffness control, robot specification, using a Mitsubishi F/T sensor $1$F-FS$001$-W$200$

% \begin{figure}
%     \centering
%     \includegraphics[width=0.45\textwidth]{Figures/control_block.png} % 
%     \caption{A schematic showing the implemented control block diagram. }
%     \label{fig:control_block}
% \end{figure}

\subsection{CIBO for Uncertain Mass and CoM Parameters}
% We first show the results of our proposed controller for gear 1 in \fig{fig:openloop_result}. 

\fig{fig:openloop_result} shows the trajectory of frictional stability margin of gear 1 obtained from the proposed robust CIBO considering uncertain mass and uncertain CoM location, and the benchmark optimization. Overall, CIBO could generate more robust trajectories. 
For example, at $t=0$ s, $f_{nB}$ in (a) is almost zero so that the stability margin obtained from \eq{fnb_cond_mass} is almost zero. In contrast,  CIBO could realize non-zero $f_{nB}$ as shown as a red arrow in (b). 
In (d), to increase the stability margin, the finger position $\revise{P}_{y}^{\revise{O}}$ moves on the face of gear 1 so that the controller can increase the stability margin more than the benchmark optimization. This would not happen if we do not consider complementarity constraints \eq{slippingP}. 
% In fact, in this problem setting, we only 
% finds infeasible solutions if we get rid of \eq{lippingP} and ensure that point $P$ is always sticking. 
Also, our obtained $\epsilon_{+}, \epsilon_{-}, r_+, r_-$ follows bounds of stability margin. It means that CIBO can successfully design a controller that maximizes the worst stability margin given the best stability margin for each time-step.  
% 

% In (a), at $t=0$ s, $f_{nB}$ is almost zero so that the stability margin obtained from \eq{fnb_cond_mass} would be zero. In contrast, in (b), our proposed optimization could realize non-zero $f_{nB}$ as shown as a red arrow in (b). In (c), at $t=25$ s,  the trajectory has smallest stability margin. In (d), our bilevel optimization is able to increase the stability margin at $t=25$ s.

% \fig{fig:benchmark_vs_robust_com} also explains why the proposed bilevel optimization achieves the robust solution. 

\fig{fig:benchmark_vs_robust_com} shows that both the benchmark and CIBO actually change the finger position $\revise{P}_{y}^{\revise{O}}$ by considering complementarity constraints \eq{slippingP}. In fact, we observed that at $t=25$ s, $\revise{P}_{y}^{\revise{O}}$ in both results moves to the negative value to maintain the stability of the object.
In practice, we are unable to find any feasible solutions with fixed $\revise{P}_{y}^{\revise{O}}$, instead of using \eq{slippingP}. Thus, \eq{slippingP} is critically important to find a feasible solution. 

Next, we discuss how much CIBO improves the worst-case stability margin.
% Regarding optimality of robustness, 
The trajectories of $f_{nP}$ in \fig{fig:benchmark_vs_robust_com} show that the magnitude of $f_{nP}$ from CIBO increase at $t=25$ s to improve the worst-case stability margin. On the other hand, $f_{nP}$ from the benchmark optimization does not increase at $t=25$ s. Hence, we verify that by increasing normal force, the robot could successfully robustify the pivoting manipulation. 
This result can be also understood in \fig{fig:openloop_result} (c) and (d) where the stability margin in (d) at $t=25$ s is larger than that in (c), as discussed above.

\tab{epsilon} and \tab{epsilon_peg} summarize the computed stability margin from \fig{fig:openloop_result}. In \tab{epsilon}, for the case where CIBO considers uncertainty of mass, we observe that the value of $\epsilon_-^*$ from CIBO is smaller than that from the benchmark optimization although the sum of the stability margin $\epsilon_+^* + \epsilon_-^* $ from CIBO is greater than that from the benchmark optimization. This result means that CIBO can actually improve the worst-case performance by sacrificing the general performance of the controller. Regarding the case where we consider the uncertain CoM location, CIBO outperforms the benchmark trajectory optimization in both $r_+^*, r_-^*$. For peg 1, the bilevel optimizer without using mode sequence-based optimization (i.e., hierarchical optimization) finds trajectories that have larger stability margins for both uncertain mass and CoM location as shown in \tab{epsilon_peg}. The trajectory of stability margin obtained from CIBO considering mass uncertainty is illustrated in \fig{fig:openloop_result_peg1}.
We discuss the results using CIBO with mode-sequence based optimization in Sec~\ref{sec:mode_seq_result}.
%how much our proposed controller is robust under uncertain parameters. 

\subsection{CIBO with Different Manipulator Initial State}

\begin{figure}
    \centering    \includegraphics[width=0.48\textwidth]{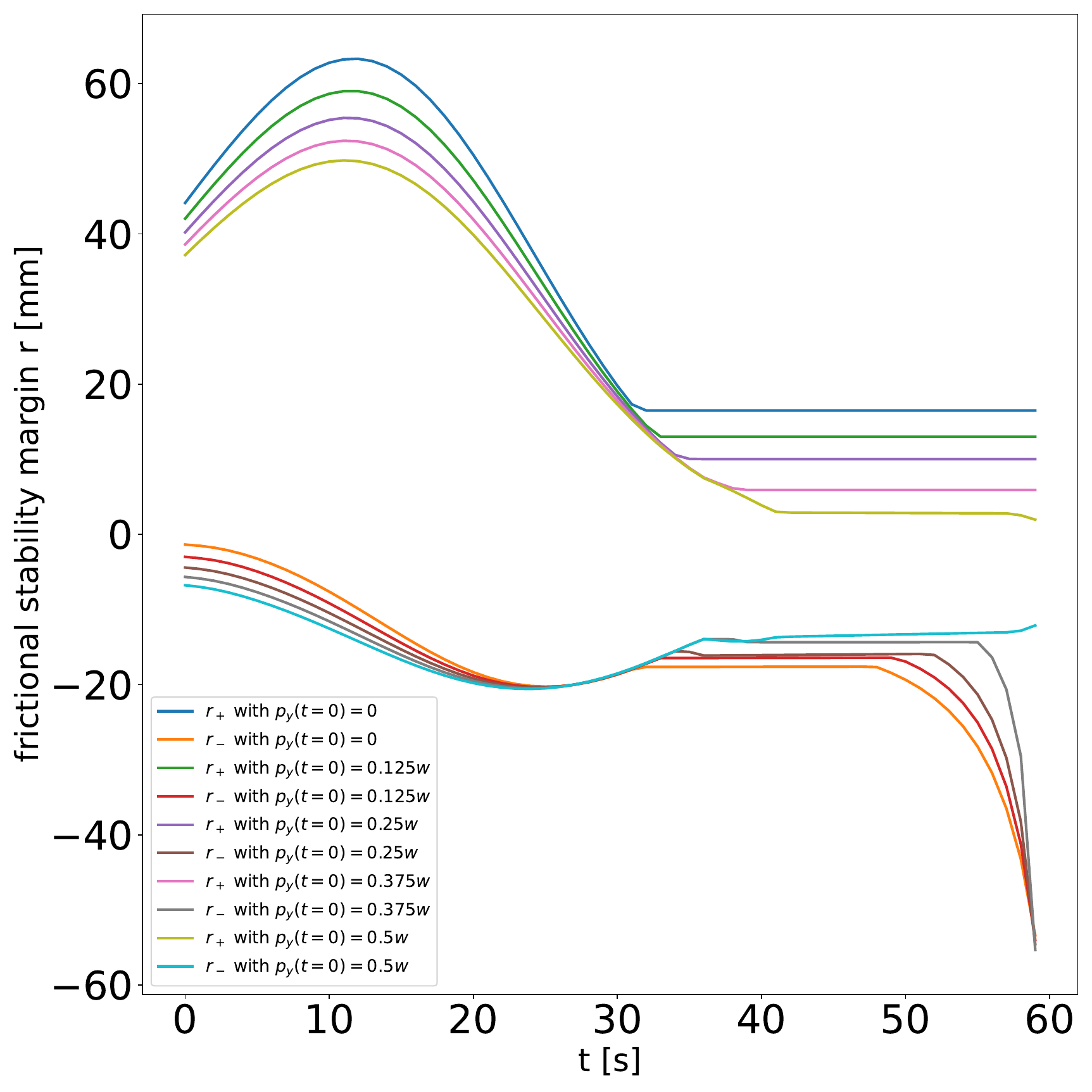} % 
    \caption{Time history of frictional stability margin considering CoM location with different initial manipulator position $\revise{P}_{y}^{\revise{O}}(t=0)$.}
    \label{fig:diiferent_py_t0}
\end{figure}

\begin{table}[t]
    \caption{{Computed worst-case stability margin considering uncertain CoM location with different $\revise{P}_{y}^{\revise{O}}$ at $t=0$ over the control horizon obtained from optimization for gear 1.}}
    \centering
    \begin{tabular}{c|c}
     & $r_{+}^*$, $r_{-}^*$ [mm]\\
         \hline\hline Ours  with $\revise{P}_{y}^{\revise{O}}(t=0) = 0$  & 16.47, 1.36\\
         % \hline Ours  \eq{kkt_convertion} with mass uncertainty & 0.34, 0.50 & N/A \\
         \hline Ours  with $\revise{P}_{y}^{\revise{O}}(t=0) = 0.125w$  & 12.99, 2.98\\
         \hline Ours  with $\revise{P}_{y}^{\revise{O}}(t=0) = 0.25w$  & 10.00, 4.41\\
         \hline Ours  with $\revise{P}_{y}^{\revise{O}}(t=0) = 0.375w$  & 5.94, 5.67\\
         \hline Ours  with $\revise{P}_{y}^{\revise{O}}(t=0) = 0.5w$  & 1.94, 6.77
    \end{tabular}
    \label{py_table}
\end{table}

 %  CIBO is conditioned with states at $t = 0$ (i.e., $x_0$), which is true for other trajectory optimization frameworks.
% We discuss how initial states have an effect on the stability margin during pivoting. 
We believe that the efficiency of the optimization depends on the initial location of the manipulator finger. This is because the stability margin depends on the manipulation finger location, which is partially governed by its location at $t = 0$. Thus we present some results by randomizing over the manipulator finger location at $t = 0$.
We sample initial state of finger position $\revise{P}_{y}^{\revise{O}}(t=0)$ from a discrete uniform distribution with the range of $\revise{P}_{y}^{\revise{O}}(t=0) \in [-0.5w, -0.375w, -0.25w, -0.125w, \ldots, 0.5w]$. 
% As illustrated in \fig{fig:mechanics_pivoting_eq}, $\revise{P}_{y}^{\revise{O}}(t=0)=0.5w$ indicates that the contact location of finger tip is at $C_1$. 
Then we run CIBO considering CoM location uncertainty.

\fig{fig:diiferent_py_t0} illustrates the time history of stability margin with different $\revise{P}_{y}^{\revise{O}}(t=0)$. CIBO is not able to find feasible solutions with $\revise{P}_{y}^{\revise{O}}(t=0) < 0$. It makes sense since there may not be enough moment for the desired motion if $\revise{P}_{y}^{\revise{O}}(t=0) < 0$.

\fig{fig:diiferent_py_t0} shows that different $\revise{P}_{y}^{\revise{O}}(t=0)$ leads to different stability margin over the time horizon. 
\tab{py_table} summarizes the worst-case stability margin over the trajectory obtained from \fig{fig:diiferent_py_t0}. \tab{py_table} also shows that the worst-case stability margin is different with different $\revise{P}_{y}^{\revise{O}}(t=0)$. 
Finding a good $\revise{P}_{y}^{\revise{O}}(t=0)$ is not trivial and it requires domain knowledge. Thus, ideally, we should formulate CIBO where $\revise{P}_{y}^{\revise{O}}(t=0)$ is also a decision variable so that the solver can optimize the trajectory over $\revise{P}_{y}^{\revise{O}}(t=0)$ as well. 

Since CIBO is non-convex optimization, it is still possible that a feasible solution exists for $\revise{P}_{y}^{\revise{O}}(t=0) < 0$. However, we can at least argue that it is much more difficult to find a feasible solution with $\revise{P}_{y}^{\revise{O}}(t=0) < 0$ than that with $\revise{P}_{y}^{\revise{O}}(t=0) \geq 0$.

% Another important limitation regarding the complexity of dynamics is that CIBO is conditioned with states at $t = 0$ (i.e., $x_0$), which is true for other trajectory optimization frameworks. During the implementation of CIBO, we observed that it is not trivial to find "good" $x_0$ and the behavior of CIBO dramatically changes as  $x_0$ changes. For example, for a certain $x_0$, CIBO is able to find a solution while for another $x_0$, CIBO is not able to find a solution. Finding a good $x_0$ is not trivial at all and it requires domain knowledge. Thus, ideally, we should formulate CIBO where $x_0$ is also a decision variable so that the solver can optimize the trajectory over $x_0$ as well. 

\subsection{\revise{CIBO for Uncertain CoM parameters with Different Mass and Friction of Object}}
\revise{We first study how stability margin with uncertain CoM location changes with different mass parameters. We sample the mass of the object from a discrete uniform distribution with range of $m \in [0.1, 0.12, 0.14, 0.16, 0.18, 0.2]$ kg. Then we run CIBO considering CoM location uncertainty.}

\begin{figure}[t]
    \centering
    \begin{subfigure}{0.5\textwidth}
        \centering
        \includegraphics[width=\textwidth]{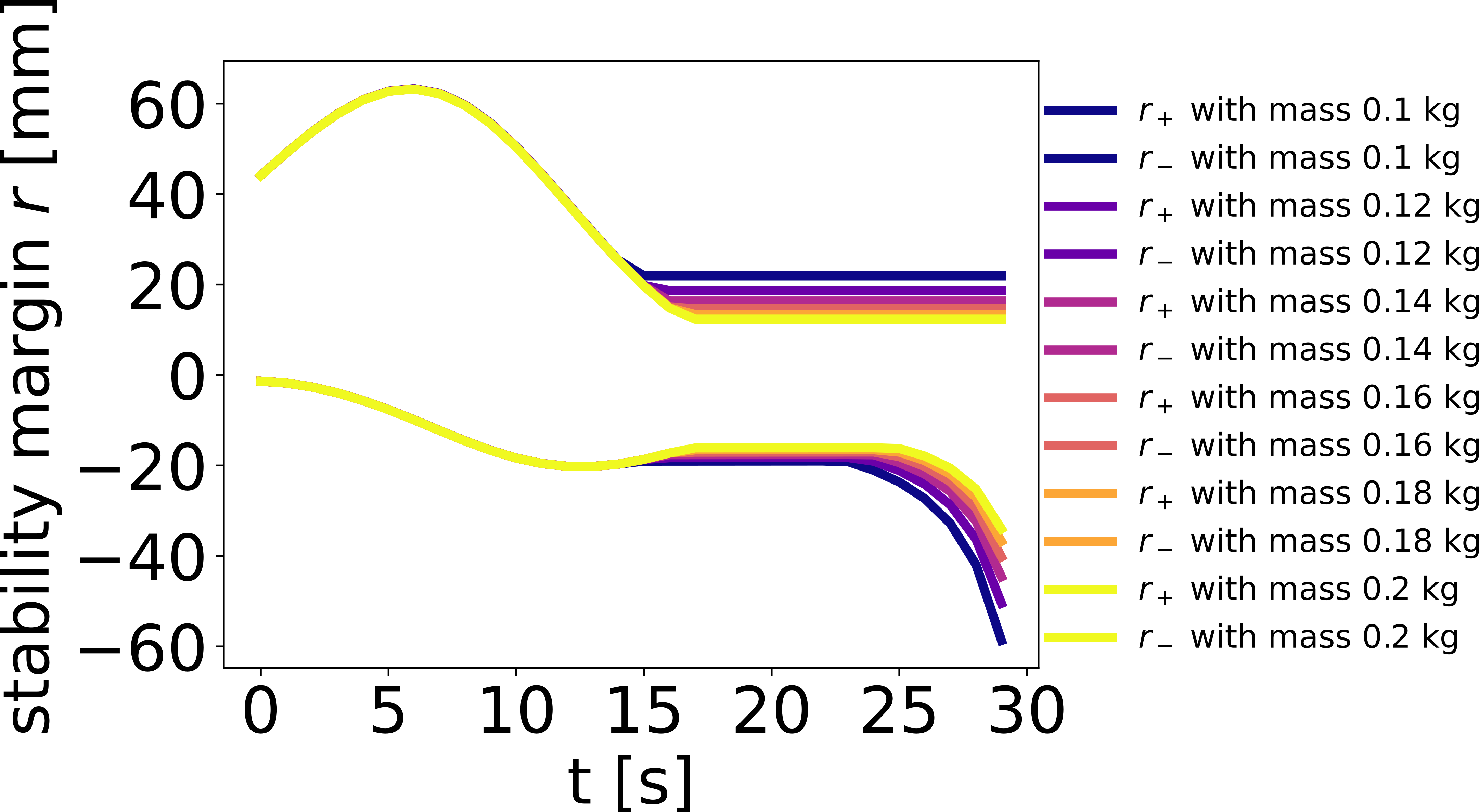} % Replace 'image1' with your image file name
        \caption{\revise{Time history of stability margin considering CoM location with different mass. The trajectory with the same color means that the same mass is used in the CIBO. The trajectories where $r>0$ are the trajectories of $r_+$ and the the trajectories where $r<0$ are the trajectories of $r_-$.}}
        \label{fig:sub1_cibo_mass}
    \end{subfigure}
    % \hfill
    % \begin{subfigure}{0.241\textwidth}
    %     \centering
    %     \includegraphics[width=\textwidth]{Figures/q_2_pivot_com_uncertainty_different_mass.pdf} % Replace 'image2' with your image file name
    %     \caption{Subfigure 2}
    %     \label{fig:sub2}
    % \end{subfigure}

        \hfill

    % \vspace{0.5cm} % Adjust vertical space between rows

    \begin{subfigure}{0.241\textwidth}
        \centering
        \includegraphics[width=\textwidth]{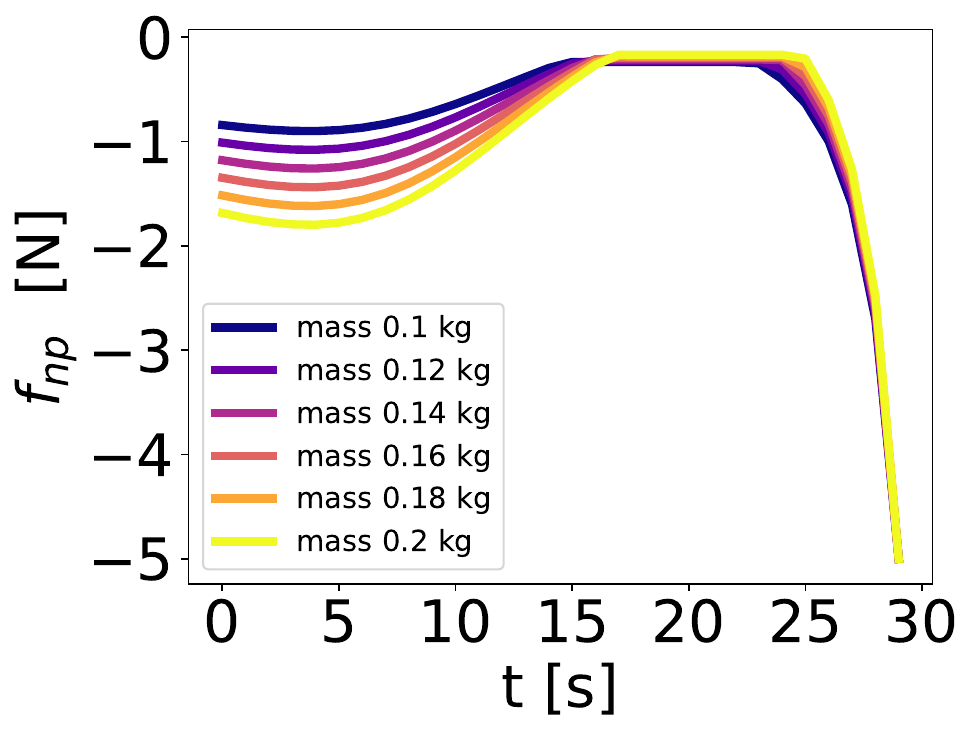} % Replace 'image3' with your image file name
        \caption{\revise{Time history of $f_{nP}^O$}}
        \label{fig:sub3_cibo_mass}
    \end{subfigure}
    \hfill
    \begin{subfigure}{0.241\textwidth}
        \centering
        \includegraphics[width=\textwidth]{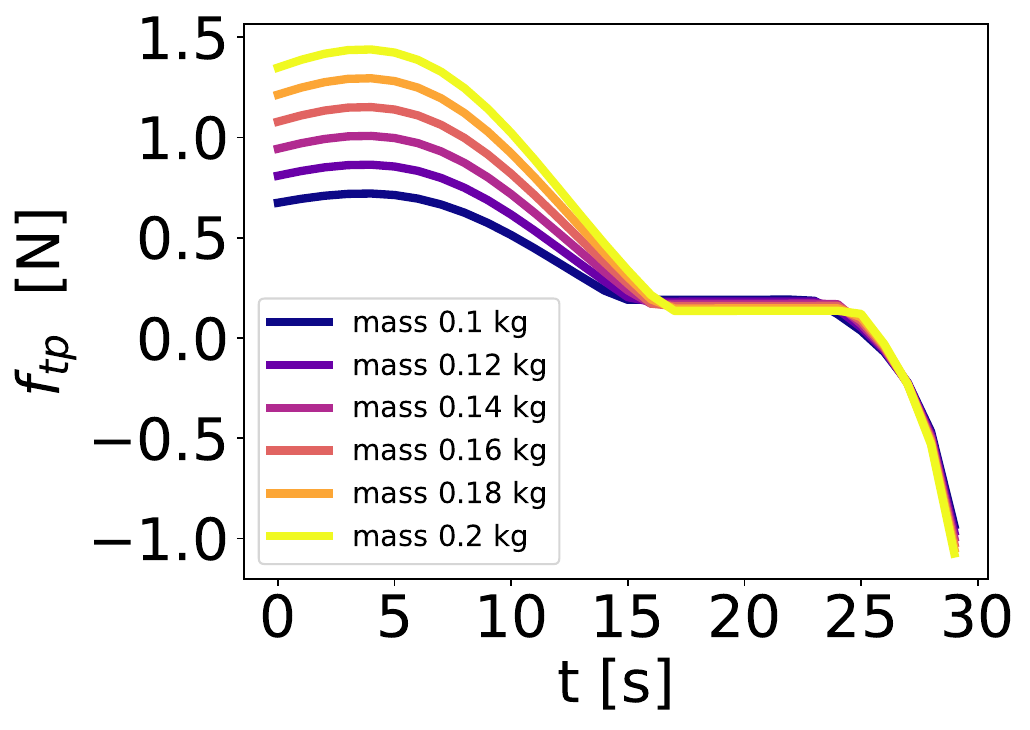} % Replace 'image4' with your image file name
        \caption{\revise{Time history of $f_{tP}^O$}}
        \label{fig:sub4_cibo_mass}
    \end{subfigure}
    \caption{\revise{Results of CIBO considering CoM location with different mass.}}
    \label{fig:subplot_cibo_mass}
\end{figure}

\revise{\fig{fig:subplot_cibo_mass} shows the time history of stability margin and contact forces over the time horizon. 
% We observe that at $t \in [0, 15]$ s, the stability margin with different masses shows the same value.
For this analysis, the projection of CoM lies on the contact $B$ (i.e., $C_x^B = 0$.) at $t = 15$ s. At $t \in [0, 15]$  s (i.e., $C_x^B > 0$), the robot has to execute the contact forces to support the object against gravity. In fact, \fig{fig:sub3_cibo_mass} and \fig{fig:sub4_cibo_mass} show that the contact forces increase as mass increases. Since other parameters of the system are the same, the CIBO designs the trajectory whose stability margin is the same with different mass by changing the contact forces from the robot. At $t \in [15, 30]$ s,  the upper-bound of stability margin $r_+$ shows the larger value with the lighter object, and the lower-bound of stability margin $r_-$ also shows the larger value with the lighter mass of the object. This makes sense because as the object becomes lighter, the system allows for a longer moment arm $r$ in \revise{quasi-static} equilibrium.}
% since $C_x^B < 0$,

\begin{figure}[t]
    \centering
    \begin{subfigure}{0.5\textwidth}
        \centering
        \includegraphics[width=\textwidth]{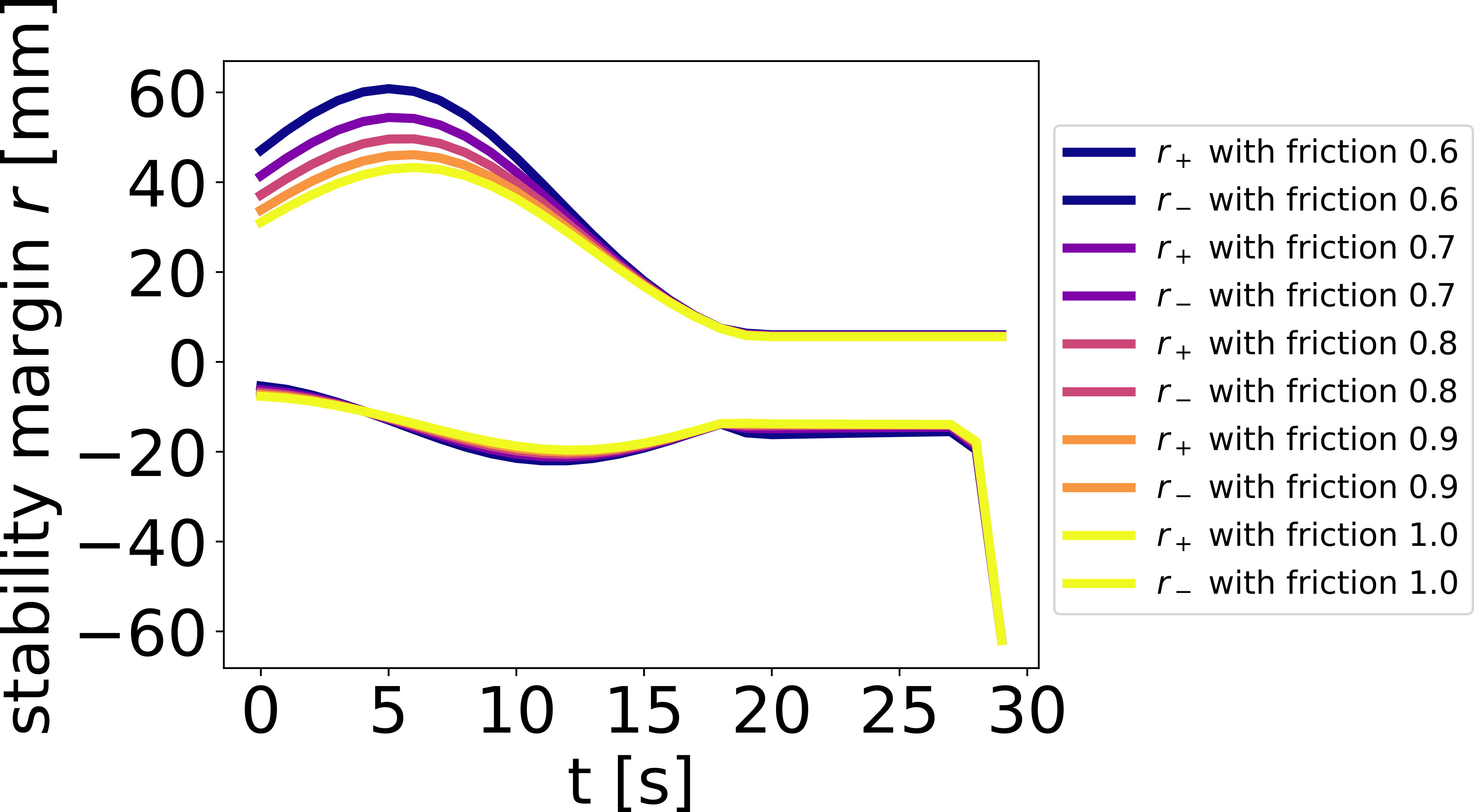} % Replace 'image1' with your image file name
        \caption{\revise{Time history of stability margin considering CoM location with different friction at $P$. The trajectory with the same color means that the same mass is used in the CIBO. The trajectories where $r>0$ are the trajectories of $r_+$ and the the trajectories where $r<0$ are the trajectories of $r_-$.}}
        \label{fig:sub1_cibo_friction}
    \end{subfigure}
        \hfill

    \begin{subfigure}{0.241\textwidth}
        \centering
        \includegraphics[width=\textwidth]{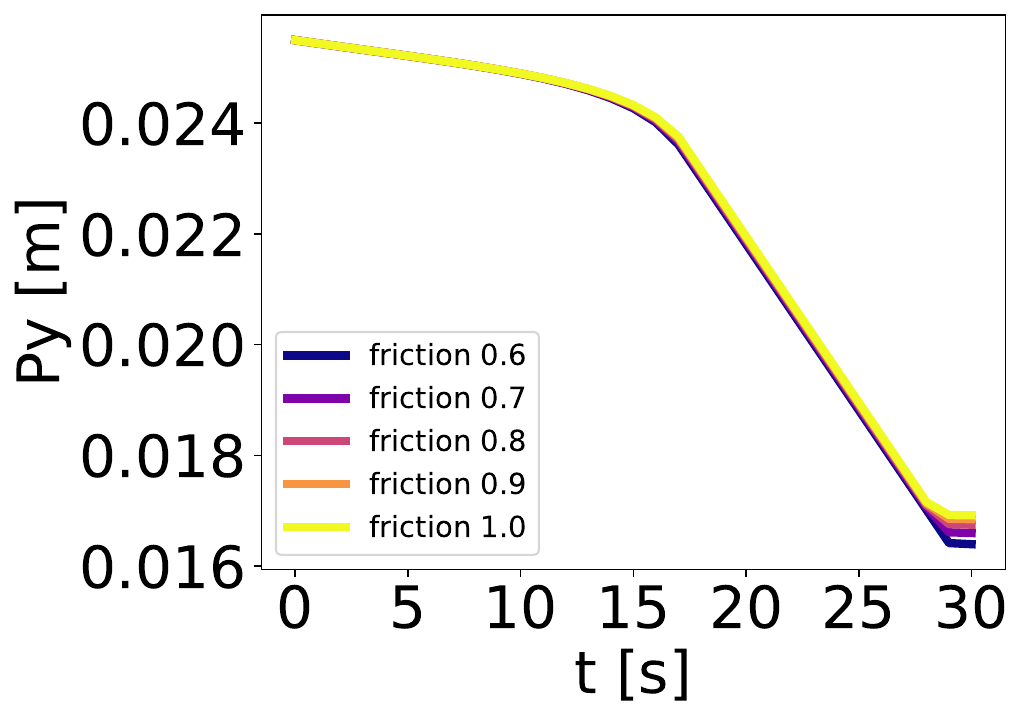} 
        \caption{\revise{Time history of $\revise{P}_{y}^{\revise{O}}$}}
        \label{fig:sub3_cibo_friction}
    \end{subfigure}
    \hfill
    \begin{subfigure}{0.241\textwidth}
        \centering
        \includegraphics[width=\textwidth]{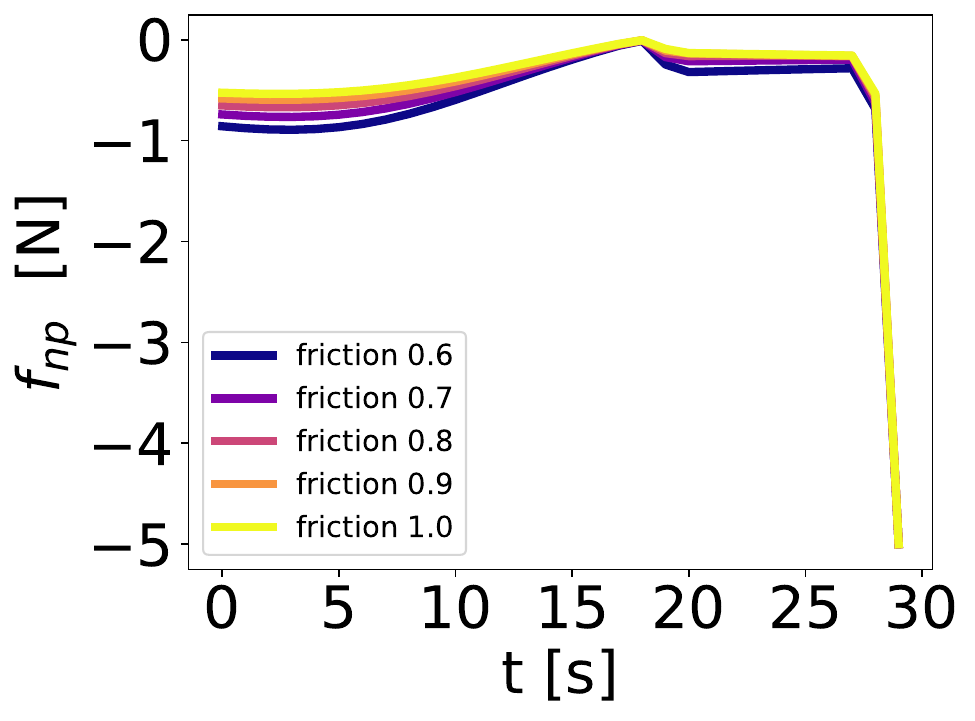} 
        \caption{\revise{Time history of $f_{nP}^O$}}
        \label{fig:sub4_cibo_friction}
    \end{subfigure}
    \caption{\revise{Results of CIBO considering CoM location with different friction.}}
    \label{fig:subplot_cibo_friction}
\end{figure}

\revise{Second, we study how stability margin with uncertain CoM location changes with different coefficients of friction between the object and the robot finger (i.e., $\mu_P$ at contact $P$ in \fig{fig:mechanics_pivoting_eq}). We sample the friction of the object from a discrete uniform distribution with a range of $\mu_P \in [0.6, 0.7, 0.8, 0.9, 1.0]$. Then we run CIBO considering CoM location uncertainty.}

\revise{\fig{fig:subplot_cibo_friction} shows the time history of stability margin, finger contact location $\revise{P}_{y}^{\revise{O}}$, and the contact normal force $f_{nP}^O$ over the time horizon. 
We observe that the different friction leads to different trajectories of the stability margin. In particular, we observe that the CIBO considering the lower $\mu_P$ results in a larger $r_+$. As \fig{fig:sub3_cibo_friction}, the finger keeps moving during the manipulation to complete the pivoting. It means that the complementarity constraints at $P$ \eq{slippingP} are always equality constraints like \eq{slipping_friction_cone}, $f_{tP}^O = \mu_P f_{nP}^O$. With the small $\mu_P$, the robot can execute the large $f_{nP}^O$ with the small $f_{tP}^O$, which is beneficial at  $t \in [0, 18]$ s to avoid losing the contact $A$,  before the projection of CoM lies on the contact $B$. 
% while we observe almost the same value of $r_i$ over the time horizon. 
% We observe that at $t \in [0, 15]$ s, the stability margin with different masses shows the same value.
}

\subsection{CIBO for Uncertain Friction Parameters}
\begin{figure}
     \centering
     \begin{subfigure}[b]{0.241\textwidth}
         \centering
         \includegraphics[width=\textwidth]{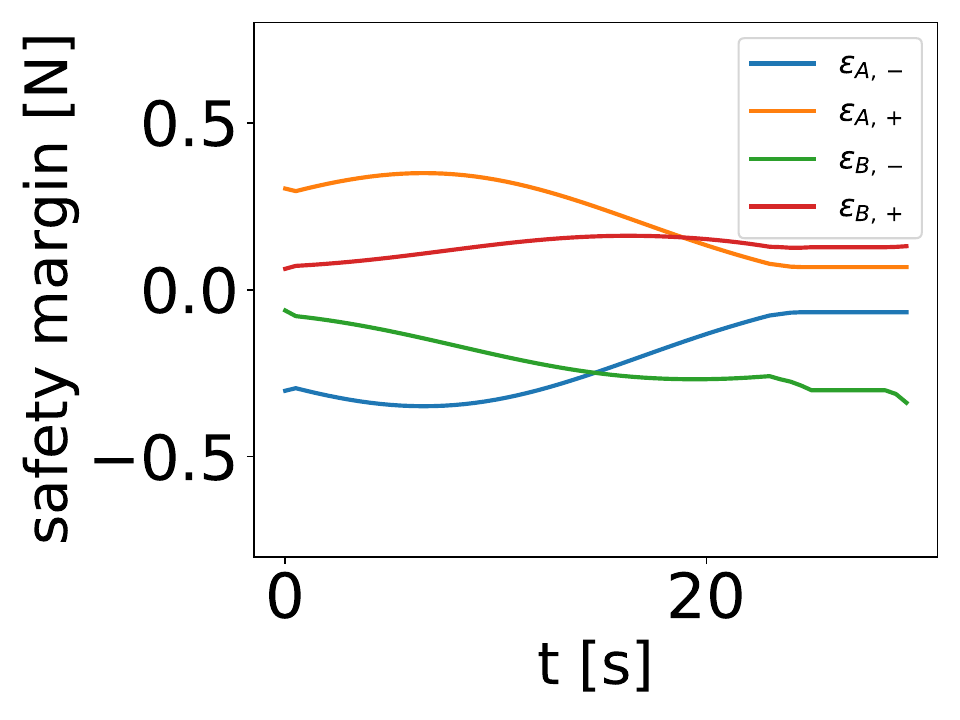}
         \caption{}
         \label{fig:y equals x}
     \end{subfigure}
     \hfill
     \begin{subfigure}[b]{0.241\textwidth}
         \centering
         \includegraphics[width=\textwidth]{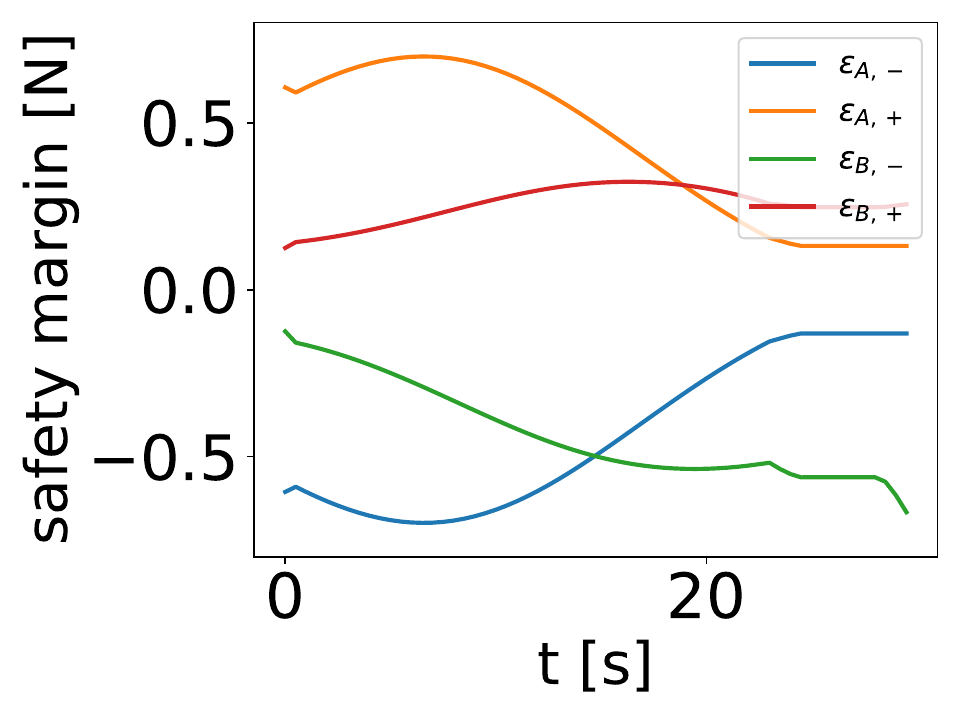}
         \caption{}
         \label{fig:three sin x}
     \end{subfigure}
        \caption{Trajectory of frictional stability margin of (a) gear 1 and (b) gear 3, based on uncertain friction obtained from CIBO \eq{eq:bilvel_friction2}, respectively. }
        \label{fig:result_friction}
\end{figure}

\fig{fig:result_friction} shows the time history of frictional stability margin of gear 1 and gear 3 using \eq{eq:bilvel_friction2}. CIBO could successfully design an optimal open-loop trajectory by improving the worst-case performance of stability margin.
We observe that \fig{fig:result_friction} (b) shows a larger stability margin compared to  (a). This result makes sense since in (b), we consider gear 3 whose weight is heavier than the weight of gear 1 and thus we get stability margins which are bigger than those obtained for (a). 

% \begin{figure}
% \centering\includegraphics[width=0.49\textwidth]{Figures/friction_result.png} % 
%     \caption{Trajectory of frictional stability margin of (a) gear 1 and (b) gear 3, based on uncertain friction obtained from our proposed bilevel optimization \eq{eq:bilvel_friction2}, respectively. }
%     \label{fig:result_friction}
% \end{figure}

\subsection{\revise{CIBO for Uncertain Finger Contact Location}}\label{sec:result_bilevel_opt_uncertain_finger_contact_location}
\begin{figure}
     \centering
     \begin{subfigure}[b]{0.241\textwidth}
         \centering
         \includegraphics[width=\textwidth]{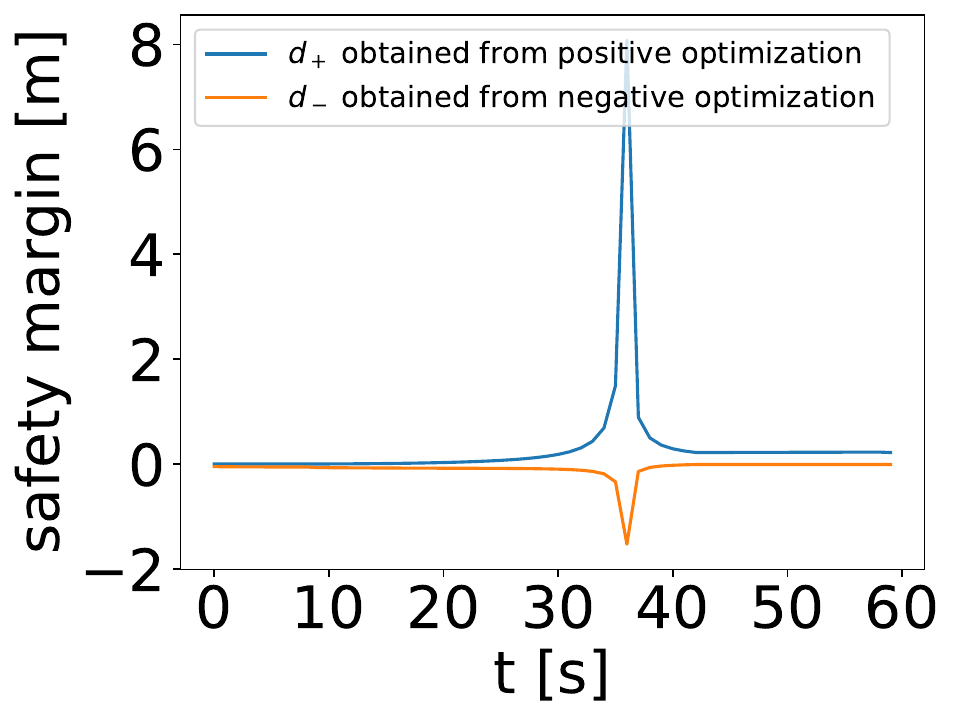}
         \caption{}
         \label{fig:finger_contact_location_margin}
     \end{subfigure}
     \hfill
     \begin{subfigure}[b]{0.241\textwidth}
         \centering
         \includegraphics[width=\textwidth]{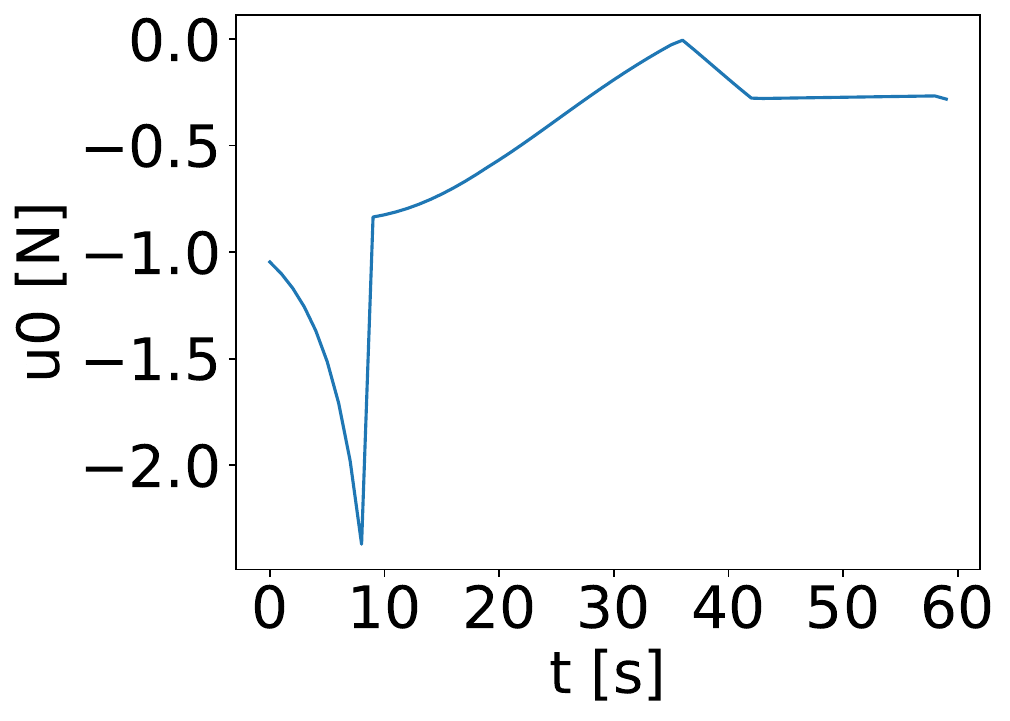}
         \caption{}
         \label{fig:contact_force}
     \end{subfigure}
        \caption{
        \revise{
        We consider CIBO with uncertain finger contact location. (a): Time history of frictional stability margin. (b) Time history of normal force at the finger. }
        }
        \label{fig:finger_contact_location}
\end{figure}

\revise{In this section, we present results for pivoting manipulation under uncertain finger contact location. \fig{fig:finger_contact_location} shows the time history of the stability margin of gear 2 using \eq{kkt_convertion}. Our CIBO could successfully design a controller for an uncertain contact location. Also, \fig{fig:finger_contact_location} shows that stability margin has a quite large value at $t = 37$ s. At $t = 37$ s, the controller makes the finger move with zero normal force, resulting in a large stability margin as we explain in Sec~\ref{sec:stability_margin_finger_contact_location}. }

% \begin{figure}
%     \centering    \includegraphics[width=0.3\textwidth]{Figures/y_16_pivot_different_noise_finger_contact_location.pdf} % 
%     \caption{Time history of frictional stability margin considering finger contact location.}
%     \label{fig:finger_contact_location}
% \end{figure}

\subsection{CIBO over Mode Sequences for Non-Convex Objects}\label{sec:mode_seq_result}
\begin{figure*}
    \centering
    \includegraphics[width=0.99\textwidth]{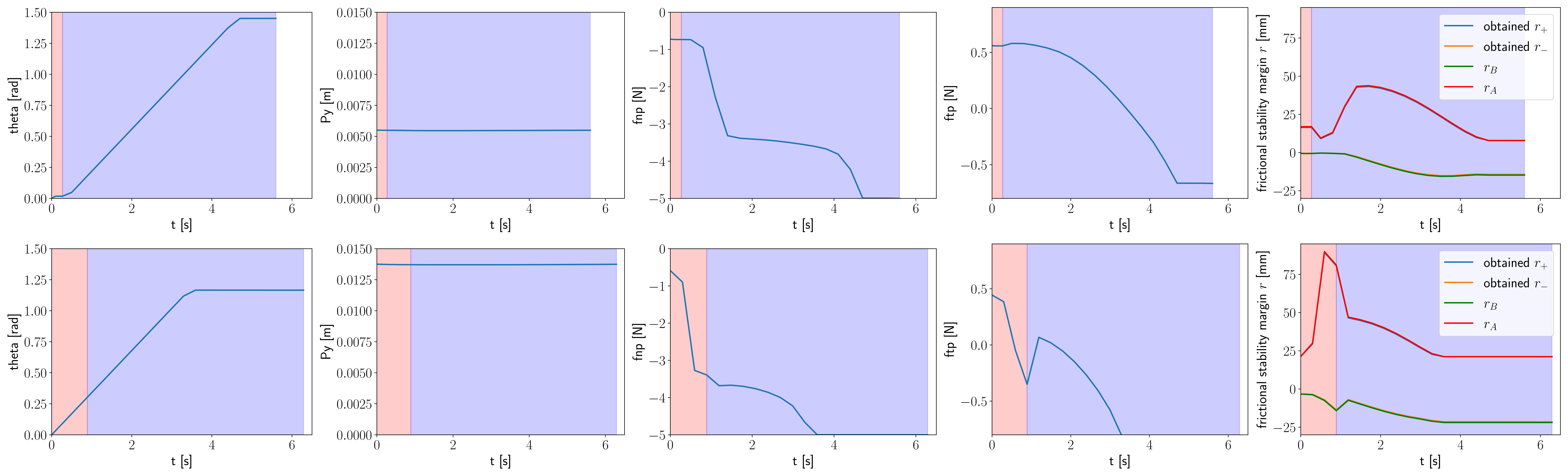} % 
    \caption{We show the time history of object angle, finger position, contact forces from a manipulator, and frictional stability margins. The top row shows the result with peg 2 and the bottom one shows the result with peg 3. The pink and blue shade regions represent that the system follows mode 1 and mode 2, respectively.}
    \label{fig:mode_result}
\end{figure*}

% what we want to say:
% 1. using mode-based opt, you can don't need to tune parameters anymore
% 2. maybe you can show the better margin
% 3. highlight the change of mode with different shapes of objects 
In this section, we present results for objects with non-convex geometry using the mode-based optimization presented in Section~\ref{subsec:mode_based_optimization}.
\fig{fig:mode_result} shows the time history of states, control inputs, and frictional stability margins for pegs whose geometry are non-convex and the contact sets change over time. First of all, we can observe that CIBO in \eq{eq:mode_change} could successfully optimize the stability margin over trajectory while it optimizes the time duration of each mode. We observe that $\frac{T_1}{T_1 + T_2}$ (i.e., the ratio of mode 1 over the horizon) of peg 2 is much smaller than that of peg 3 since $\gamma$ (see \fig{fig:mode_concept} for the definition of $\gamma$) of peg 2 is smaller than that of peg 3 and thus, it spends less time in mode 1.   \fig{fig:mode_result} shows that $f_{tP}$ of peg 3 dramatically changes at $t = T_1$ s while that of peg 1 does not. 
% This result also makes sense because the shape of peg 3 is quite non-convex (\devesh{Yuki, can you please check--Is it right to compare the convexity of objects? Im not sure if we can use such words. please check}) and thus $f_{tP}$ needs to change accordingly once the contact mode switches from mode 1 to mode 2. 
In contrast, the shape of peg 2 has smaller $\gamma$ (i.e., less non-convex shape) and it can be regarded as a rectangle shape. Thus, the effect of contact mode is less, leading to a smaller change of $f_{tP}$ at $t = T_1$ s. 

In order to show that we can find solutions much more effortlessly using \eq{eq:mode_change} compared to two-stage optimization (that was earlier used in~\cite{9811812}), we sample 20 different $p_{y}$ at $t = 0$ s and count the number of times the benchmark two-stage optimization problem and the proposed optimization problem over the mode sequences \eq{eq:mode_change} can find feasible solution. We observed that the benchmark two-stage optimization problem found feasible solutions only 2 times while the mode-based optimization using \eq{eq:mode_change} was successfully able to find feasible 18 out of 20 times. Therefore, we verify that our proposed optimization problem enables to find solutions much more effortlessly. The benchmark method requires careful selection of parameters to ensure feasibility (as was explained in~\cite{9811812}).

% It is really worth noting that using \eq{eq:mode_change}, we can find solutions much more effortlessly. Before using REFER FIGURE, we had to manually indicate the initial and goal states of for mode 1

\subsection{\revise{CIBO for Uncertain Mass on a Slope}}
\begin{figure}[t]
     \centering
     \begin{subfigure}[b]{0.241\textwidth}
         \centering
         \includegraphics[width=\textwidth]{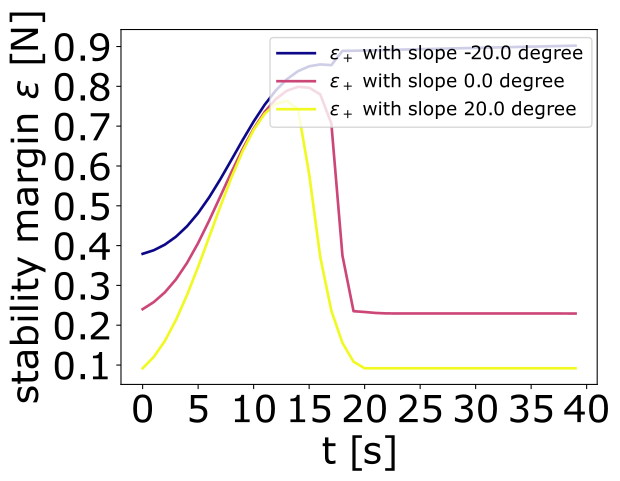}
         \caption{}
         \label{fig:mass_slope_uncertainty_positive}
     \end{subfigure}
     \hfill
     \begin{subfigure}[b]{0.241\textwidth}
         \centering
         \includegraphics[width=\textwidth]{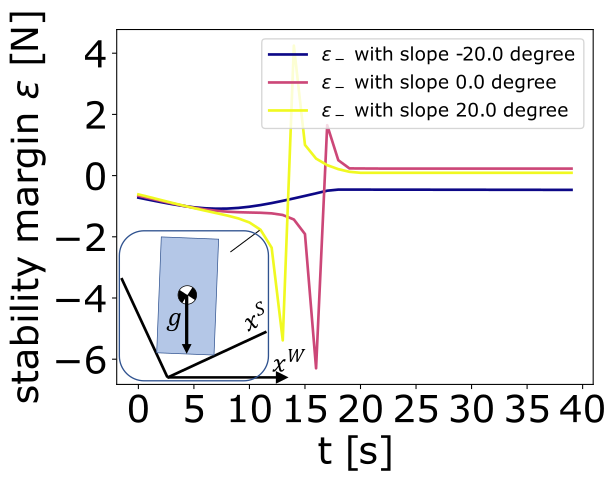}
         \caption{}
         \label{fig:mass_slope_uncertainty_negative}
     \end{subfigure}
        \caption{
        \revise{
        We consider CIBO with uncertain mass on varying angles of slope. (a): Time history of stability margin, $\epsilon_+$. (b) Time history of stability margin, $\epsilon_-$. The case where the object is on the slope whose angle of slope is $20\degree$ is illustrated in \fig{fig:mass_slope_uncertainty_negative}.}
        }
        \label{fig:mass_slope_uncertainty}
\end{figure}

\revise{We present results of objects with uncertain mass with varying angles of slope discussed in Sec~\ref{sec:sec_uncertain_mass_slope}. We consider gear 2 with $\phi = [-20\degree, 0\degree, 20\degree]$ as an angle of slope. 
}

\revise{\fig{fig:mass_slope_uncertainty_positive} and \fig{fig:mass_slope_uncertainty_negative} shows the time history of the stability margin $\epsilon_+$ and $\epsilon_-$, respectively. \fig{fig:mass_slope_uncertainty_positive} shows that the smaller $\phi$ is, the larger $\epsilon_+$ is during the manipulation. $\epsilon_+$ under mass uncertainty considers if contact $B$ is losing as we discuss in \eq{fnb_cond_mass}. \fig{fig:mass_slope_uncertainty_positive} means that contact $B$ can more easily lose contact as $phi$ increases. This makes sense because the larger the angle of slope $\phi$ is, the larger the moment which makes the object rotate along the counter-clockwise direction, resulting in the loss of contact at $B$.
Similarly, $\epsilon_-$ under mass uncertainty considers if contact $A$ is losing as we discuss in \eq{fna_cond_mass}. \fig{fig:mass_slope_uncertainty_negative} means that contact $A$ can more easily lose contact as $\phi$ decreases at $t = \in [0, 15]$ s.
This makes sense because the smaller the angle of slope $\phi$ is, the larger the moment which makes the object rotate along the clockwise direction, resulting in the loss of contact at $A$.
% Note that we observe the 
% 
% Based on the above discussion, we argue that our proposed CIBO can consider uncertain mass with different 
}

\subsection{CIBO for Patch Contact}
% what we want to say:
% patch contact has different trajectory
% analyze the reason and why it does not work
% here we talk about
% 1. patch one shows the better result and successfully 
% 2. not so much
% 3. over the trajectory still shows the better although optimization only cares about the worst-case

\tab{patch_table} shows the computed stability margin considering patch contact shows the greater margins for both positive and negative directions. Hence, we verify that our optimization can still work with patch contact and design the robust controller for maximizing the worst-case stability margin. 
Intuitively, this result makes sense since the contact area increases and the pivoting system has a larger physically feasible space, resulting in a greater stability margin. 

\begin{figure}[t]
    \centering    \includegraphics[width=0.45\textwidth]{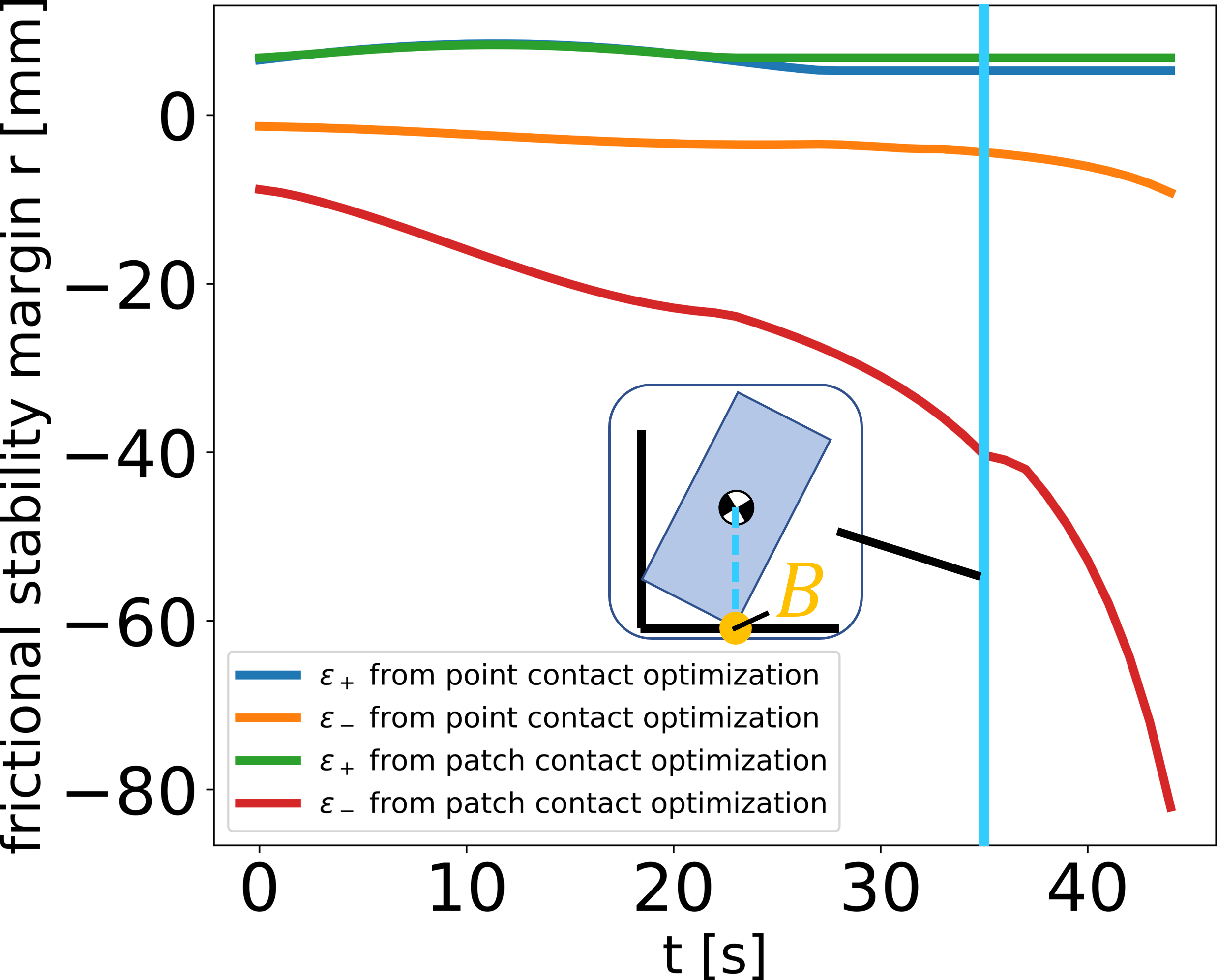} % 
    \caption{Trajectory of frictional stability margin of gear 2 based on uncertain CoM obtained from CIBO using point contact model and patch contact model, respectively. The vertical blue line represents the moment when the projection of CoM lies on the contact $B$.}
    \label{fig:patchcontact_gear2}
\end{figure}

\fig{fig:patchcontact_gear2} illustrates the time history of frictional stability margin of gear 2 from CIBO with considering point contact and with considering patch contact.  Both CIBO with point contact and patch contact have the smallest (i.e., worst-case) stability margin at $t = 0$. However, CIBO with patch contact shows a greater margin at $t=0$, as we discuss above. In addition, over the trajectory, CIBO with patch contact shows a greater margin than that with point contact. Thus, we quantitatively verify that using patch contact is beneficial over the trajectory even though the optimization aims at maximizing the worst-case margin, not the stability margin over the trajectory. It is noted that we are not able to obtain better margins using patch contact due to the non-convexity of the underlying optimization problem.

% \textcolor{red}{YS: do we want to say that patch contact does not always give you better results? in general patch shows the better often, but not always, due to non-convexity of optimization}

% \textcolor{blue}{DJ: We can make such a comment. We should also include the generic contact patch model. Let me work on that. }

\begin{table}[t]
    \caption{{Average Solving Time (AST) comparison between benchmark optimization \eq{equation_control} and CIBO under mass uncertainty using \eq{kkt_convertion} with gear 2.}}
    \centering
    \begin{tabular}{c|c|c}
   $N$  &  AST (s) of \eq{equation_control} & AST (s) of \eq{kkt_convertion}\\
         \hline\hline 
       30 & 0.21 & 0.38 \\
       \hline
      60 & 0.50 & 0.68 \\
      \hline
         120 & 1.01 & 1.24  \\
    \end{tabular}
    \label{computation_benchmark_vs_CIBO}
\end{table}

\subsection{Computation Results}\label{subsec::computation}
\tab{computation_benchmark_vs_CIBO} compares the computation time between benchmark optimization \eq{equation_control} and CIBO under mass uncertainty using \eq{kkt_convertion} for gear 2. Overall, \eq{kkt_convertion} is not so computationally demanding compared to \eq{equation_control}. 
However, as you can see in \tab{computation_friction} and \tab{computation_mode}, once the optimization problem has too many complementarity constraints because of the KKT condition, we clearly observe that the computational time increases. 

\tab{computation_friction} and \tab{computation_mode} shows the computational results for CIBO considering frictional uncertainty \eq{eq:bilvel_friction2} and bilevel optimization over mode sequences  \eq{eq:mode_change}, respectively. 

In general, the computational time for CIBO is larger than the benchmark optimization as CIBO has larger number of complementarity constraints. In the future, we will try to work on better warm-starting strategies so that we might be able to accelerate the optimization.

% Although we do not think our optimization can run in receding-horizon fashion like Model Predictive Control (MPC), we might be able to accelerate the optimization process with a better warm-start strategy. 

\begin{table}[t]
    \caption{{NLP specification for CIBO under frictional uncertainty using \eq{eq:bilvel_friction2} with gear 1.}}
    \centering
    \begin{tabular}{c|c|c|c}
   $N$  & $\#$ of Variables & $\#$ of Constraints & Average Solving Time (s)\\
         \hline\hline 
       30 & 2339 & 2280 & 1.9\\
       \hline
      60 & 4679 & 4560 & 10.6\\
      \hline
         120 & 9359 & 9130 & 30.9 \\
    \end{tabular}
    \label{computation_friction}
\end{table}

\begin{table}[t]
    \caption{{NLP specification for CIBO over mode sequences considering uncertain CoM location using \eq{eq:mode_change} with peg 3.}}
    \centering
    \begin{tabular}{c|c|c|c}
   $N$  & $\#$ of Variables & $\#$ of Constraints & Average Solving Time (s)\\
         \hline\hline 
       30 & 1648 & 1590 & 3.68\\
       \hline
      60 & 3298 & 3180 & 61.6\\
      \hline
         120 & 6598 & 6360 & 73.0 \\
    \end{tabular}
    \label{computation_mode}
\end{table}

\begin{table}[t]
    \caption{{Computed worst-case stability margin considering uncertain CoM location over the control horizon obtained from optimization for gear 2.}}
    \centering
    \begin{tabular}{c|c}
     & $r_{+}^*$, $r_{-}^*$ [mm]\\
         \hline\hline Ours  with point contact  & 5.27, 1.31\\
         % \hline Ours  \eq{kkt_convertion} with mass uncertainty & 0.34, 0.50 & N/A \\
         \hline Ours  with patch contact  & 6.81, 8.82
    \end{tabular}
    \label{patch_table}
\end{table}

\subsection{Hardware Experiments}
We implement our controller using a 6 DoF manipulator to demonstrate the efficacy of our proposed method. In particular, we perform a set of experiments to compare our method against a baseline method using gear 1. 
% \textcolor{red}{Devesh, could you put this explanation in the experiment setup section in section VI-A?}
% \revise{For all these experiments, we design a reference trajectory using the stiffness matrix of the robot. More specifically, we use the following relationship, $x_r[k]=x_{\text{CIBO}}[k]+f_{\text{CIBO}}[k]K_s^{-1}$, where, $x_{\text{CIBO}}$ is the position trajectory, $f_{\text{CIBO}}$ is the force trajectory obtained by CIBO whereas $K_s$ is the stiffness matrix for the robot which is appropriately chosen.}
% We obtain robust trajectories for the objects in \tab{parameter_table} from  \eq{kkt_convertion} and also obtain the benchmark trajectories from \eq{equation_control} using the same hyper parameters.
%  We evaluate the performance of different algorithms using gear 1.
 To evaluate robustness for objects with unknown mass, we solve the optimization with mass different from the true mass of the object and implement the obtained trajectory on the object. 
 % \revise{The true mass of gear 1 is $140$ g, which is used in the trajectory optimization. Then, }
%  To evaluate robustness of the bilevel technique, we solve optimization problem for gear with mass different from the true mass and implement the obtained trajectories. 
 We implement trajectories obtained from the two different optimization techniques 
 using 4 different mass values, \revise{$m = \{100, 110, 140, 170\}$ g. Then, we implement the obtained trajectory on the object with known mass. Note that the actual mass of gear 1 is 140 g. We test the trajectories over 10 trials for the two different methods.} 
%   We implement trajectories obtained from our proposed bilevel optimization and benchmark optimization using 4 different mass values.
%  Thus, in total we implement 8 different trajectories from the two different optimization methods.
 %Because we aim at generating trajectories of pivoting that would make a body stable even under unknown physical parameters (i.e., here, it is mass),

% \begin{table}[t]
%     \caption{{Number of successful pivoting attempts of gear 1 over 10 trials for the two different methods. To evaluate robustness for objects with unknown mass, we solve the optimization with  mass different from the known object and implement the obtained trajectory on the object with known mass. Note that the actual mass of gear 1 is 140 gm. }}
%     \centering
%     \begin{tabular}{c|c|c}
%  & CIBO & Benchmark Optimization\\
%          \hline\hline $m=100$ g  & 10 / 10 & 0 / 10 \\
%          \hline $m=110$ g  & 10 / 10 & 0 / 10 \\
%          \hline $m=140$ g  & 10 / 10 & 0 / 10 \\
%          \hline $m=170$ g  & 10 / 10 & 0 / 10 
%     \end{tabular}
%     \label{hardware_result}
% \end{table}

% \revise{We observe that }
% \tab{hardware_result} shows the success rate of pivoting for the hardware experiments. 
We observe that our proposed bilevel optimization is able to achieve 100 $\%$ success rates for all $4$ mass values while benchmark optimization cannot realize stable pivoting \revise{for all $4$ mass values over 10 trials}. 
% It means that our proposed bilevel optimization is actually able to enhance the robustness of the trajectory based on the frictional stability. 
Note that the benchmark trajectory optimization also generates trajectories with non-zero frictional stability margin but they failed to pivot the object. The reason would be that the system has a number of uncertainties such as incorrect coefficient of friction, sensor noise in the F/T sensor (for implementing the force controller), etc. which are not considered in the model. We believe that these uncertainties make the objects unstable leading to the failure of pivoting.  In contrast, even though CIBO also does not consider these uncertainties, it generates more robust trajectories and we believe that this additional robustness could account for the unknown uncertainty in the real hardware.  We also observe that the trajectories generated by benchmark optimization can successfully realize pivoting if the manipulator uses patch contact during manipulation (thus getting more stability).
%wrong physical parameters (e.g., coefficient of friction) and sensor noise, which are not modeled in frictional stability yet. 

\begin{figure}[t]
    \centering
    \includegraphics[width=0.48\textwidth]{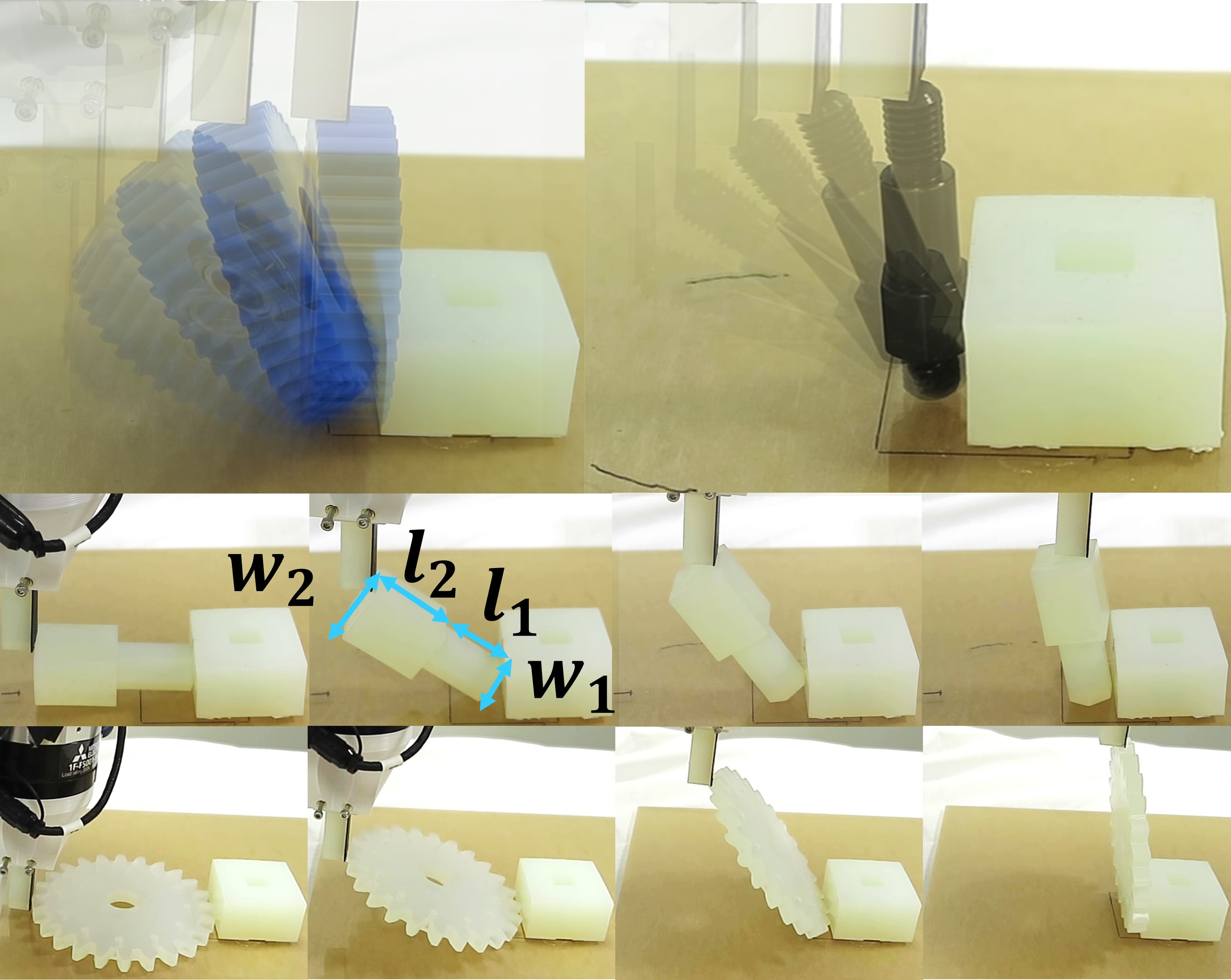} % 
    \caption{Snapshots of hardware experiments. We show snapshots of the white peg and gear (instead of overlaid images) for clarity.}
    \label{fig:hardwareresults}
\end{figure}

\begin{figure}
    \centering
    \includegraphics[width=0.48\textwidth]{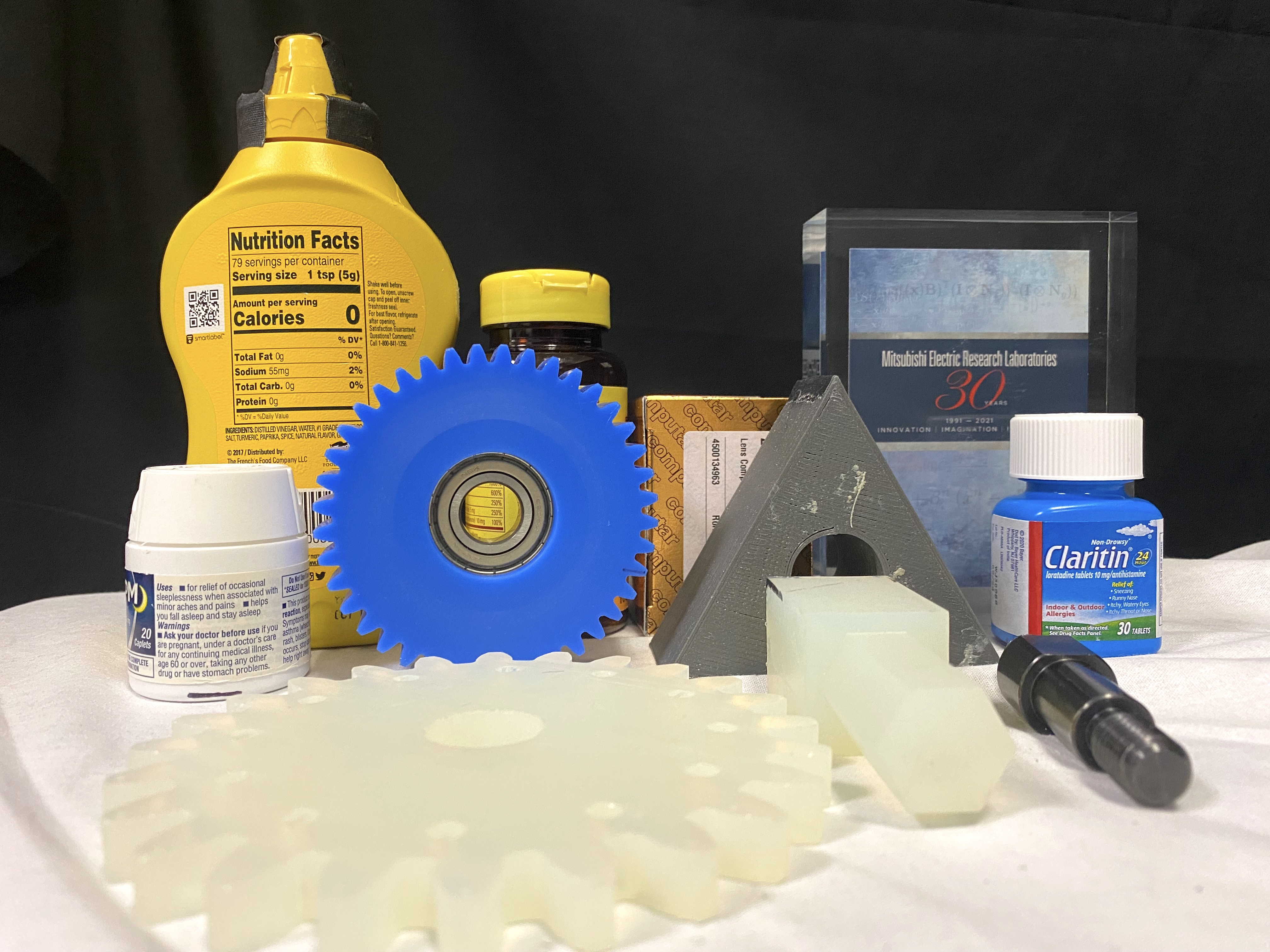} % 
    \caption{The different objects used in hardware evaluation of the proposed method. Please check the hardware experiments results in the video at this link \url{https://www.youtube.com/watch?v=ojlZDaGytSY}.}
    \label{fig:hardware_objects}
\end{figure}

We perform hardware experiments with additional objects to evaluate the generalization of the proposed planning method. All the objects used in the hardware experiments are shown in \fig{fig:hardware_objects}. 
 \fig{fig:hardwareresults} shows the snapshots of hardware experiments for the 4 objects detailed in \tab{parameter_table}. We observe that our bilevel optimization can successfully pivot all the objects during hardware experiments (see \fig{fig:hardwareresults} and the videos). This shows that we can use the proposed method with objects with different size and shape.

%Our bilevel optimization could successfully generate robust trajectories and we could verify their robustness on different objects in real hardware experiments under mass and CoM location uncertainty by providing the inaccurate parameters with the optimizer.  

\subsection{Recovery from Disturbance during Execution}\label{sec:error_recovery}
\begin{figure}
    \centering
    \includegraphics[width=0.42\textwidth]{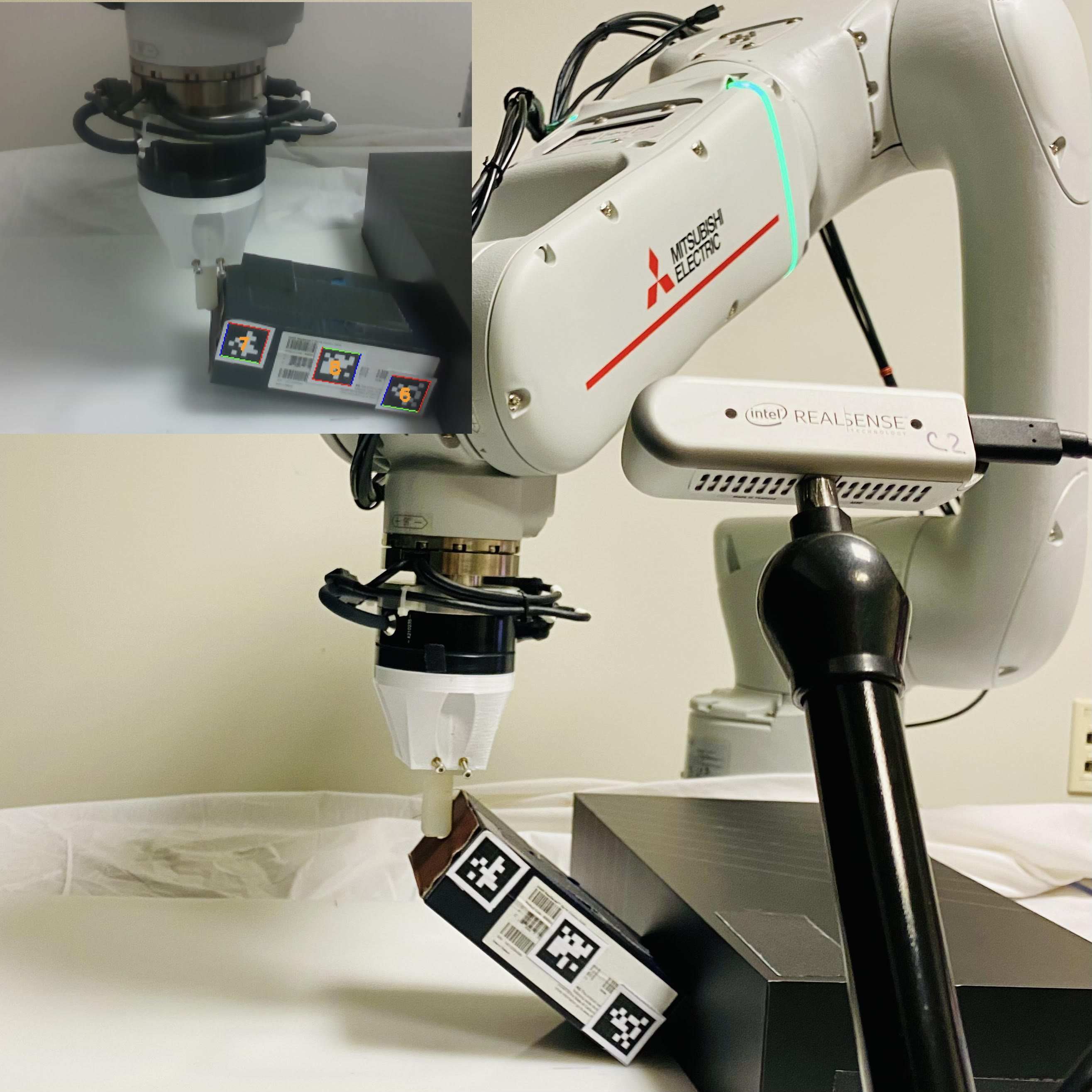} % 
    \caption{\revise{The vision-based feedback pivoting system which can observe the state in real-time and adapt to recover from disturbances during execution. The inset image shows the tracking of the object using an Apriltag system.}}
    \label{fig:camera_tracking_system}
\end{figure}

\begin{figure}
    \centering
    \includegraphics[width=0.48\textwidth]{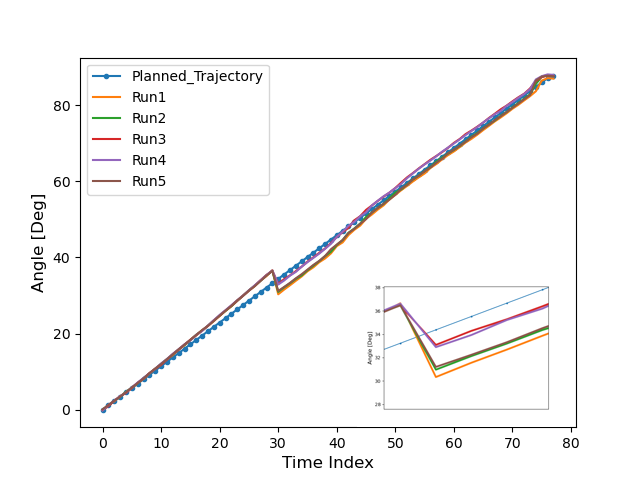} % 
    \caption{\revise{This plot shows recovery from disturbance applied during execution of the robust trajectories obtained from CIBO. The inset image shows the amount of disturbance applied during execution at $t = 30$. As could be seen in the plots, the robot we could successfully recover in all test runs.}}
    \label{fig:error_plot_planning}
\end{figure}

\revise{In the next set of hardware experiments, we present the recovery of the proposed controller from disturbances applied on the object during execution. For performing these experiments we use a cuboid object (see \fig{fig:camera_tracking_system} for the experimental setup, $l=110$, $w=55$, and $m=110$ g). During the execution of the trajectory, we apply random disturbances and record the object orientation using a vision-tracking system. The results from $5$ runs of the trajectory are shown in \fig{fig:error_plot_planning}. As can be seen in the figure, we apply external disturbance on every run of the trajectory at $t=30$. It is noted that the disturbance can not be large enough which results in loss of contact. As long as the contact between the object and the robot is maintained, the robust planner can successfully recover from the disturbance applied during execution and can reach the desired goal (see \fig{fig:error_plot_planning}).}

\begin{figure}
    \centering
    \includegraphics[width=0.48\textwidth]{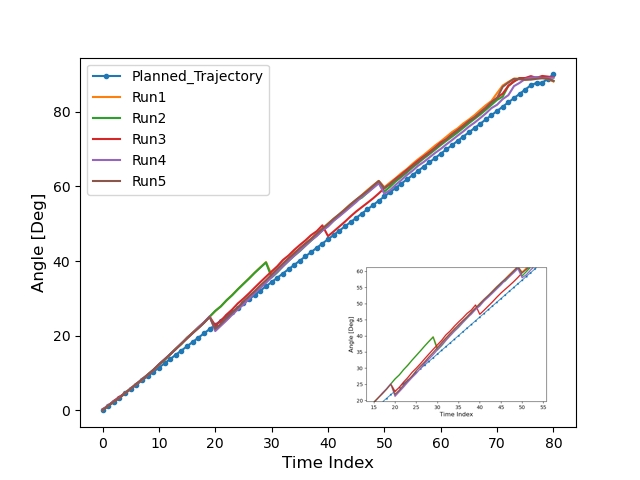} % 
    \caption{\revise{We run the proposed CIBO in MPC fashion with state feedback using the vision system shown in \fig{fig:camera_tracking_system}. We apply multiple disturbances between $t = 20$ and $t = 50$ during execution which could be seen in the zoomed image. We show that due to the closed-loop nature of the controller, the controller is always able to guide the object to the desired goal. }}
    \label{fig:error_plot_mpc}
\end{figure}

\revise{Furthermore, we also implement the algorithm in an MPC fashion to understand if it implements the algorithm in a closed-loop fashion as well as its performance. We use an initial reference trajectory planned by CIBO to initialize the controller. During online execution, we use a trajectory tracking cost function for CIBO. In particular, the vision system is used to estimate $\theta^W$ of the object. For brevity, we abbreviate the superscript $W$ here. The following cost function is used for CIBO:
\begin{equation}
   % c_{\text{MPC}}[k]=
   \lambda_1\sum_{t=k}^T \left(\hat{\theta}_t-\theta^\star_t\right)^2 +\lambda_2 u^2 \nonumber
\end{equation}
where $\theta^\star$ is the initial planned orientation trajectory of the object obtained by CIBO. We optimize the controller after every $10$ control steps till the object reaches the goal. Results of $5$ such runs are shown in \fig{fig:error_plot_mpc}. We apply random disturbances during execution between $t=20$ and $t=50$ as could be seen in the plot (please see the image inset in \fig{fig:error_plot_mpc}). As we can observe from these plots, the controller is successfully able to recover from these disturbances and thus, the controller can always guide the system from any initial state to the desired goal state. This shows that the proposed controller can be used in closed-loop to perform the desired pivoting manipulation. Note that we can not precisely estimate slip between the robot and the object accurately using the vision system. We believe we can generate more complex recovery behavior using additional slip information. However, designing such an estimator requires new hardware and additional work on tactile estimation~\cite{ota2023tactile} which is left as a future exercise. 
}

\section{Discussion and Future Work}\label{sec:discussion}
Generalizable manipulation through contact requires that robots be able to incorporate and account for uncertainties during planning. However, designing the robust controller for achieving such manipulation remains an open problem and remains largely unexplored.  
This paper presents \textit{frictional stability}-aware optimization, a strategy that exploits friction for robust planning of pivoting manipulation. By considering a variety sources of uncertainty such as mass, CoM location, \revise{finger contact location,} and friction coefficients, we discussed the stability margin for pivoting manipulation with slipping contact. We presented CIBO, which solves novel bilevel optimization for pivoting manipulation while optimizing the worst-case stability margin of pivoting manipulation for (non-convex) objects. The proposed method was evaluated in simulation using several test settings. We showed that our proposed bilevel optimization method is able to design trajectories which are robust to larger uncertainties compared to a baseline trajectory optimization method. The proposed method was also demonstrated on a physical robotic system by implementing the computed trajectories on a large variety of objects of different geometries and physical properties. \revise{Furthermore, we also designed an MPC controller using the proposed algorithm which can successfully tracking and regulate the pose of objects during manipulation providing additional robustness during execution.}

Although this paper focuses on pivoting manipulation as a demonstration of our framework, our work can be generalized to other manipulation primitives such as pivoting with one-point contact, pushing, and grasping. This is because our stability margin analysis and CIBO are derived from \revise{quasi-static equilibrium}
\eq{force_eq} and the corresponding friction cone constraints \eq{general_FC}. These conditions are very common across most manipulation problems, and thus our framework can be applicable to the aforementioned manipulation primitives as long as they satisfy \eq{force_eq} and \eq{general_FC}.

There are a number of limitations in this work:

\textbf{Contact-Rich CIBO}
This work assumes that dynamics of an object with quasi-static equilibrium. For objects with non-convex geometry, CIBO is still able to design robust open-loop controller using mode-based optimization. CIBO using mode-based optimization is able to find feasible solutions if users can provide CIBO with physically feasible mode sequences. However, for objects with a very complex shape, it is quite challenging to identify mode sequences prior to optimization~\cite{zhang2023simultaneous}. As a result, CIBO might not be able to find feasible solutions. In order to avoid providing mode sequences, CIBO needs to consider mode sequences by itself. This can be realized by considering complementarity constraints or integer constraints inside the lower-level optimization problem of CIBO. However, as we explained in Sec~\ref{subsec:mode_based_optimization}, CIBO considering these non-convex constraints inside the lower-level optimization problem is not guaranteed to find globally optimal safety margins. 

% Another important limitation regarding the complexity of dynamics is that CIBO is conditioned with states at $t = 0$ (i.e., $x_0$), which is true for other trajectory optimization frameworks. During the implementation of CIBO, we observed that it is not trivial to find "good" $x_0$ and the behavior of CIBO dramatically changes as  $x_0$ changes. For example, for a certain $x_0$, CIBO is able to find a solution while for another $x_0$, CIBO is not able to find a solution. Finding a good $x_0$ is not trivial at all and it requires domain knowledge. Thus, ideally, we should formulate CIBO where $x_0$ is also a decision variable so that the solver can optimize the trajectory over $x_0$ as well. 

\revise{
\textbf{Geometric Uncertainty}
Although we consider a variety of uncertainties, we do not explicitly consider the uncertainty in the geometry of an object. Because geometric uncertainty is one of the main uncertainties due to imperfect vision sensing and can change the contact mode such as from making to breaking contact, considering geometric uncertainty is important. This paper discusses the uncertainty in the finger contact location, which indirectly considers geometric uncertainty. This is because the relative pose between the object and the robot changes due to the finger contact location uncertainty. Since geometric uncertainty also changes the relative pose, we believe considering the uncertainty in robot finger contact location is one way to get started working on geometric uncertainty. 
}

\textbf{Dynamic Manipulation with Uncertainty Propagation}
In this work, we make quasi-static assumption during manipulation. The natural extension of this work is to relax this assumption and consider quasi-dynamic model during manipulation.
To work on these cases, we need to explicitly consider dynamic version of the stability margin. However, this is not trivial. We need to understand how we can propagate uncertainty for contact dynamics as it is not well understood. The stability margin needs to incorporate this uncertainty propagation for such cases. See \cite{shirai2023covariance} for more discussion on uncertainty propagation for contact-rich dynamical systems.
% However, this requires the modification of the model

\textbf{Accurate Contact Mechanics} One of the contributions of this paper is that we consider patch contact. However, in reality, the robot should be able to switch contact mode from patch contact to point contact and vice versa. This enables CIBO to have a larger stability margin but, again, makes the lower-level optimization of CIBO non-convex. 

Another limitation here is modeling of compliant contact. We observed that introducing compliant contact improves the stability of the object. However, modeling compliant contact is difficult. One approach to model compliant contact can be learning-based approach. One of the assumptions of this work is that we consider pivoting in 2D. Thus, extending our work in 3D is promising, which requires the discussion of generalized friction cones \cite{doi:10.1177/027836499401300306}.  

\textbf{Closed Loop Tactile Control} We also implemented the proposed CIBO in a closed-loop fashion with real-time feedback from a vision system.  
\revise{However, once uncertainty of the system is \textit{too} large (e.g., the mass of the object used in CIBO and the actual mass of the object is so different), we observed slipping between the robot end-effector and the object. It is difficult to observe slip using vision sensors alone. 
% However, for most of practical applications we want closed-loop control of manipulation using sensory feedback. 
% Without a closed-loop controller, even the robust trajectories need to be initialized precisely and the system can not recover from a failure. 
In the future, we would like to design a closed-loop controller using tactile sensing to obtain the slip information between the robot and the object for precise closed-loop control~\cite{shirai2023tactile, https://doi.org/10.48550/arxiv.2303.03385}.}

\bibliographystyle{IEEEtran}
\bibliography{main}

% Generated by IEEEtran.bst, version: 1.14 (2015/08/26)
\begin{thebibliography}{10}
\providecommand{\url}[1]{#1}
\csname url@samestyle\endcsname
\providecommand{\newblock}{\relax}
\providecommand{\bibinfo}[2]{#2}
\providecommand{\BIBentrySTDinterwordspacing}{\spaceskip=0pt\relax}
\providecommand{\BIBentryALTinterwordstretchfactor}{4}
\providecommand{\BIBentryALTinterwordspacing}{\spaceskip=\fontdimen2\font plus
\BIBentryALTinterwordstretchfactor\fontdimen3\font minus \fontdimen4\font\relax}
\providecommand{\BIBforeignlanguage}[2]{{%
\expandafter\ifx\csname l@#1\endcsname\relax
\typeout{** WARNING: IEEEtran.bst: No hyphenation pattern has been}%
\typeout{** loaded for the language `#1'. Using the pattern for}%
\typeout{** the default language instead.}%
\else
\language=\csname l@#1\endcsname
\fi
#2}}
\providecommand{\BIBdecl}{\relax}
\BIBdecl

\bibitem{mason2018toward}
M.~T. Mason, ``Toward robotic manipulation,'' \emph{Annual Review of Control, Robotics, and Autonomous Systems}, vol.~1, pp. 1--28, 2018.

\bibitem{drnach2021robust}
L.~Drnach and Y.~Zhao, ``Robust trajectory optimization over uncertain terrain with stochastic complementarity,'' \emph{IEEE Robotics and Automation Letters}, vol.~6, no.~2, pp. 1168--1175, 2021.

\bibitem{9812069}
A.~U. Raghunathan, D.~K. Jha, and D.~Romeres, ``Pyrobocop: Python-based robotic control \& optimization package for manipulation,'' in \emph{2022 International Conference on Robotics and Automation (ICRA)}, 2022, pp. 985--991.

\bibitem{yuki2021chance}
Y.~Shirai, D.~K. Jha, A.~U. Raghunathan, and D.~Romeres, ``Chance-constrained optimization in contact-rich systems,'' in \emph{2023 American Control Conference (ACC)}, 2023, pp. 14--21.

\bibitem{vidyasagar2002nonlinear}
M.~Vidyasagar, \emph{Nonlinear systems analysis}.\hskip 1em plus 0.5em minus 0.4em\relax SIAM, 2002.

\bibitem{raghunathan2020stability}
A.~U. Raghunathan and J.~T. Linderoth, ``Stability analysis of discrete-time linear complementarity systems,'' \emph{arXiv preprint arXiv:2012.13287}, 2020.

\bibitem{hou2018fast}
Y.~Hou, Z.~Jia, and M.~T. Mason, ``Fast planning for 3d any-pose-reorienting using pivoting,'' in \emph{2018 IEEE International Conference on Robotics and Automation (ICRA)}.\hskip 1em plus 0.5em minus 0.4em\relax IEEE, 2018, pp. 1631--1638.

\bibitem{hogan2020tactile}
F.~R. Hogan, J.~Ballester, S.~Dong, and A.~Rodriguez, ``Tactile dexterity: Manipulation primitives with tactile feedback,'' in \emph{2020 IEEE international conference on robotics and automation (ICRA)}.\hskip 1em plus 0.5em minus 0.4em\relax IEEE, 2020, pp. 8863--8869.

\bibitem{shirai2023tactile}
Y.~Shirai, D.~K. Jha, A.~U. Raghunathan, and D.~Hong, ``Tactile tool manipulation,'' in \emph{2023 IEEE International Conference on Robotics and Automation (ICRA)}, 2023, pp. 12\,597--12\,603.

\bibitem{9811812}
Y.~Shirai, D.~K. Jha, A.~U. Raghunathan, and D.~Romeres, ``Robust pivoting: Exploiting frictional stability using bilevel optimization,'' in \emph{2022 International Conference on Robotics and Automation (ICRA)}, 2022, pp. 992--998.

\bibitem{todorov2010implicit}
E.~Todorov, ``Implicit nonlinear complementarity: A new approach to contact dynamics,'' in \emph{2010 IEEE international conference on robotics and automation}.\hskip 1em plus 0.5em minus 0.4em\relax IEEE, 2010, pp. 2322--2329.

\bibitem{drumwright2011evaluation}
E.~Drumwright and D.~A. Shell, ``An evaluation of methods for modeling contact in multibody simulation,'' in \emph{2011 IEEE International Conference on Robotics and Automation}.\hskip 1em plus 0.5em minus 0.4em\relax IEEE, 2011, pp. 1695--1701.

\bibitem{9366782}
C.~de~Farias, N.~Marturi, R.~Stolkin, and Y.~Bekiroglu, ``Simultaneous tactile exploration and grasp refinement for unknown objects,'' \emph{IEEE Robotics and Automation Letters}, vol.~6, no.~2, pp. 3349--3356, 2021.

\bibitem{shirai2022iros}
Y.~Shirai, X.~Lin, A.~Schperberg, Y.~Tanaka, H.~Kato, V.~Vichathorn, and D.~Hong, ``Simultaneous contact-rich grasping and locomotion via distributed optimization enabling free-climbing for multi-limbed robots,'' in \emph{2022 IEEE/RSJ International Conference on Intelligent Robots and Systems (IROS)}, 2022, pp. 13\,563--13\,570.

\bibitem{9561521}
Y.~Zhu, Z.~Pan, and K.~Hauser, ``Contact-implicit trajectory optimization with learned deformable contacts using bilevel optimization,'' in \emph{2021 IEEE International Conference on Robotics and Automation (ICRA)}, 2021, pp. 9921--9927.

\bibitem{9739950}
S.~Shield, A.~M. Johnson, and A.~Patel, ``Contact-implicit direct collocation with a discontinuous velocity state,'' \emph{IEEE Robotics and Automation Letters}, vol.~7, no.~2, pp. 5779--5786, 2022.

\bibitem{Yoshida_Regrasp2009}
E.~Yoshida, M.~Poirier, J.-P. Laumond, O.~Kanoun, F.~Lamiraux, R.~Alami, and K.~Yokoi, ``Regrasp planning for pivoting manipulation by a humanoid robot,'' in \emph{2009 IEEE International Conference on Robotics and Automation}, 2009, pp. 2467--2472.

\bibitem{Dafle_gripper2015}
N.~Chavan-Dafle, M.~T. Mason, H.~Staab, G.~Rossano, and A.~Rodriguez, ``A two-phase gripper to reorient and grasp,'' in \emph{2015 IEEE International Conference on Automation Science and Engineering (CASE)}, 2015, pp. 1249--1255.

\bibitem{jin2021trajectory}
S.~Jin, D.~Romeres, A.~Ragunathan, D.~K. Jha, and M.~Tomizuka, ``Trajectory optimization for manipulation of deformable objects: Assembly of belt drive units,'' \emph{arXiv preprint arXiv:2106.00898}, 2021.

\bibitem{posa2014direct}
M.~Posa, C.~Cantu, and R.~Tedrake, ``A direct method for trajectory optimization of rigid bodies through contact,'' \emph{The International Journal of Robotics Research}, vol.~33, no.~1, pp. 69--81, 2014.

\bibitem{8648229}
A.~Patel, S.~L. Shield, S.~Kazi, A.~M. Johnson, and L.~T. Biegler, ``Contact-implicit trajectory optimization using orthogonal collocation,'' \emph{IEEE Robotics and Automation Letters}, vol.~4, no.~2, pp. 2242--2249, 2019.

\bibitem{shirai2023covariance}
Y.~Shirai, D.~K. Jha, and A.~U. Raghunathan, ``Covariance steering for uncertain contact-rich systems,'' in \emph{2023 IEEE International Conference on Robotics and Automation (ICRA)}, 2023, pp. 7923--7929.

\bibitem{4598894}
T.~Bretl and S.~Lall, ``Testing static equilibrium for legged robots,'' \emph{IEEE Transactions on Robotics}, vol.~24, no.~4, pp. 794--807, 2008.

\bibitem{8383993}
A.~Del~Prete, S.~Tonneau, and N.~Mansard, ``Zero step capturability for legged robots in multicontact,'' \emph{IEEE Transactions on Robotics}, vol.~34, no.~4, pp. 1021--1034, 2018.

\bibitem{8416785}
K.~Hauser, S.~Wang, and M.~R. Cutkosky, ``Efficient equilibrium testing under adhesion and anisotropy using empirical contact force models,'' \emph{IEEE Transactions on Robotics}, vol.~34, no.~5, pp. 1157--1169, 2018.

\bibitem{8358969}
R.~Orsolino, M.~Focchi, C.~Mastalli, H.~Dai, D.~G. Caldwell, and C.~Semini, ``Application of wrench-based feasibility analysis to the online trajectory optimization of legged robots,'' \emph{IEEE Robotics and Automation Letters}, vol.~3, no.~4, pp. 3363--3370, 2018.

\bibitem{9113247}
Y.~Shirai, X.~Lin, Y.~Tanaka, A.~Mehta, and D.~Hong, ``Risk-aware motion planning for a limbed robot with stochastic gripping forces using nonlinear programming,'' \emph{IEEE Robotics and Automation Letters}, vol.~5, no.~4, pp. 4994--5001, 2020.

\bibitem{dai2016planning}
H.~Dai and R.~Tedrake, ``Planning robust walking motion on uneven terrain via convex optimization,'' in \emph{2016 IEEE-RAS 16th International Conference on Humanoid Robots (Humanoids)}.\hskip 1em plus 0.5em minus 0.4em\relax IEEE, 2016, pp. 579--586.

\bibitem{8289420}
H.~Audren and A.~Kheddar, ``3-d robust stability polyhedron in multicontact,'' \emph{IEEE Transactions on Robotics}, vol.~34, no.~2, pp. 388--403, 2018.

\bibitem{hou2020manipulation}
Y.~Hou, Z.~Jia, and M.~Mason, ``Manipulation with shared grasping,'' in \emph{Robotics: Science and Systems}, 2020.

\bibitem{donlon2018gelslim}
E.~Donlon, S.~Dong, M.~Liu, J.~Li, E.~Adelson, and A.~Rodriguez, ``Gelslim: A high-resolution, compact, robust, and calibrated tactile-sensing finger,'' in \emph{2018 IEEE/RSJ International Conference on Intelligent Robots and Systems (IROS)}.\hskip 1em plus 0.5em minus 0.4em\relax IEEE, 2018, pp. 1927--1934.

\bibitem{li2020f}
W.~Li, A.~Alomainy, I.~Vitanov, Y.~Noh, P.~Qi, and K.~Althoefer, ``F-touch sensor: Concurrent geometry perception and multi-axis force measurement,'' \emph{IEEE Sensors Journal}, vol.~21, no.~4, pp. 4300--4309, 2020.

\bibitem{DanicaIn-hand2015}
F.~E. Viña~B., Y.~Karayiannidis, K.~Pauwels, C.~Smith, and D.~Kragic, ``In-hand manipulation using gravity and controlled slip,'' in \emph{2015 IEEE/RSJ International Conference on Intelligent Robots and Systems (IROS)}, 2015, pp. 5636--5641.

\bibitem{DanicaIn-hand2016}
F.~E. Viña~B., Y.~Karayiannidis, C.~Smith, and D.~Kragic, ``Adaptive control for pivoting with visual and tactile feedback,'' in \emph{2016 IEEE International Conference on Robotics and Automation (ICRA)}, 2016, pp. 399--406.

\bibitem{Cruciani_inhand2017}
S.~Cruciani and C.~Smith, ``In-hand manipulation using three-stages open loop pivoting,'' in \emph{2017 IEEE/RSJ International Conference on Intelligent Robots and Systems (IROS)}, 2017, pp. 1244--1251.

\bibitem{Dafle_Extrinsic2014}
N.~C. Dafle, A.~Rodriguez, R.~Paolini, B.~Tang, S.~S. Srinivasa, M.~Erdmann, M.~T. Mason, I.~Lundberg, H.~Staab, and T.~Fuhlbrigge, ``Extrinsic dexterity: In-hand manipulation with external forces,'' in \emph{2014 IEEE International Conference on Robotics and Automation (ICRA)}, 2014, pp. 1578--1585.

\bibitem{aceituno2020global}
B.~Aceituno-Cabezas and A.~Rodriguez, ``A global quasi-dynamic model for contact-trajectory optimization,'' in \emph{Robotics: Science and Systems (RSS)}, 2020.

\bibitem{han2020local}
W.~Han and R.~Tedrake, ``Local trajectory stabilization for dexterous manipulation via piecewise affine approximations,'' in \emph{2020 IEEE International Conference on Robotics and Automation (ICRA)}.\hskip 1em plus 0.5em minus 0.4em\relax IEEE, 2020, pp. 8884--8891.

\bibitem{yalmip_2018}
\BIBentryALTinterwordspacing
``Equalities with uncertainty,'' Sep 2018. [Online]. Available: \url{https://yalmip.github.io/equalityinuncertainty}
\BIBentrySTDinterwordspacing

\bibitem{80fe29bf9dc245ffa5c8bd7b3eee2902}
A.~W{\"a}chter and L.~Biegler, ``\BIBforeignlanguage{English (US)}{On the implementation of an interior-point filter line-search algorithm for large-scale nonlinear programming},'' \emph{\BIBforeignlanguage{English (US)}{Mathematical Programming}}, vol. 106, no.~1, pp. 25--57, May 2006.

\bibitem{hsl2007collection}
A.~HSL, ``collection of fortran codes for large-scale scientific computation,'' \emph{See http://www. hsl. rl. ac. uk}, 2007.

\bibitem{stiffness_control}
J.~K. Salisbury, ``Active stiffness control of a manipulator in cartesian coordinates,'' in \emph{1980 19th IEEE Conference on Decision and Control including the Symposium on Adaptive Processes}, 1980, pp. 95--100.

\bibitem{9838102}
D.~K. Jha, D.~Romeres, W.~Yerazunis, and D.~Nikovski, ``Imitation and supervised learning of compliance for robotic assembly,'' in \emph{2022 European Control Conference (ECC)}, 2022, pp. 1882--1889.

\bibitem{https://doi.org/10.48550/arxiv.2204.10447}
\BIBentryALTinterwordspacing
D.~K. Jha, D.~Romeres, S.~Jain, W.~Yerazunis, and D.~Nikovski, ``Design of adaptive compliance controllers for safe robotic assembly,'' 2022. [Online]. Available: \url{https://arxiv.org/abs/2204.10447}
\BIBentrySTDinterwordspacing

\bibitem{apriltag_2011icra}
E.~Olson, ``Apriltag: A robust and flexible visual fiducial system,'' in \emph{2011 IEEE International Conference on Robotics and Automation}, 2011, pp. 3400--3407.

\bibitem{ota2023tactile}
K.~Ota, D.~K. Jha, H.-Y. Tung, and J.~Tenenbaum, ``{Tactile-Filter: Interactive Tactile Perception for Part Mating},'' in \emph{Proceedings of Robotics: Science and Systems}, Daegu, Republic of Korea, July 2023.

\bibitem{zhang2023simultaneous}
M.~Zhang, D.~K. Jha, A.~U. Raghunathan, and K.~Hauser, ``{Simultaneous Trajectory Optimization and Contact Selection for Multi-Modal Manipulation Planning},'' in \emph{Proceedings of Robotics: Science and Systems}, Daegu, Republic of Korea, July 2023.

\bibitem{doi:10.1177/027836499401300306}
M.~Erdmann, ``On a representation of friction in configuration space,'' \emph{The International Journal of Robotics Research}, vol.~13, no.~3, pp. 240--271, 1994.

\bibitem{https://doi.org/10.48550/arxiv.2303.03385}
\BIBentryALTinterwordspacing
S.~Kim, D.~K. Jha, D.~Romeres, P.~Patre, and A.~Rodriguez, ``Simultaneous tactile estimation and control of extrinsic contact,'' 2023. [Online]. Available: \url{https://arxiv.org/abs/2303.03385}
\BIBentrySTDinterwordspacing

\end{thebibliography}

\begin{IEEEbiography}[{\includegraphics[width=1in,height=1.25in,clip,keepaspectratio]{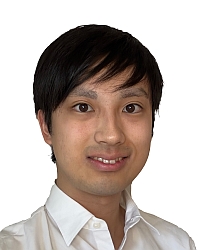}}]{Yuki Shirai}
(Member, IEEE) received the B.E. in mechanical and aerospace engineering from Tohoku University, Sendai, Japan, in 2018, and the M.S. and the Ph.D. in mechanical engineering from the University of California Los Angeles, Los Angeles, CA, USA, in 2019 and 2024, respectively.

He is currently a Postdoctoral Research Fellow with Optimization \& Intelligent Robotics team at Mitsubishi Electric Research Laboratories. His research interests lie in the intersection of optimization and learning for contact-rich robotic manipulation and locomotion. 
\end{IEEEbiography}

\begin{IEEEbiography}[{\includegraphics[width=1in,height=1.25in,clip,keepaspectratio]{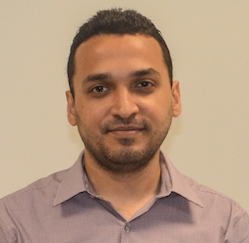}}]{Devesh Jha} (Senior Member, IEEE) is a Senior Principal Research Scientist at Mitsubishi Electric Research Laboratories (MERL) in Cambridge, MA, USA. He received PhD in Mechanical Engineering from Penn State in Decemeber 2016. He received M.S. degrees in Mechanical Engineering and Mathematics also from Penn State. 

His research interests are in the areas of Robotics and Machine Learning. He is a recipient of several best paper awards including the Kalman Best Paper Award 2019 from the American Society of Mechanical Engineers (ASME) Dynamic Systems and Control Division. He is a senior member of IEEE and an associate editor of IEEE Robotics and Automation Letters (RA-L).
\end{IEEEbiography}

\begin{IEEEbiography}[{\includegraphics[width=1in,height=1.25in,clip,keepaspectratio]{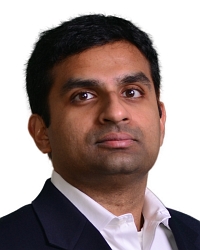}}]{Arvind U Raghunathan} received his Bachelor of Technology degree (summa cum laude) from Indian Institute of Technology, Madras and Ph.D. degree Carnegie Mellon University both in Chemical Engineering in 1999 and 2004, respectively. He is currently the Senior Principal Research Scientist and Senior Team Leader of the Optimization \& Intelligent Robotics Team at Mitsubishi Electric Research Laboratories, Cambridge, MA, USA. 

His research interests include development of optimization algorithms and their applications to electric power operations,  control of robotic systems, and operations of transportation systems. Arvind's research has been recognized with the 2022 IEEE Control Systems Society Roberto Tempo Best CDC Paper Award.
\end{IEEEbiography}

\end{document}